\documentclass{IEEEoj}
\usepackage{cite}
\usepackage{amsmath,amssymb,amsfonts}
\usepackage{algorithmic}
\usepackage{graphicx,color}
\usepackage{textcomp}

\usepackage[utf8]{inputenc} % allow utf-8 input
\usepackage[T1]{fontenc}    % use 8-bit T1 fonts
\usepackage{booktabs}
\usepackage{array}
\usepackage{tabularx}
\usepackage{xcolor}
\usepackage{colortbl}
\usepackage{caption}
\usepackage{ragged2e}
\usepackage{makecell}

\usepackage{hyperref}       % hyperlinks
\usepackage{url}            %     % blackboard math symbols
\usepackage{nicefrac}       % compact symbols for 1/2, etc.
\usepackage{microtype}      % microtypography

\usepackage{lipsum}

\def\BibTeX{{\rm B\kern-.05em{\sc i\kern-.025em b}\kern-.08em
    T\kern-.1667em\lower.7ex\hbox{E}\kern-.125emX}}
\AtBeginDocument{\definecolor{ojcolor}{cmyk}{0.93,0.59,0.15,0.02}}
\def\OJlogo{\vspace{-14pt}\includegraphics[height=28pt]{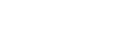}}

\begin{document}
% \receiveddate{XX Month, XXXX}
% \reviseddate{XX Month, XXXX}
% \accepteddate{XX Month, XXXX}
% \publisheddate{XX Month, XXXX}
% \currentdate{XX Month, XXXX}
% \doiinfo{OJITS.2022.1234567}

\title{Artificial Intelligence for Modeling and Simulation of Mixed Automated and Human Traffic}

\author{Saeed Rahmani\authorrefmark{1}, 
Shiva Rasouli\authorrefmark{2}, 
Daphne Cornelisse\authorrefmark{3}, 
Eugene Vinitsky\authorrefmark{3}, 
Bart van Arem\authorrefmark{1}, Simeon C. Calvert\authorrefmark{1}}
\affil{Department of Transport \& Planning, Delft University of Technology, Delft, the Netherlands.}
\affil{Department of Industrial and Systems Engineering, University of Michigan, Dearborn, Michigan, USA}
\affil{Tandon School of Engineering, New York University, New York, United States.}
\corresp{CORRESPONDING AUTHOR: Saeed Rahmani (e-mail: s.rahmani@tudelft.nl).}
\authornote{This work was supported by EU Horizon 2020 Project Hi-Drive.}
\markboth{Artificial Intelligence for Modeling and Simulation of Mixed Automated and Human Traffic}{Rahmani \textit{et al.}}

\begin{abstract}
Autonomous vehicles (AVs) are now operating on public roads, which makes their testing and validation more critical than ever. Simulation offers a safe and controlled environment for evaluating AV performance in varied conditions. However, existing simulation tools mainly focus on graphical realism and rely on simple rule-based models and therefore fail to accurately represent the complexity of driving behaviors and interactions. Artificial intelligence (AI) has shown strong potential to address these limitations; however, despite the rapid progress across AI methodologies, a comprehensive survey of their application to mixed autonomy traffic simulation remains lacking. Existing surveys either focus on simulation tools without examining the AI methods behind them, or cover ego-centric decision-making without addressing the broader challenge of modeling surrounding traffic. Moreover, they do not offer a unified taxonomy of AI methods covering individual behavior modeling to full scene simulation. To address these gaps, this survey provides a structured review and synthesis of AI methods for modeling AV and human driving behavior in mixed autonomy traffic simulation. We introduce a taxonomy that organizes methods into three families: \textit{agent-level behavior models}, \textit{environment-level simulation methods}, and \textit{cognitive and physics-informed methods}. The survey analyzes how existing simulation platforms fall short of the needs of mixed autonomy research and outlines directions to narrow this gap. It also provides a chronological overview of AI methods and reviews evaluation protocols and metrics, simulation tools, and datasets. By covering both traffic engineering and computer science perspectives, we aim to bridge the gap between these two communities.
\end{abstract}



\maketitle

\section{INTRODUCTION}

Autonomous vehicles (AVs) are now operating on public roads, a development that is reshaping transportation systems worldwide. As their deployment accelerates, the demand for rigorous, scalable, and safe testing environments increases. Simulation has emerged as a viable tool for this purpose by offering a cost-effective and controlled environment for evaluating AV performance across a wide range of conditions and driving scenarios\cite{farah2022modeling, kazemkhani2025gpudrive}. 

Nevertheless, a fundamental challenge in current microscopic traffic simulation is the realism of traffic participants' behaviors. Existing popular simulation platforms like SUMO~\cite{lopez2018microscopic}, VISSIM~\cite{vissim2021}, and CARLA~\cite{dosovitskiy2017carla} typically model other vehicles' behavior using simple rule-based models or by replaying recorded trajectories. This simplified assumption fails to accurately represent the dynamic, interactive complexity of real-world driving. Traditional rule-based models, such as the Intelligent Driver Model for car-following~\cite{treiber2000congested} or MOBIL for lane changing~\cite{kesting2007general}, capture important aggregate phenomena but struggle to reproduce the full distribution of individual-level behavioral variability and adaptation observed in real traffic~\cite{schwarting2019social, simon1955behavioral}. This limitation becomes especially important in the era of mixed autonomy as AVs are gradually integrated into human-dominated road networks and must navigate environments characterized by behavioral heterogeneity.

To address these limitations, researchers in automated driving and traffic simulation have increasingly relied on artificial intelligence (AI) and deep learning methods. Recent breakthroughs in generative AI, foundation models, and learned world models have opened up new possibilities for synthesizing diverse, realistic, and highly interactive traffic agents and scenarios~\cite{mao2023agentdriver, wen2023dilu,gu2022stochastic, jiang2023motiondiffuser}.  Techniques such as imitation learning~\cite{pomerleau1989alvinn, bojarski2016end, codevilla2018end}, reinforcement learning~\cite{shalev2016safe, kendall2019learning}, and world models~\cite{russell2025gaia2} are now widely used in simulation and, increasingly, in real-world AV deployments. Nevertheless, despite this rapid methodological progress, a comprehensive and structured taxonomy that links these advances to the specific challenges of mixed autonomy traffic simulation and evaluation remains missing. Existing surveys either focus on simulation tools without examining the AI methods behind them~\cite{li2024choose, farah2022modeling}, or they focus on AV decision-making from an ego-centric perspective without addressing how surrounding traffic should be modeled or the specific requirements that arise in interactive simulation and mixed autonomy traffic~\cite{di2021mixed, schwarting2018planning}. Moreover, existing studies do not cover the full methodological range, from individual agent behavior models to environment-level scene generation, and from purely data-driven approaches to cognitive and physics-informed models.

Motivated by these gaps, this survey provides a comprehensive review and taxonomy of AI-driven methods for modeling both automated and human-driven vehicle behavior in mixed autonomy traffic simulation. Our methodological taxonomy uses several complementary dimensions. At the agent level, we distinguish between \textit{single-agent methods}, which model individual vehicle behavior, and \textit{multi-agent methods}, which explicitly capture the behaviors and bidirectional interactions among multiple agents. We then separate agent-level behavior models from \textit{environment-level} methods, such as generative world models and scenario generation frameworks. To complement these pure data-driven approaches, we survey \textit{cognitive and physics-informed} methods that incorporate theories of human attention, risk perception, decision-making, and vehicle dynamics, into learning-based architectures. Within each category, we classify methods by their core technical formulation, cover the formal problem setup, highlight representative studies, and discuss current limitations and open research directions. Beyond the methodological overview and synthesis, we provide a chronological investigation of these methods and review evaluation and benchmarking practices, simulation tools, and publicly available datasets that support the development, training, and validation of these models.

Through these contributions, this survey aims to serve as a unified and cross-disciplinary resource that connects methodological advances in machine learning with the practical demands of mixed autonomy traffic simulation. It is intended for researchers and practitioners in transportation and traffic engineering, robotics, and computer science who seek a common methodological understanding across these disciplines. The rest of this paper is organized as follows. Section~\ref{sec:methodology} describes our scope and taxonomy, as well as how we structure the survey. Section~\ref{sec:related_surveys} reviews related surveys. Sections~\ref{sec:single_agent}--\ref{sec:cognitive_ai} form the methodological core and cover agent-level, environment-level, and cognitive and physics-informed methods.
Sections~\ref{sec:timeline}--\ref{sec:dataset_sim} present a chronological overview of method evolution, evaluation and benchmarking practices, and simulation tools and datasets. Section~\ref{sec:discussion} discusses open challenges and future directions, and Section~\ref{sec:conclusion} concludes the survey.

\section{SURVEY SCOPE AND METHODOLOGY}
\label{sec:methodology}

\begin{figure*}
    \centering
    \includegraphics[width=0.95\linewidth]{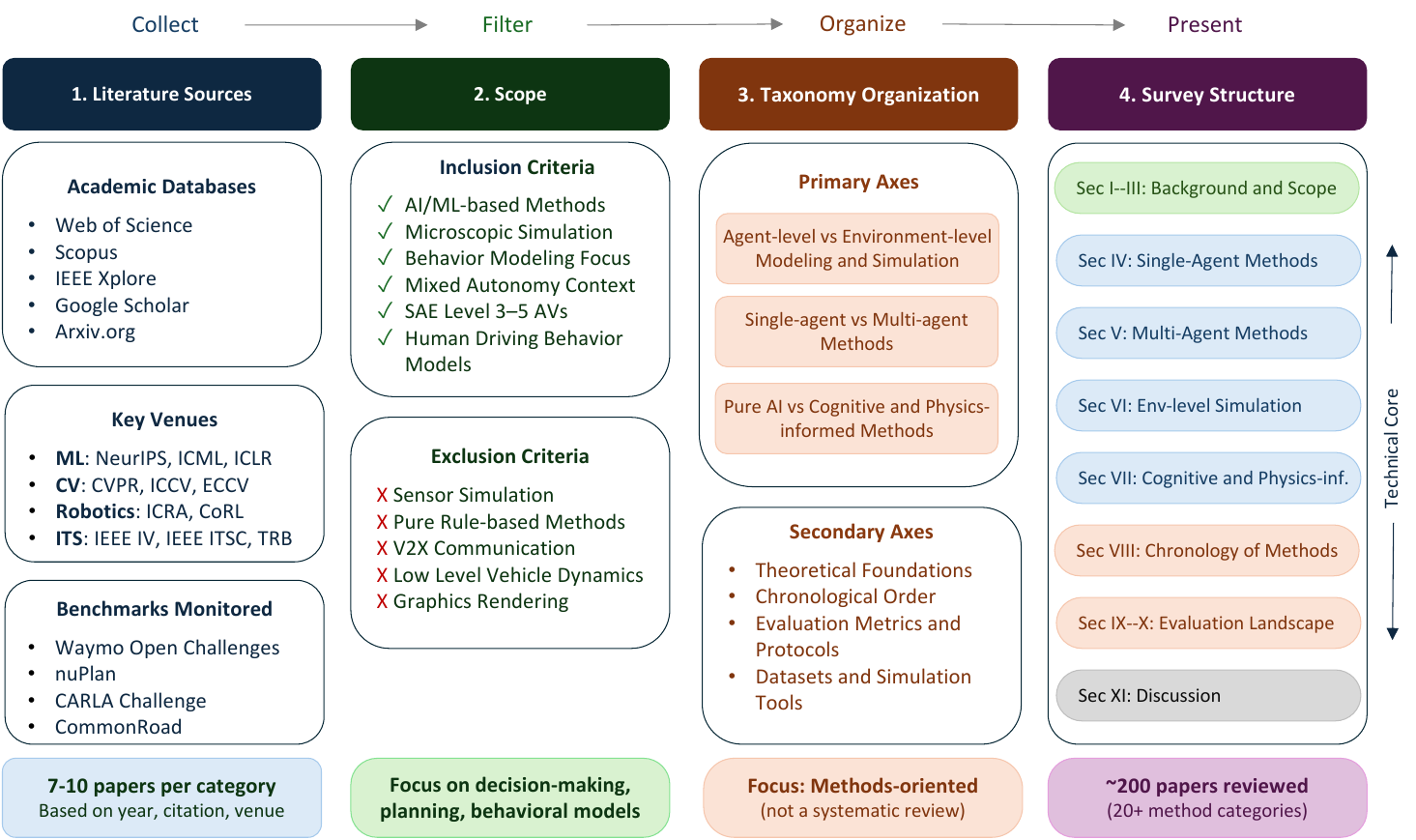}
    \caption{Survey methodology, scope, and structure}
    \label{fig:survey_methodology_structure}
\end{figure*}

Surveying AI methods for mixed autonomy traffic simulation requires navigating a broad and fragmented literature spanning transportation engineering, machine learning, and cognitive science. This survey therefore adopts a \emph{methods-oriented} approach rather than a systematic review protocol, with the goal of organizing these methods into a coherent taxonomy, defining core ideas and terminology, discussing strengths and limitations, highlighting representative studies, and outlining future directions. The following subsections detail our scope, literature selection criteria, and the resulting taxonomy, and Figure~\ref{fig:survey_methodology_structure} provides an overview of survey methodology, scope, and structure. 

\subsection{Scope}
\label{subsec:scope}

Our focus is on AI-based methods for \textit{behavior modeling} in microscopic simulation of mixed autonomy traffic. We cover both human driver models (SAE Level~0) and automated vehicle models, with emphasis on SAE Level~4 and~5 systems where the automated driving system performs the full dynamic driving task without requiring a handover to a human operator. We also include SAE Level~3 automated vehicles when they operate in automated mode. As a result, methods that model control transitions between automation and human drivers are outside the scope of this survey.

The survey covers machine learning and deep learning methods used for decision-making, motion planning, trajectory generation, and prediction. While the main focus is motorized vehicles, we also include methods for pedestrians and cyclists, where necessary, due to their role in mixed traffic. We also review evaluation and benchmarking practices, simulation platforms, and public datasets used to develop and validate these models.

Several topics are outside the scope of this survey. These include sensor and perception simulation, such as camera, LiDAR, and radar synthesis, detailed vehicle dynamics and low-level control, vehicle-to-everything communication protocols, graphics rendering and visualization, real-world deployment and regulatory validation frameworks, and non-ground transportation systems. Traditional rule-based models are discussed for context, but they are not the primary focus.

\subsection{Literature Search and Selection Criteria}
\label{subsec:literature_search}

We collected literature through searches in Google Scholar, Web of Science, Scopus, IEEE Xplore, and arXiv. We used keyword queries that combine automated driving and mixed autonomy terms (for example, ``automated driving,'' ``autonomous or automated vehicles,'' and ``mixed autonomy'') with representative AI method keywords and behavior-modeling task keywords (for example, motion prediction, trajectory generation, driver modeling, and scenario generation). We monitored key venues from both the machine learning and transportation communities. Citation tracking and snow balling was also used from foundational and recent papers to identify additional relevant works. We also tracked major benchmarks and challenges, including the Waymo Open Motion Dataset~\cite{ettinger2021womd}, Argoverse~\cite{wilson2021argoverse2}, nuPlan~\cite{caesar2021nuplan}, and the Waymo Open Sim Agents Challenge (WOSAC)~\cite{montali2023wosac}. For cognitive and physics-informed methods, we also explored the human factors and cognitive science literature, including journals focusing on Human Factors and Cognitive Science.

For each methodological category, we selected 7--15 representative papers based on five criteria: (1)~a foundational contribution that introduced a new formulation or problem setting, (2)~a clear methodological step beyond prior work, (3)~impact and community adoption (citation count), (4)~recent work with strong results on recognized benchmarks, and (5)~coverage of different technical approaches within the category. For emerging topics such as foundation-model-based driving and cognitive and learning integration, we placed more weight on novelty and coverage, and less weight on citation counts.

\subsection{Taxonomy and Survey Structure}
\label{subsec:taxonomy_structure}

\begin{figure*}
    \centering
    \includegraphics[width=0.95\linewidth]{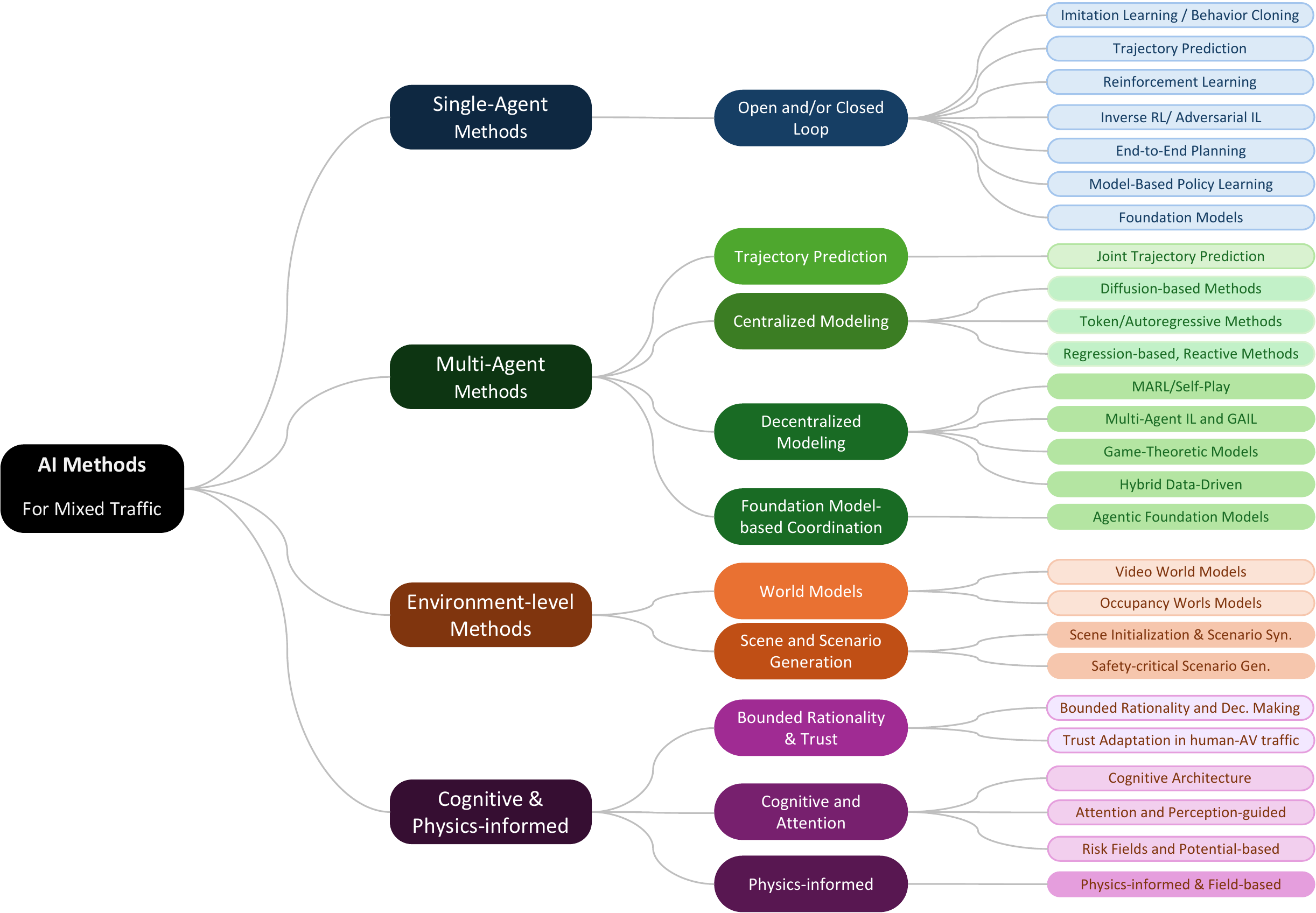}
    \caption{Taxonomy of AI methods for modeling mixed-autonomy traffic}
    \label{fig:taxonomy}
\end{figure*}

The survey is organized around a taxonomy that follows three main axes. These axes reflect key differences in problem formulation and in how models are used in simulation. Figure~\ref{fig:taxonomy} provides an overview of the full taxonomy, and the section structure follows the same organization.

The first axis distinguishes \emph{single-agent} and \emph{multi-agent} methods. Single-agent methods (Section~\ref{sec:single_agent}) focus on learning the behavior of one agent, and they treat other agents as part of the environment. Multi-agent methods (Section~\ref{sec:multi_agent}) model interactions explicitly and aim to capture coupled decision-making among agents. Within multi-agent methods, we separate joint trajectory forecasting from interactive simulation methods. We then divide interactive simulation into centralized approaches, where one learned model produces the next states for all agents, and decentralized approaches, where each agent has its own policy and scene-level behavior results from their coupled execution.

The second axis distinguishes \emph{agent-level} and \emph{environment-level} methods. Agent-level methods (Sections~\ref{sec:single_agent}--\ref{sec:multi_agent}) output agent actions or trajectories, which define behavior in the scene. Environment-level methods (Section~\ref{sec:environment_level}) model how the world evolves and provide the conditions under which behavior models operate. In this part, we cover generative world models and traffic scenario generation methods. 

The third axis distinguishes \emph{data-driven} and \emph{theory-informed} methods. The main body of the survey in Sections \ref{sec:single_agent} to \ref{sec:environment_level} covers data-driven methods organized by the first two axes. Section~\ref{sec:cognitive_ai} covers cognitive and physics-informed methods that incorporate domain knowledge from cognitive psychology, human factors, and traffic physics. 

Within each category, we organize methods by their main technical foundation, and identify classes of methods, such as imitation learning, reinforcement learning, diffusion-based generation, and game-theoretic reasoning. We treat the open-loop versus closed-loop distinction as an important deployment choice, but not as a main taxonomy dimension, since many methods can support both settings. Beyond this taxonomy of methods, Section~\ref{sec:timeline} summarizes the evolution of methods over time. Section~\ref{sec:evaluation} reviews evaluation and benchmarking practices, including open-loop metrics, closed-loop protocols, and realism assessment, and Section~\ref{sec:dataset_sim} reviews simulation tools, datasets, and benchmarks used to train and validate these models.

\section{RELATED SURVEYS}
\label{sec:related_surveys}

Table~\ref{tab:related_surveys} summarizes the scope of the most relevant existing surveys and highlights their contributions and the complementary coverage offered by this survey. Early foundational surveys established the landscape of AV decision-making and control. Schwarting et al.~\cite{schwarting2018planning} provided an overview of planning and decision-making paradigms, covering optimization-based, sampling-based, and early learning-based methods. This study adopts a purely ego-vehicle perspective and pre-dated the transformer and foundation model era. Kuutti et al.~\cite{kuutti2020survey} surveyed deep learning applications to autonomous vehicle control by covering end-to-end learning, behavior cloning, and deep RL for lateral and longitudinal control tasks. However, they do not cover traffic agent modeling or multi-agent interaction and remain focused on ego vehicle's control. Grigorescu et al.~\cite{grigorescu2020survey} surveyed deep learning across the full autonomous driving stack, perception, planning, and control, but without addressing multi-agent simulation or mixed autonomy aspects. Kiran et al.~\cite{kiran2021drl_survey} taxonomized deep reinforcement learning for autonomous driving and extended the scope to cover behavior cloning, inverse RL, and simulation platforms, but the treatment remains limited to reinforcement learning, ego-centric, and does not consider surrounding traffic modeling. More recently, Chen et al.~\cite{chen2024e2e_survey} reviewed over 270 papers on end-to-end autonomous driving, providing an in-depth analysis of imitation learning, reinforcement learning, and world model-based approaches, but their scope is restricted to the end-to-end paradigm and is ego-centric only.

On the traffic simulation and mixed autonomy side, Di and Shi~\cite{di2021mixed} provided the first survey explicitly bridging transportation engineering and AI for mixed autonomy AV control. Although their cross-disciplinary scope is the closest antecedent to ours, the focus is on \textit{AV control policy} design rather than on the broader landscape of behavior and interaction modeling methods for traffic simulation and multi-agent setup. Moreover, recent foundation and world model techniques, as well as environment-level simulation methods, are not discussed in their work. Farah et al.~\cite{farah2022modeling} addressed the practical challenges of modeling automated driving in microscopic traffic simulations for traffic performance evaluations, identified six key modeling aspects, and reviewing how existing simulation studies handle mixed traffic with varying automation levels. Their work highlights the gap between simulation practice and behavioral realism but does not cover AI or deep learning methods. Chao et al.~\cite{chao2020visual} surveyed visual traffic simulation models at multiple scales and discussed early data-driven animation techniques for AV testing. But their study stays focused on graphical simulation. Chen et al.~\cite{chen2024datadriven} provided a review of data-driven traffic simulation and covered imitation learning, RL, deep generative models, though without addressing cognitive or physics-informed methods, world models, and environment-level simulation and scenario generation techniques. Other related, but more narrowly scoped, surveys include the work of
Ding et al.~\cite{ding2023scenario} on safety-critical scenario generation methods, Li et al.~\cite{li2024choose}, who reviewed three decades of open-source AV simulators. They categorized over 20 platforms but did not examine the AI methods behind the simulated behaviors.

In summary, prior surveys have made valuable contributions by mapping key aspects of the automated driving and AI landscape, and this survey builds upon their foundations. However, taken together, they leave several important gaps when viewed from a simulation and evaluation perspective, especially in the context of mixed automated and human traffic. First, to our knowledge, no existing survey spans the complete methodological spectrum from agent-level methods (single-agent and multi-agent) through environment-level approaches (world models and scenario generation) to cognitive and physics-informed models within one unified taxonomy. Second, they do not address the full set of modeling and deployment requirements that arise in traffic simulation, such as the open-loop versus closed-loop gap, counterfactual validity under intervention, long-horizon rollout stability, and interaction consistency among heterogeneous agents. Third, prior works generally lack a dedicated, mixed-autonomy-focused review of evaluation practice, including the metrics, benchmarks, datasets, and simulation tools through which models are validated and compared in interactive traffic settings. Finally, our survey is distinguished by its explicit focus on behavior modeling for both automated vehicles and human drivers, treating the interaction dynamics between these heterogeneous agents as the central object of study rather than adopting a purely ego-vehicle or purely traffic-flow perspective.

\begin{table*}[t]
\centering
\caption{Comparison of this survey with related surveys across key dimensions. \checkmark\,= covered; (\checkmark)\,= partially covered; -- = not covered.}
\label{tab:related_surveys}
\renewcommand{\arraystretch}{1.15}
\footnotesize
\begin{tabular}{l c c c c c c c c}
\hline
\textbf{Survey} &
\parbox{1.3cm}{\vspace{5pt}\centering\textbf{Mixed\\Autonomy\\Focus}\vspace{5pt}} &
\parbox{1.3cm}{\vspace{5pt}\centering\textbf{Traffic\\Agent\\Model.}\vspace{5pt}} &
\parbox{1.3cm}{\vspace{5pt}\centering\textbf{Multi-Agent\\Methods}\vspace{5pt}} &
\parbox{1.1cm}{\vspace{5pt}\centering\textbf{World\\Models}\vspace{5pt}} &
\parbox{1.3cm}{\vspace{5pt}\centering\textbf{Scenario\\Generation}\vspace{5pt}} &
\parbox{1.3cm}{\vspace{5pt}\centering\textbf{Cognitive /\\Physics}\vspace{5pt}} &
\parbox{1.3cm}{\vspace{5pt}\centering\textbf{Sim. Tools\\\& Datasets}\vspace{5pt}} &
\parbox{1.1cm}{\vspace{5pt}\centering\textbf{Cross-\\Discipl.}\vspace{5pt}} \\
\hline
Schwarting et al.~\cite{schwarting2018planning} & -- & (\checkmark) & -- & -- & -- & -- & -- & -- \\
Kuutti et al.~\cite{kuutti2020survey} & -- & -- & -- & -- & -- & -- & \checkmark & -- \\
Grigorescu et al.~\cite{grigorescu2020survey} & -- & (\checkmark) & -- & -- & -- & -- & (\checkmark) & -- \\
Chao et al.~\cite{chao2020visual} & (\checkmark) & \checkmark & (\checkmark) & -- & -- & -- & \checkmark & (\checkmark) \\
Di and Shi~\cite{di2021mixed} & \checkmark & -- & (\checkmark) & (\checkmark) & -- & (\checkmark) & -- & \checkmark \\
Kiran et al.~\cite{kiran2021drl_survey} & -- & -- & -- & -- & -- & -- & (\checkmark) & -- \\
Farah et al.~\cite{farah2022modeling} & \checkmark & -- & -- & -- & -- & -- & \checkmark & (\checkmark) \\
Ding et al.~\cite{ding2023scenario} & -- & -- & (\checkmark) & -- & \checkmark & -- & (\checkmark) & -- \\
Li et al.~\cite{li2024choose} & -- & -- & -- & -- & -- & -- & \checkmark & -- \\
Chen et al.~\cite{chen2024e2e_survey} & -- & (\checkmark) & -- & (\checkmark) & -- & -- & (\checkmark) & -- \\
Chen et al.~\cite{chen2024datadriven} & -- & \checkmark & (\checkmark) & -- & -- & -- & \checkmark & -- \\
\hline
\textbf{This survey} & \checkmark & \checkmark & \checkmark & \checkmark & \checkmark & \checkmark & \checkmark & \checkmark \\
\hline
\end{tabular}
\end{table*}

\section{SINGLE AGENT METHODS}
\label{sec:single_agent}

Single-agent methods focus on modeling the behavior of an individual vehicle. These methods treat other agents as part of the environment's dynamics, assuming their behaviors are fixed, predefined, or modeled separately. In mixed autonomy traffic simulation, single-agent methods serve two primary purposes. First, they develop and train driving policies for the AV itself, enabling decision-making and trajectory planning in response to surrounding traffic. Second, they model individual human drivers or other road users; these individual models are then replicated across multiple agents to produce the surrounding traffic that populates the simulated environment.

While this simplification may not capture full multi-agent interaction complexity, single-agent approaches offer practical advantages: computational efficiency, simpler training pipelines, and, importantly, closer alignment with real-world operation in mixed autonomy. In practice, an automated vehicle under test (often called the ego vehicle) does not have direct access to other road users’ internal states, such as their intentions, goals, or planned maneuvers; it must instead act based on partial observations and uncertain inferences from motion cues, context, and traffic rules. Single-agent formulations naturally match this information structure by learning policies conditioned on observable history and scene context, without assuming access to other agents’ latent intent. Also, when deployed across multiple instances, each controlled by an independent single-agent model, these methods create diverse traffic scenarios without explicitly modeling inter-agent coordination.

In the following sub-sections, we review single-agent methods and summarize their main assumptions, modeling choices, and key characteristics. A comparative summary of these methods is provided in Table~\ref{tab:single_agent_summary}.

\begin{table*}[t]
\centering
\caption{Summary of single-agent behavior modeling methods for mixed autonomy traffic simulation.}
\label{tab:single_agent_summary}
\renewcommand{\arraystretch}{1.25}
\setlength{\tabcolsep}{3.5pt}
% \footnotesize
\scriptsize
\begin{tabularx}{\textwidth}{
    >{\RaggedRight\arraybackslash}p{1.9cm}
    >{\RaggedRight\arraybackslash}p{1.7cm}
    >{\centering\arraybackslash}p{0.8cm}
    >{\RaggedRight\arraybackslash}X
    >{\RaggedRight\arraybackslash}X
    >{\RaggedRight\arraybackslash}p{3.1cm}
}
\toprule
\textbf{Method} &
\textbf{Training Signal} &
\textbf{Loop} &
\textbf{Strengths} &
\textbf{Limitations} &
\textbf{Representative Studies} \\
\midrule

Imitation Learning / Behavior Cloning &
Expert demonstrations (supervised) &
OL, CL &
Simple and scalable; direct data-to-policy mapping; strong baselines with large datasets &
Covariate shift and compounding errors; causal confusion; limited robustness to out-of-distribution states &
ALVINN~\cite{pomerleau1989alvinn}, PilotNet~\cite{bojarski2016end}, CIL~\cite{codevilla2018end}, ChauffeurNet~\cite{bansal2019chauffeurnet}, DAgger~\cite{ross2011dagger}, TransFuser~\cite{prakash2021multi}, Urban Driver~\cite{scheel2022urbandriver} \\

Trajectory Prediction &
Logged trajectories (supervised) &
OL &
Multi-modal forecasting; mature standardized benchmarks; captures distributional structure of future motion &
Marginal prediction; non-reactive; displacement metrics miss interactive quality &
CoverNet~\cite{phan2019covernet}, MultiPath~\cite{chai2019multipath}, TNT~\cite{zhao2021tnt}, DenseTNT~\cite{gu2021densetnt}, LaPred~\cite{kim2021lapred}, PRIME~\cite{song2022prime}, MID~\cite{gu2022stochastic} \\

Reinforcement Learning &
Reward signal (trial-and-error) &
CL &
Discovers novel strategies via exploration; no expert demonstrations needed; inherently reactive &
Sample inefficiency; reward engineering difficulty; safety during exploration; sim-to-real gap &
Kendall et al.~\cite{kendall2019learning}, Roach~\cite{zhang2021roach}, BC-SAC~\cite{bronstein2022imitation}, Think2Drive~\cite{li2024think2drive}, HAIM-DRL~\cite{wu2024haim}, Safe RL~\cite{shalev2016safe} \\

IRL / Adversarial Imitation &
Expert demos (adversarial matching) &
CL &
Recovers interpretable reward functions; robust to compounding errors; captures driving preferences &
Training instability; mode collapse; sensitivity to discriminator design; scaling difficulty &
MaxEnt IRL~\cite{ziebart2008maximum}, GAIL~\cite{ho2016generative}, Kuefler et al.~\cite{kuefler2017imitating}, PS-GAIL~\cite{bhattacharyya2018multi}, AIRL~\cite{fu2017learning}, RAIL~\cite{bhattacharyya2019simulating}\\

End-to-End Learned Planners &
Expert demos + perception labels &
OL, CL &
Joint perception--planning; end-to-end gradient flow; unified architecture &
Ego-status bias; open-loop metric gap; limited closed-loop generalization &
NMP~\cite{zeng2019end}, UniAD~\cite{hu2023uniad}, VAD~\cite{jiang2023vad}, SparseDrive~\cite{sun2024sparsedrive}, InterFuser~\cite{shao2023safety}, ThinkTwice~\cite{jia2023thinktwice}, NAVSIM~\cite{dauner2024navsim} \\

Model-Based Policy Learning &
Learned world model + reward &
CL &
Sample-efficient via imagination; decouples dynamics learning from policy optimization &
Model bias; compounding errors in imagined rollouts; sensitive to model accuracy &
World Models~\cite{ha2018worldmodels}, Dreamer~\cite{hafner2019dreamer}, DreamerV3~\cite{hafner2023dreamerv3}, MILE~\cite{hu2022mile}, AdaWM~\cite{wang2025adawm}, Imagine-2-Drive~\cite{garg2024imagine2drive} \\

Foundation Model Approaches &
Pretrained LLM/VLM + prompts &
OL, CL &
Commonsense reasoning; interpretable chain-of-thought; instruction-following capabilities &
Latency bottleneck; grounding gap; hallucination risk; impractical for multi-agent at scale &
Agent-Driver~\cite{mao2023agentdriver}, DiLu~\cite{wen2023dilu}, LanguageMPC~\cite{sha2023languagempc}, GPT-Driver~\cite{mao2023gptdriver}, DriveGPT4~\cite{xu2024drivegpt4} \\

\bottomrule
\end{tabularx}
\par\vspace{3pt}
{\scriptsize \textit{Loop}: OL = open-loop, CL = closed-loop. Methods listed under both can operate in either mode depending on deployment context.}
\end{table*}

\subsection{Imitation Learning}
\label{sec:il}

Imitation learning (IL) methods enable agents to directly learn driving behavior by observing expert
demonstrations from real-world datasets. The
simplest and most widely used form, behavior cloning (BC), frames the problem as
supervised learning: given a dataset
$\mathcal{D} = \{(s_i, a_i)\}_{i=1}^{N}$ of $N$ state-action pairs
collected from expert demonstrations, BC learns a policy $\pi_\theta$ parameterized by
$\theta$ by minimizing the empirical loss:
\begin{equation}
    \mathcal{L}(\theta) = \frac{1}{N} \sum_{i=1}^{N}
        \ell\!\left(\pi_\theta(s_i),\; a_i\right)
\end{equation}
where $\ell(\cdot,\cdot)$ is typically mean squared error for continuous actions or
cross-entropy for discrete actions. In driving contexts, the state $s_i$ commonly includes
sensor inputs or bird's-eye view representations
along with vehicle state and contextual data. The action $a_i$ comprises control
commands (steering, throttle, brake) or planned trajectory waypoints. Imitation learning
can operate in both open-loop and closed-loop settings: in open-loop mode, a trained
policy generates a \textit{fixed} trajectory from an initial observation without environmental
feedback; in closed-loop mode, the policy is queried at each simulation step, receiving
updated observations and producing reactive behavior. For mixed autonomy simulation,
IL enables driver models that replicate behavioral patterns from naturalistic
driving datasets, and when deployed across multiple independent instances, these models
can populate simulated environments with diverse, human-like traffic participants.

Imitation learning for automated driving began with ALVINN~\cite{pomerleau1989alvinn}, which showed that neural networks could map raw camera images directly to steering commands. PilotNet~\cite{bojarski2016end} scaled this idea with deep CNNs trained on large datasets and demonstrated that end-to-end policies could generalize across diverse road conditions. A key limitation of these reactive approaches, that is the inability to reason about goals, was addressed by Conditional Imitation Learning (CIL)~\cite{codevilla2018end}. CIL conditions the policy on high-level navigation commands through a branched architecture and enables goal-directed behavior. ChauffeurNet~\cite{bansal2019chauffeurnet} moved away from raw sensor input to mid-level bird's-eye view representations and introduced trajectory perturbation as a form of data augmentation to reduce distributional shift. ``Learning by Cheating''~\cite{chen2020learning} proposed a two-stage, teacher-student training paradigm in which a privileged agent with access to ground-truth environment information supervises a purely vision-based sensorimotor agent. Their framework substantially outperforming prior methods on the CARLA \cite{dosovitskiy2017carla} and NoCrash benchmarks. TransFuser~\cite{prakash2021multi} then brought multi-modal perception into this framework by fusing camera and LiDAR features through transformer attention at multiple resolutions.

A fundamental limitation of standard behavior cloning is \textit{covariate shift}. The policy is trained on expert-induced state distributions, whereas during deployment the agent may encounter states that differ from those seen in the training data, either due to differences in the initial state distribution or deviations caused by the learned policy itself. This mismatch can lead to compounding errors.

Several methods
address this by incorporating closed-loop feedback during training. Dataset Aggregation
(DAgger)~\cite{ross2011dagger} alternates between rolling out the learner's current policy
and querying an expert for corrective labels in the visited states. DART~\cite{laskey2017dart} injects structured noise into expert
demonstrations during collection, HG-DAgger~\cite{kelly2019hgd} improves scalability by learning an intervention rule that requests expert control only in states predicted to be risky, and MEGA-DAgger~\cite{sun2023megadagger} extends DAgger to settings with multiple imperfect experts. Beyond interactive correction,
Urban Driver~\cite{scheel2022urbandriver} presents an offline approach that builds a
differentiable data-driven simulator from perception outputs and HD maps.

Despite its simplicity and scalability, imitation learning faces several persistent challenges with direct implications for simulation quality. \textit{Distributional mismatch} can cause BC-trained agents behave unrealistically when the simulation diverges from recorded scenarios, limiting their utility as reactive traffic participants~\cite{codevilla2019exploring}. This may arise from covariate shift, dataset bias, and causal confusion. Interactive correction methods such as DAgger~\cite{ross2011dagger} mitigate this but require repeated expert queries in simulation, which is expensive and difficult to scale for high-dimensional observations and long-horizon tasks. \textit{Data coverage limitations} compound this problem: expert data quality and consistency are critical but is typically unavailable for rare edge cases, which are precisely the scenarios most relevant for AV safety testing. Together, these challenges mean that IL-based agents may match logged driving distributions while failing to respond plausibly to interventions, which is the core requirement for closed-loop simulation. Future directions include uncertainty quantification for detecting out-of-distribution operation, retrieval-based methods that ground predictions in similar recorded situations, and hybrid approaches that use IL as initialization for reinforcement or adversarial training. The next section discusses trajectory prediction, a related family that shares the supervised training signal of behavior cloning but differs in deployment role and the core problem it addresses.

\subsection{Trajectory Prediction}
\label{sec:traj_pred}

Trajectory prediction forecasts future positions of agents given observed motion history and environmental context. Although both trajectory prediction and imitation learning learn from recorded expert data and can produce future waypoints as output, they differ in role and training objective. An imitation learning policy is trained as an \textit{actor} that produces actions for an agent to execute. A trajectory prediction model is trained as a \textit{passive observer} that forecasts where other agents will be; because it does not act, it does not compound its own errors across time steps in the same way. When a trajectory prediction model is deployed in closed-loop by re-querying it at each time step, it is operationally equivalent to an imitation learning policy, but without the training-time mechanisms, such as DAgger or adversarial correction, that handle the resulting distributional mismatch. Single-agent trajectory prediction models learn a conditional distribution over future trajectories: $P(Y \mid X, C)$, where $X = \{x_1, x_2, \ldots, x_{T_h}\}$ represents the observed trajectory over history horizon $T_h$, $C$ encodes contextual information, and $Y = \{y_1, y_2, \ldots, y_{T_f}\}$ denotes the predicted future trajectory over forecast horizon $T_f$. These models are predominantly evaluated in open-loop settings, where predicted trajectories are compared against ground-truth recordings using displacement metrics.

Foundational approaches established key paradigms for encoding agent history, scene
context, and multi-modal output. CoverNet~\cite{phan2019covernet} treats prediction
as classification over a fixed set of trajectory anchors derived from expert data. An important aspect is multimodality: at any decision
point, an agent may turn left, go straight, or turn right, each representing a valid
future. 
MultiPath~\cite{chai2019multipath} extended CoverNet with learnable anchor trajectories
corresponding to distinct behavioral modes to enable efficient multi-modal predictions. Goal-conditioned approaches improve accuracy for long-horizon predictions by decomposing the task into two sequential stages: the model first predicts a plausible goal location, such as a target position or waypoint, and then generates a trajectory conditioned on that endpoint. This two-stage structure anchors the trajectory to a specific destination, which constrains the prediction space and reduces the uncertainty that accumulates in free rollouts over long time horizons. TNT~\cite{zhao2021tnt} and DenseTNT~\cite{gu2021densetnt} predict target goal locations or probability distributions from lane
centerlines and generate trajectories conditioned on selected endpoints. 
Similarly, LaPred~\cite{kim2021lapred} leveraged lane-aware representations to structure predictions
in highway and urban environments.

Recently, generative models offer alternative
approaches to capturing trajectory multimodality.
PRIME~\cite{song2022prime} improved robustness under imperfect tracking by constraining predictions to
dynamically feasible trajectories. MID~\cite{gu2022stochastic} formulated the trajectory prediction task as a reverse process of motion
indeterminacy diffusion that progressively refines noisy samples into determinate
trajectories. These generative approaches capture richer distributional structure
than deterministic or anchor-based methods, but typically at higher computational cost.

Despite significant progress, \textit{single-agent} trajectory prediction faces fundamental
limitations for simulation applications. The independence assumption among agents could potentially generate inconsistent or colliding trajectories. Rare events and safety-critical scenarios remain challenging due to
underrepresentation in training data. As a result, models also show limited generalization across geographic regions and
driving cultures. These limitations motivate joint multi-agent prediction
methods that explicitly model
inter-agent dependencies, and closed-loop agent models that adapt to evolving simulation
states.

\subsection{Reinforcement Learning}
\label{sec:rl}

Reinforcement learning (RL) provides a framework for learning driving policies through
trial-and-error interaction with the environment, guided by reward signals rather than
relying on labeled data. In RL, an agent observes state $s_t$ at time step $t$, executes action
$a_t$ according to policy $\pi_\theta(a_t|s_t)$ parameterized by $\theta$, receives scalar
reward $r_t$, and transitions to new state $s_{t+1}$. The objective is to maximize
expected cumulative discounted reward:
\begin{equation}
    J(\theta) = \mathbb{E}_{\pi_\theta}\left[\sum_{t=0}^{T} \gamma^t r_t\right]
\end{equation}
where $\gamma \in [0,1]$ is the discount factor and the expectation is taken over
trajectories generated by policy $\pi_\theta$. For automated driving, reward functions encode
objectives such as progress toward goal, lane keeping, collision avoidance, and passenger
comfort. Unlike imitation learning, which requires expert demonstrations, RL discovers policies through exploration. This is a strength enabling discovery of novel strategies but also a drawback as it requires extensive interaction with the environment to learn a robust policy. RL methods are inherently closed-loop because the agent continuously receives environmental feedback and adapts its actions.

Foundational work demonstrated RL viability for increasingly complex driving tasks.
Kendall et al.~\cite{kendall2019learning} achieved the first successful deep RL
application to real-world driving by training DDPG agents for lane following using monocular
camera input. Toromanoff et al.~\cite{toromanoff2020end} introduced implicit affordances
combined with Rainbow-IQN-Apex, achieving the first RL agent capable of end-to-end urban
driving including traffic light detection. Zhang et al.~\cite{zhang2021roach} introduced
Roach, an RL expert trained using PPO that serves as a superior ``coach'' for imitation
learning. Waymo's
BC-SAC~\cite{bronstein2022imitation} combined behavior cloning initialization with Soft
Actor-Critic on over 100,000 miles of real-world data and showed that RL reward signals
significantly improve robustness in challenging scenarios. More recently, Li et
al.~\cite{li2024think2drive} proposed Think2Drive, a model-based RL approach using latent
world models, which illustrates the potential of imagination-based policy learning for driving. Safety-aware
formulations have also emerged as viable paradigms. Shalev-Shwartz et
al.~\cite{shalev2016safe} introduced hard safety constraints into RL driving policies, and
Wen et al.~\cite{wen2020safe} developed Parallel Constrained Policy Optimization to
enforce safety constraints during training. Recently, human-in-the-loop approaches offer complementary
efficiency gains. HAIM-DRL~\cite{wu2024haim} treats humans as AI mentors providing sparse
interventions during training to guide the exploration toward safer and more efficient policies.

Sample
efficiency in RL is a primary concern. RL training requires millions of environment interactions with
heavy simulation reliance and associated sim-to-real gap issues. Reward engineering is also
non-trivial; simple rewards may lead to unintended behaviors while complex multi-objective rewards are hard to tune and may produce brittle
policies. Generalization to unseen environments is also a challenge similar to other learning-based methods. For mixed autonomy simulation specifically, current RL methods focus primarily on
ego-vehicle control rather than generating diverse traffic agent behaviors. Therefore, future directions could focus on sample-efficient algorithms through world models and offline RL, improved sim-to-real
transfer, safe exploration guarantees, and generative diverse and human-like behaviors.

\subsection{Inverse Reinforcement Learning and Adversarial Imitation Learning}
\label{sec:irl_gail}

Inverse reinforcement learning (IRL) infers reward functions from expert demonstrations by
assuming that observed behavior is approximately optimal with respect to unknown rewards.
Unlike behavior cloning, which directly maps states to actions, IRL recovers the underlying objectives driving expert behavior, and therefore, enables generalization to novel states by
optimizing the recovered rewards. Also, unlike standard RL, which requires manually specifying
reward functions, IRL extracts these objectives from data, which is a significant advantage for
driving, where human preferences involve complex trade-offs
between speed, comfort, safety, and social norms. Mathematically, given expert demonstrations
$\mathcal{D}_E = \{(s_i, a_i)\}_{i=1}^{N}$, IRL seeks to find reward function $r_\phi$
parameterized by $\phi$ such that the expert policy outperforms alternatives:
\begin{equation}
    \phi^* = \arg\max_{\phi} \left[ \mathbb{E}_{\pi_E}[r_\phi] - \max_{\pi} \mathbb{E}_{\pi}[r_\phi] \right]
\end{equation}
where $\pi_E$ denotes the expert policy and the expectation is over trajectories. However, this
formulation is inherently ill-posed, which means that many reward functions can explain the same behavior.
Maximum Entropy IRL~\cite{ziebart2008maximum} resolves this ambiguity by selecting reward
functions that make the expert demonstrations most likely under maximum entropy trajectory
distributions.

Adversarial imitation learning methods reframe the imitation problem as a two-player game: a policy generates behavior while a discriminator tries to distinguish it from expert demonstrations. Generative Adversarial Imitation Learning
GAIL~\cite{ho2016generative} bypasses explicit reward recovery by directly matching the learner's state-action occupancy measure to that of the expert. Instead, it directly trains the policy to produce the same distribution of (state, action) pairs that the expert visits. The discriminator is what drives this. it learns to tell the policy's behavior apart from the expert's, and the policy is trained to fool it, which pushes the two distributions closer together. The GAIL objective is:
\begin{equation}
\begin{aligned}
    \min_{\pi_\theta} \max_{D_\psi}\;& \mathbb{E}_{\pi_\theta}\left[\log D_\psi(s, a)\right] \\
    &+ \mathbb{E}_{\pi_E}\left[\log(1-D_\psi(s, a))\right]
    - \lambda \mathcal{H}(\pi_\theta)
\end{aligned}
\end{equation}
where $D_\psi$ is a discriminator parameterized by $\psi$ distinguishing expert from
learner state-action pairs, and $\mathcal{H}(\pi_\theta)$ denotes the entropy of policy
$\pi_\theta$, weighted by $\lambda > 0$. Adversarial Inverse Reinforcement Learning
(AIRL)~\cite{fu2017learning} extends this framework by structuring the discriminator to
recover a disentangled reward function that transfers across environments with different
dynamics. For mixed autonomy simulation, these methods are particularly valuable as they
learn interpretable reward functions capturing human driving preferences and enable generation of realistic and diverse traffic
agent behaviors without requiring hand-crafted reward engineering.

IRL and adversarial IL methods are inherently closed-loop during training, as the learner
must interact with the environment to generate trajectories for discriminator comparison.
Once a reward function is recovered, it can be used to train new policies via standard RL
in closed-loop settings, or the GAIL-trained policy itself can serve as a closed-loop
simulation agent. However, recovered reward functions can also be applied in open-loop
contexts, for instance, to score or rank pre-generated trajectory candidates without
environmental feedback.

Kuefler et al.~\cite{kuefler2017imitating} first applied GAIL to highway driving using
NGSIM data and demonstrated that adversarial imitation produces more realistic lane-keeping
and car-following behavior than behavior cloning, particularly over longer rollout
horizons where BC suffers from compounding errors. PS-GAIL~\cite{bhattacharyya2018multi} extended this to multi-agent settings via shared policy networks and curriculum learning. Bhattacharyya et al.~\cite{bhattacharyya2019simulating} further augmented GAIL with semantic rewards (RAIL) to disentangle latent driving style factors for controllable behavior generation. Huang et al.~\cite{huang2021driving} modeled driving intentions as discrete latent variables to enable learning personalized rewards that generalize to unseen conditions. Wang et al.~\cite{wang2020adversarial} and Sackmann et al.~\cite{sackmann2022modeling} both applied AIRL to highway scenarios, the former augmenting with semantic rewards for stability, the latter recovering interpretable reward functions capturing diverse driving styles.

For simulation applications, especially in the context of mixed autonomy traffic, IRL and adversarial IL face challenges worth noting. Training instability remains significant, particularly in multi-agent settings where discriminators must distinguish expert behavior while each agent's behavior is highly dependent on surrounding vehicles, which makes the discriminator's feedback noisy and unreliable. Multi-agent extensions that explicitly address discriminator instability arising from irrelevant agent interactions are discussed in Section~\ref{subsec:decentralized_methods}. Moreover, recovered
rewards may not transfer across different road geometries or traffic densities, which limits
generalization. For multi-agent simulation, scaling remains difficult. Training
independent GAIL agents leads to non-stationary dynamics as each agent's learning changes
the environment for others, and centralized approaches can become computationally
expensive and disregard the decentralized nature of mixed autonomy traffic. Future directions may include more stable multi-agent adversarial training,
combining IRL with offline RL to reduce environment interaction requirements, and learning
hierarchical reward structures that separate strategic intent from tactical control. Also,
integrating safety constraints into the reward recovery process seems a promising direction.

\subsection{End-to-End Learned Planners}
\label{sec:end_to_end}

End-to-end learned planners represent a major paradigm within single-agent AV behavior modeling, directly mapping sensor observations to planned trajectories through unified differentiable architectures. By jointly learning perception, prediction, and planning, these systems produce driving policies that can serve as the automated ego vehicle in mixed autonomy simulation, as alternative to rule-based autopilots in tools like CARLA or SUMO. While end-to-end planners are often trained using imitation learning objectives, they are treated as a distinct section in this survey because they are defined by their architectural scope and can be trained with IL, RL, or hybrid objectives. However, IL is defined by its learning signal, that is mimicking expert behavior, and applies regardless of architecture. End-to-end planners learn a mapping: \begin{equation} \pi_\theta: (I, M, E) \rightarrow \tau \end{equation} where $I$ represents sensor inputs, $M$ denotes map information, $E$ captures ego vehicle state, and $\tau = \{y_1, y_2, \ldots, y_{T_p}\}$ is the planned trajectory over planning horizon $T_p$. The policy $\pi_\theta$ is parameterized by neural network weights $\theta$. These models can be evaluated in either open-loop mode or closed-loop mode.

NMP~\cite{zeng2019end} established the foundational idea that planning quality depends on rich intermediate scene representations, encoding LiDAR and HD maps into learned cost volumes that score candidate trajectories. TransFuser~\cite{prakash2021multi, chitta2022transfuser} and InterFuser~\cite{shao2023safety} demonstrated that multi-modal transformer fusion across camera and LiDAR streams produces interpretable, safety-aware AV behavior. ST-P3~\cite{hu2022stp3} introduced spatial-temporal feature learning that tightly couples scene understanding with trajectory generation.  A pivotal
contribution, UniAD~\cite{hu2023uniad}, introduced a
planning-oriented philosophy by hierarchically organizing tracking, mapping, motion
forecasting, and occupancy prediction to facilitate planning. It established the paradigm
that intermediate task supervision improves planning quality. VAD~\cite{jiang2023vad} and SparseDrive~\cite{sun2024sparsedrive} demonstrated that efficient vectorized and sparse representations can sustain high behavioral quality, a property that matters for simulation deployments where computational budget is shared across many concurrently running agents. ThinkTwice~\cite{jia2023thinktwice} addressed the refinement problem by conditioning trajectory generation on imagined future scenes and produced more consistent and anticipatory AV behavior. PP\&P~\cite{sadat2020perceive} contributed the complementary insight that inspectable intermediate representations enable verification of what the agent perceives before its behavioral decisions are executed. This property is particularly important for simulation-based AV testing workflows. 

More recently, diffusion-based architectures have been applied directly to the planning task by leveraging iterative denoising to model multi-modal driving behavior distributions more expressively than regression-based decoders. Diffusion Planner~\cite{zheng2025diffusionplanner} jointly models ego planning and neighboring vehicle prediction within a single diffusion transformer, using classifier guidance to adapt driving style (safety, comfort, speed) without retraining. DiffusionDrive~\cite{liao2025diffusiondrive} addresses the computational bottleneck of iterative denoising through truncated diffusion with anchored Gaussian priors, reducing the reverse process to two steps and enabling real-time inference.

End-to-end planners have shown impressive benchmark performance; however, recent studies have exposed fundamental
evaluation concerns. Zhai et al.~\cite{zhai2023rethinking} and Zhiqi et
al.~\cite{li2024ego} demonstrated that simple MLP-based models using only ego vehicle
state can achieve comparable L2 displacement error to sophisticated perception-based
methods on nuScenes. This suggests that standard open-loop metrics may not adequately
measure planning capability. This ``ego status bias'' arises because datasets are
dominated by straight driving scenarios where maintaining current velocity is nearly
optimal. NAVSIM~\cite{dauner2024navsim} proposed a middle ground through non-reactive
simulation with safety-aware metrics (progress, time-to-collision), finding that
simpler methods can match complex architectures on challenging scenarios. In other words, the gap between open-loop metrics and closed-loop performance means models with
low displacement error may still produce unsafe behavior when deployed reactively. For
simulation applications, end-to-end planners provide architectural insights for building
traffic agents but require closed-loop training mechanisms to serve as truly reactive
simulation participants.

\subsection{Model-Based Policy Learning}
\label{sec:single_model_based_policy_learning}

Model-based policy learning trains driving policies by leveraging learned dynamics models, also known as ``world models'', to predict how the environment evolves under candidate actions, rather
than requiring direct interaction with the environment for every policy update. A common
formulation introduces latent state $z_t$ summarizing observation history, using encoder
$e_\theta$ mapping observations to latent states and latent transition model $p_\theta$
modeling dynamics in latent space:
\begin{equation}
    z_t = e_\theta(o_{\le t}), \quad z_{t+1} \sim p_\theta(z_{t+1}\mid z_t, a_t)
\end{equation}
where $o_{\le t} = \{o_1, o_2, \ldots, o_t\}$ denotes the observation history up to time
$t$, and $a_t$ is the action taken at time $t$. The ego policy
$\pi_\phi(a_t\mid z_t)$ parameterized by $\phi$ is optimized by rolling out imagined
trajectories in latent space and maximizing expected return:
\begin{equation}
    J(\phi) = \mathbb{E}_{z_t, \{a_t, z_{t+1}\}}\left[\sum_{k=0}^{H} \gamma^k r(z_{t+k}, a_{t+k})\right]
\end{equation}
where $H$ is the imagination horizon, $r(z_t, a_t)$ is a reward function operating on
latent states, and the expectation is over latent rollouts generated by the learned
dynamics model and policy. This separates learning predictive models of the world from
learning policies through imagination within those models. Compared to model-free RL,
which requires direct environment interaction for every gradient step, policy improvement
here happens through simulated rollouts inside learned models, which substantially improves
sample efficiency but makes performance sensitive to model accuracy and bias.

Model-based approaches are fundamentally closed-loop methods. The learned world model
provides the feedback loop during imagination-based training, and the resulting policy is
deployed as a reactive agent that conditions actions on current observations. However, the
world model itself can also be used in open-loop fashion, for instance, to generate
multiple candidate future trajectories under different action sequences for offline
scoring or trajectory ranking without environmental feedback. 
The key distinction from the generative world models surveyed in 
Section~\ref{sec:worldmodels_sim} is functional rather than architectural: 
here the world model serves as a \emph{training substrate} for a specific 
ego policy, whereas 
Section~\ref{sec:worldmodels_sim} covers world models deployed as 
\emph{reusable simulation infrastructure} serving multiple downstream 
applications including policy evaluation, synthetic data generation, and 
counterfactual testing. Some model families, such as latent dynamics models and  
autoregressive token predictors, could appear in both contexts.

Ha and Schmidhuber~\cite{ha2018worldmodels} introduced the foundational paradigm of
training recurrent generative world models and optimizing compact controllers using
extracted latent features. The Dreamer line of work established scalable recipes for
jointly learning latent dynamics and improving behavior through differentiable
imagination. Dreamer~\cite{hafner2019dreamer} learns a recurrent state-space model (RSSM)
and optimizes actors using value gradients backpropagated through imagined trajectories.
DreamerV2~\cite{hafner2021dreamerv2} strengthened this approach with discrete latent
representations to improve model expressiveness.
DreamerV3~\cite{hafner2023dreamerv3} demonstrated robustness across diverse domains with a
single hyperparameter configuration and established world-model-based RL as a general-purpose
approach. For automated driving specifically, MILE~\cite{hu2022mile} adapts
imagination-based learning to urban driving by jointly learning compact latent world models
and ego policies from offline demonstrations. AdaWM~\cite{wang2025adawm} targets the distribution mismatch
between pretrained dynamics models and planning policies and proposed adaptive finetuning
strategies that improve policy performance without retraining the world model from scratch.
Imagine-2-Drive~\cite{garg2024imagine2drive} pairs high-fidelity world models with diffusion-based policy actors and leverage the multi-modal generation capability of
diffusion models to represent diverse action modes within the imagined rollouts.

World models and model-based approaches are becoming increasingly popular, but they also face challenges. Model bias and compounding errors, where policies can learn to exploit
inaccuracies in the learned dynamics can produce actions that appear optimal in imagination
but fail in the real environment. Learning dynamics that remain valid under interventions
is particularly difficult simply because the model has not observed
the consequences of actions outside the data distribution. Multi-modality in traffic
evolution requires uncertainty-aware rollouts and robust planning that does not rely only
to a single imagined future. Future directions include stronger uncertainty calibration to
detect when imagination diverges from reality, hybrid training that mixes offline world
model priors with online correction through limited real interaction, and explicit safety
constraints that remain stable under distribution shift between imagined and real dynamics.

\subsection{Foundation Model Approaches}
\label{sec:single_foundation_models}

Foundation model approaches leverage large pretrained models, such as large language
models (LLMs) or vision-language models (VLMs), as high-level decision-making or
reasoning modules for ego vehicles. The key idea is that pretraining on
internet-scale data provides these models with commonsense reasoning, semantic
understanding, and instruction-following capabilities that are difficult to acquire from
driving data alone.  
% It is important to distinguish two related but architecturally different paradigms. The first uses pretrained LLMs or VLMs with prompting and optional finetuning and rely on knowledge and reasoning capabilities acquired during pretraining. The second
% trains GPT-style autoregressive architectures on driving-specific data and borrow the
% transformer backbone and next-token prediction objective but not the pretrained weights or
% broad language understanding. Here, we focus primarily on the former, as the latter is more closely related to the token-based methods discussed in
% Section~\ref{subsec:multi_closed_loop_centralized}.

A generic formulation treats the foundation model as a conditional
decision function over structured scene representations:
\begin{equation}
    y_t = f_{\theta}\big(\mathcal{P}(o_{\le t}, m, g), \mathcal{M}_t\big)
\end{equation}
where $o_{\le t}$ denotes the observation history, $m$ represents the map context, $g$
encodes navigation goals or mission specifications, $\mathcal{P}(\cdot)$ is a prompt
builder that structures these inputs into a format suitable for the foundation model,
$\mathcal{M}_t$ denotes memory or retrieved knowledge at time $t$, and $f_\theta$ is the
pretrained foundation model parameterized by $\theta$. The output $y_t$ can take several
forms: discrete maneuver decisions (e.g., ``change lane left,'' ``yield to pedestrian''),
constraints or cost weights for downstream optimizers, or draft trajectory waypoints.
% Critically, $y_t$ is typically not used as direct vehicle control but is translated into
% continuous actions through conventional controllers:
% \begin{equation}
%     a_t = \mathcal{C}(y_t, s_t)
% \end{equation}
% where $\mathcal{C}$ is a controller (e.g., MPC, PID) that maps high-level decisions to
% control actions $a_t$ given current state $s_t$. This separation assigns foundation models
% responsibility for high-level semantic reasoning while low-level controllers enforce
% vehicle dynamics and safety constraints---a design choice that mitigates the risk of
% physically implausible outputs from models not trained on dynamics.
Foundation model approaches can operate in both open-loop and closed-loop settings.
% In open-loop mode, the model generates a complete plan or sequence of decisions from an
% initial scene description without subsequent environmental feedback. In closed-loop mode,
% the model is queried at each simulation step with updated observations, enabling reactive
% behavior. 
However, the computational cost of LLM inference poses a significant bottleneck for real-time
closed-loop deployment, and most current systems operate at reduced frequency or with
asynchronous decision-making.

This class of models has become very popular in recent years. Agent-Driver~\cite{mao2023agentdriver} proposes an LLM-centered driving agent equipped
with tool libraries (for perception and prediction queries), memory modules storing past
experiences, and explicit chain-of-thought reasoning steps to yield interpretable
decisions. DiLu~\cite{wen2023dilu} develops a
knowledge-driven framework by combining LLMs with explicit Reasoning and Reflection modules. The Reasoning module generates decisions from scene descriptions, and the Reflection
module evaluates outcomes and accumulates experience memory. This enables closed-loop
improvement through self-correction. LanguageMPC~\cite{sha2023languagempc} uses LLMs as
high-level decision-makers that output situation-dependent parameters, such as target speed and desired lane, mapped into actionable commands through Model Predictive Control; effectively
using LLMs to select or adapt control objectives while preserving the optimization
structure and constraint satisfaction guarantees of MPC.

Beyond high-level reasoning, recent work explores whether LLMs can directly generate
trajectories or control signals. GPT-Driver~\cite{mao2023gptdriver} reformulates motion
planning as a language modeling problem, representing planner inputs and outputs as
language tokens and leveraging GPT-3.5 to generate driving trajectories through coordinate
descriptions. 
DriveGPT4~\cite{xu2024drivegpt4} extends this paradigm to interpretable end-to-end
driving. It processes multi-frame video inputs through a multimodal LLM to predict low-level
vehicle control signals while providing natural language explanations for vehicle actions. 
% Complementing language-based approaches,
% BEVGPT~\cite{wang2024bevgpt} proposes a generative pre-trained architecture that
% integrates prediction, decision-making, and motion planning using bird's-eye-view images
% as sole input. While BEVGPT adopts the GPT autoregressive framework, it is trained from
% scratch on driving data rather than leveraging pretrained language knowledge, positioning
% it closer to domain-specific autoregressive models than to foundation model transfer.

Reliability is a primary concern in foundation models as LLM outputs can be sensitive to prompt phrasing. Moreover, mapping
language-level reasoning into geometrically and dynamically valid actions remains
difficult as language representations do not natively encode spatial relationships with
the precision required for collision avoidance. Hallucination, where models generate
plausible-sounding but factually incorrect reasoning about scene elements, poses safety
risks that are difficult to detect without independent verification.
Latency constraints limit closed-loop deployment. Current evaluation is most limited to small scenario sets or with open-loop metrics. For mixed autonomy simulation, foundation models offer intriguing possibilities for generating diverse,
human-like reasoning patterns for traffic agents, but the computational cost of running
LLM inference for every agent at every timestep makes direct application to multi-agent
simulation impractical with current architectures. Future directions include uncertainty-aware decision interfaces that express confidence alongside decisions, formal safety wrappers that verify LLM outputs against physical constraints before execution, more efficient inference, and standardized closed-loop evaluation protocols measuring. 

\subsection{Discussions on Single-Agent Methods}
\label{subsec:single_challenges}

The fundamental limitation shared across many single-agent methods is the independence assumption. When a single-agent policy trained via imitation learning, reinforcement learning, or any other approach is deployed in a simulator where surrounding agents follow \textit{fixed replay or simple rule-based models}, the resulting traffic dynamics may diverge substantially from what would occur in the real world. The ego policy is effectively optimized against an environment whose transition dynamics $p_{\text{train}}$ do not match the true interactive dynamics $p_{\text{real}}$, and this mismatch can remain hidden during development because open-loop evaluation metrics, which compare predicted trajectories against recorded ground truth, are insensitive to interactive effects. A model that achieves low displacement error may still produce unsafe or unrealistic behavior when other agents begin responding to its actions rather than following scripted trajectories. 

This method-agnostic challenge point toward several promising research directions. Tighter integration between ego-policy learning and surrounding traffic modeling represents the most natural bridge to the multi-agent methods discussed in Section~\ref{sec:multi_agent}. Hybrid architectures that combine the semantic reasoning capabilities of foundation models with the dynamics grounding of model-based approaches and the safety guarantees of constrained optimization could address complementary weaknesses. Evaluation methodology also requires attention: closed-loop benchmarks that measure not just trajectory accuracy but counterfactual robustness would provide far more meaningful assessments of simulation readiness than current displacement-based metrics. Finally, uncertainty quantification across all method families would enable conservative fallback behaviors when models operate outside their competence, a prerequisite for any deployment in safety-critical simulation infrastructure.

% While centralized neural simulators share architectural elements with the 
% generative world models of Section~\ref{sec:worldmodels_sim}---both learn 
% scene-level transition dynamics from data---the methods here are specifically 
% designed as \emph{agent behavior simulators} that model how traffic 
% participants act and interact, rather than as environment observation 
% generators. The distinction is in the output space: centralized simulators 
% produce agent states (positions, velocities, headings), while world models 
% produce observations (images, point clouds, occupancy grids) that may 
% implicitly encode but do not explicitly represent agent-level behavior.

% Closed loop evaluation metrics and training paradigms can address this issue to some extend, but ultimately resolving it requires surrounding agents that are themselves reactive and behaviorally realistic. This is one of the core motivations for the multi-agent methods reviewed in the following section.

\section{MULTI-AGENT METHODS}
\label{sec:multi_agent}

Multi-agent methods explicitly model multiple traffic participants and their interactions, capturing the ``social contract'' of driving: how agents anticipate, respond to, and coordinate with each other. Unlike the single-agent methods of Section~\ref{sec:single_agent}, which treat surrounding traffic as part of the environment, multi-agent approaches model the bidirectional influence between agents.

We organize multi-agent methods along two architectural axes. The first distinguishes \emph{joint trajectory forecasting} (Section~\ref{subsec:joint_forecasting}), which predicts coordinated futures for multiple agents. The second divides interactive simulation into \emph{centralized} approaches (Section~\ref{subsec:centralized_simulation}), where a single joint model produces globally consistent scene evolution, and \emph{decentralized} approaches (Section~\ref{subsec:decentralized_methods}), where independent per-agent policies yield emergent coordination. We further discuss the emerging role of foundation models for multi-agent coordination (Section~\ref{subsec:multi_agentic_foundation}). Table \ref{tab:multi_agent_summary} provides a summary of methods and studies reviewed in this section.

\begin{table*}[!htbp]
\centering
\caption{Summary of multi-agent interaction methods for mixed autonomy traffic simulation.}
\label{tab:multi_agent_summary}
\renewcommand{\arraystretch}{1.25}
\setlength{\tabcolsep}{3.5pt}
% \footnotesize
\scriptsize
\begin{tabularx}{\textwidth}{
    >{\RaggedRight\arraybackslash}p{1.8cm}
    >{\RaggedRight\arraybackslash}p{2.6cm}
    >{\RaggedRight\arraybackslash}X
    >{\RaggedRight\arraybackslash}X
    >{\RaggedRight\arraybackslash}p{3.3cm}
}
\toprule
\textbf{Method} &
\textbf{Core Idea} &
\textbf{Strengths} &
\textbf{Limitations} &
\textbf{Representative Studies} \\

%% --- SUB-HEADER: Joint Trajectory Forecasting ---
\midrule
\multicolumn{5}{l}{\textbf{\textit{Joint Trajectory Forecasting (predominantly open-loop)}}} \\
\midrule

Joint Trajectory Prediction &
Predicts joint future trajectories for multiple agents; uses graph networks, attention, or transformers to model inter-agent dependencies &
Captures inter-agent dependencies; multi-modal outputs; strong benchmark performance; joint consistency &
Does not model reactive feedback; joint consistency not guaranteed under marginal losses; gap to interactive simulation &
Social LSTM~\cite{alahi2016social}, Social GAN~\cite{gupta2018socialgan}, DESIRE~\cite{lee2017desire}, Trajectron++~\cite{salzmann2020trajectronpp}, PRECOG~\cite{rhinehart2019precog}, VectorNet~\cite{gao2020vectornet}, LaneGCN~\cite{liang2020lanegcn}, Scene Transformer~\cite{ngiam2022scenetransformer}, HiVT~\cite{zhou2022hivt}, MTR++~\cite{shi2023mtrpp}, GameFormer~\cite{huang2023gameformer} \\

%% --- SUB-HEADER: Centralized Neural Simulation ---
\midrule
\multicolumn{5}{l}{\textbf{\textit{Centralized Neural Simulation (closed-loop)}}} \\
\midrule

Diffusion-Based Simulators &
Models multi-agent evolution via iterative denoising; samples from learned joint trajectory distributions with guidance for controllability &
High diversity; multi-modal generation; controllable via guidance; collision-aware sampling &
Slow sampling (iterative denoising); hard constraint satisfaction difficult; closed-loop stability over long horizons &
MotionDiffuser~\cite{jiang2023motiondiffuser}, CTG~\cite{zhong2023ctg}, CTG++~\cite{zhong2023ctgpp} \\

Token-Based / Autoregressive &
Discretizes motion into tokens; casts simulation as conditional language modeling with next-token prediction &
Scalable transformer architectures; handles variable agent counts; real-time inference possible &
Discretization error; compounding token mistakes; serialization order asymmetry; constraint satisfaction requires constrained decoding &
MotionLM~\cite{seff2023motionlm}, Trajeglish~\cite{philion2024trajeglish}, BehaviorGPT~\cite{zhou2024behaviorgpt}, SMART~\cite{wu2024smart} \\

Regression-Based Reactive &
Single learned model produces globally consistent scene updates via continuous regression; rolls forward step-by-step &
Global scene consistency; captures multi-agent dependencies; computationally efficient; responds to ego deviations &
Long-horizon drift; limited multi-modality (point estimates); counterfactual validity concerns from passive logs &
SimNet~\cite{bergamini2021simnet}, TrafficSim~\cite{suo2021trafficsim} \\

%% --- SUB-HEADER: Decentralized Simulation ---
\midrule
\multicolumn{5}{l}{\textbf{\textit{Decentralized Simulation (closed-loop)}}} \\
\midrule

Multi-Agent RL / Self-Play &
Each agent learns a decentralized policy; CTDE paradigm for training; self-play creates automatic curriculum &
Emergent coordination; robustness through competition; automatic difficulty scaling &
Non-stationarity; credit assignment difficulty; reward specification; may converge to non-human equilibria &
Safe MARL~\cite{shalev2016safe}, Flow~\cite{wu2021flow}, Nocturne~\cite{vinitsky2022nocturne}, Data-Reg.\ Self-Play~\cite{cornelisse2024human}, Robust Self-Play~\cite{cusumanotowner2025robust_selfplay} \\

Multi-Agent Imitation / GAIL &
Decentralized policies trained to match expert interaction patterns via adversarial discrimination &
Reproduces human interaction styles; scalable via parameter sharing; captures emergent behaviors &
Training instability with global discriminators; diversity collapse; irrelevant interaction misguidance &
PS-GAIL~\cite{bhattacharyya2018multi}, RAIL/Burn-InfoGAIL~\cite{bhattacharyya2023titsgail}, Symphony~\cite{igl2022symphony}, DecompGAIL~\cite{guo2025decompgail} \\

Game-Theoretic Models &
Models interaction as strategic reasoning; Nash or Stackelberg equilibria capture anticipation and mutual influence &
Principled interaction modeling; captures strategic behavior; interpretable equilibrium concepts &
Utility specification difficulty; equilibrium selection ambiguity; computational cost; bounded rationality needed &
Kita~\cite{kita1999merging}, Sadigh et al.~\cite{sadigh2016planning}, iLQGames~\cite{fridovichkeil2020ilqgames}, GameFormer~\cite{huang2023gameformer}, Level-$k$~\cite{stahl1995levels,chong2005cognitive}, BeTop~\cite{liu2024reasoning} \\

Hybrid Data-Driven Agents &
Combines imitation learning base with RL fine-tuning or return conditioning; controllable behavior knobs &
Balances realism and controllability; behavior steering via conditioning; stable closed-loop rollouts &
Controllability can leave data support; reward misspecification; calibration of latent controls &
TrafficBots~\cite{zhang2023trafficbots}, TrafficBots V1.5~\cite{zhang2024trafficbots}, CtRL-Sim~\cite{rowe2024ctrl}, TrajGen~\cite{zhang2022trajgen} \\

%% --- SUB-HEADER: Agentic Foundation-Model Coordination ---
\midrule
\multicolumn{5}{l}{\textbf{\textit{Agentic Foundation-Model Coordination (decentralized, centralized, or hierarchical)}}} \\
\midrule

Agentic Foundation-Model Coordination &
LLM/VLM agents with communication, memory, and reflection modules coordinate multi-vehicle behavior; architectures could be decentralized, centralized, and hierarchical &
Semantic reasoning; emergent social norms; interpretable decisions; flexible coordination &
Grounding difficulty; temporal inconsistency; scalability bottleneck; verification of joint outputs &
AgentsCoDriver~\cite{hu2024agentscodriver}, KoMA~\cite{jiang2024koma}, CCMA~\cite{zhang2025ccma}, Wang et al.~\cite{wang2024llmsocialnorms} \\

\bottomrule
\end{tabularx}
\end{table*}

\subsection{Joint Trajectory Forecasting}
\label{subsec:joint_forecasting}

Joint trajectory forecasting models predict the conditional distribution of future motion for multiple agents simultaneously and explicitly capture inter-agent dependencies that single-agent prediction methods (Section~\ref{sec:traj_pred}) ignore. 

Mathematically, given past states for $N$ agents and map context $m$, the goal is to learn:
\begin{equation}
    P(Y^{1:N} \mid X^{1:N}, m)
\end{equation}
where $X^{i} = \{x_1^{i}, x_2^{i}, \ldots, x_{T_h}^{i}\}$ represents the observed trajectory history for agent $i$ over horizon $T_h$, $Y^{i} = \{y_1^{i}, y_2^{i}, \ldots, y_{T_f}^{i}\}$ denotes the predicted future trajectory for agent $i$ over forecast horizon $T_f$, and $Y^{1:N} = \{Y^{1}, Y^{2}, \ldots, Y^{N}\}$ denotes the joint future for all agents. The critical distinction from single-agent trajectory prediction is how inter-agent dependencies are modeled within the joint distribution $P(Y^{1:N} \mid X^{1:N}, m)$. Factorizing this as $\prod_{i=1}^{N} P(Y^{i} \mid X^{1:N}, m)$ (predicting each agent independently given shared context) captures contextual influence but not the constraint that future trajectories must be jointly consistent. True joint prediction instead models correlations across agents, producing consistent and collision-free future trajectories, where, for instance, one agent yielding is paired with another proceeding. This joint consistency is essential for simulation.

Joint trajectory forecasting models are predominantly evaluated in open-loop settings, comparing predicted trajectories against ground-truth recordings using displacement metrics such as minADE and minFDE. However, they can also be deployed in closed-loop simulation by re-querying the model at each time step with updated observations. However, this receding-horizon approach introduces challenges around temporal consistency because the model is not trained to maintain coherent behavior across successive re-invocations. This dual-use nature makes joint forecasting a bridge between pure prediction and interactive simulation.

Foundational work in this domain established key paradigms for encoding agent interactions. Social LSTM~\cite{alahi2016social} introduced social pooling mechanisms capturing spatial dependencies among agents using LSTM encoders with grid-based pooling of neighboring hidden states. Social GAN~\cite{gupta2018socialgan} incorporated adversarial training with global pooling, using a variety loss that promotes diverse yet plausible multi-agent futures by encouraging the generator to cover multiple behavioral modes. DESIRE~\cite{lee2017desire} encoded diverse intents through stochastic latent variables with scene-level refinement that conditions trajectory generation on both learned intent and environmental context. Trajectron++~\cite{salzmann2020trajectronpp} modeled heterogeneous agent types through dynamic interaction graphs with edge types reflecting semantic relationships (vehicle--vehicle, vehicle--pedestrian). SoPhie~\cite{sadeghian2019sophie} combined social attention over agent interactions with physical attention to scene context, jointly attending to who matters and where constraints lie. PRECOG~\cite{rhinehart2019precog} used normalizing flows to produce diverse, goal-conditionable joint futures with exact likelihood computation, which enables principled probabilistic evaluation.

As the field scaled to larger agent counts and more complex road topologies, vectorized map and scene-graph representations became dominant. VectorNet~\cite{gao2020vectornet} encoded both map elements and agent trajectories as polylines processed through hierarchical graph networks. LaneGCN~\cite{liang2020lanegcn} constructed explicit lane graphs to preserve topological connectivity through multi-scale graph convolutions that propagate information along and across lanes. TNT~\cite{zhao2021tnt} decomposed prediction into target endpoint prediction from lane centerlines and trajectory completion conditioned on selected goals, structuring multi-modality around discrete spatial anchors.

In recent years, transformer-based architectures unified agent interaction modeling and scene reasoning through attention mechanisms. Scene Transformer~\cite{ngiam2022scenetransformer} introduced a unified architecture with masking strategies to enable flexible querying, which means the same model can predict marginal or joint futures by adjusting which agents are masked during decoding. AgentFormer~\cite{yuan2021agentformer} combined transformers with latent variables using stochastic attention to model both social interactions and multi-modal intent simultaneously. HiVT~\cite{zhou2022hivt} introduced a two-stage architecture separating local context extraction from global interaction modeling by using translation- and rotation-invariant representations that improve generalization across scenes. Wayformer~\cite{nayakanti2023wayformer} applied efficient attention mechanisms (including latent query attention) for scalable processing of large agent sets. QCNet~\cite{liu2023qcnet} and QCNeXt~\cite{zhong2023qcnext} emphasized structured scene querying with query-centric designs to further improve computational efficiency. MTR-A~\cite{shi2022mtra} introduced the Motion Transformer framework by combining learned intention points with iterative motion refinement, which resulted in significant performance improvement. MTR++~\cite{shi2023mtrpp} extended this foundation with symmetric scene modeling and strengthened intent querying. MotionDiffuser~\cite{jiang2023motiondiffuser} represents a growing diffusion-based direction in the latest studies. It applies denoising diffusion to joint trajectory generation for controllable, high-diversity sampling with the ability to incorporate constraints during the reverse process. GameFormer~\cite{huang2023gameformer} incorporates game-theoretic reasoning into the transformer framework by modeling strategic interactions where agents' predicted futures reflect anticipation of others' responses rather than independent extrapolation.

To summarize, joint trajectory forecasting has become increasingly popular, following the recent Waymo Open Dataset Challenges \cite{montali2023wosac, waymo2021public}. Despite this interest, core limitation of joint trajectory forecasting for simulation is still the mismatch between \emph{distributional realism} and \emph{causal reactivity}. These models can achieve low displacement error by matching data distributions under passive observation, yet the predicted futures do not reflect how agents would react if the ego vehicle deviated from its logged behavior. They model correlations observed in data but not the causal mechanisms generating those correlations. This is important, especially when simulation is used to evaluate what happens under counterfactual ego actions. Nevertheless, joint forecasting is still valuable in several simulation-adjacent roles. Multi-modal joint predictions can seed downstream planners or closed-loop simulators with plausible initial futures, which reduces the search space for trajectory optimization. They can also support scenario mining, where identifying interactions with high uncertainty or predicted conflict flags situations that warrant closer testing. Finally, aggregated occupancy or flow forecasts derived from joint predictions offer lightweight, risk-aware scene summaries that planners can consume without reasoning over individual agent trajectories explicitly.

Subsequent studies could explore training objectives that explicitly penalize joint inconsistency (e.g., collision-aware losses), uncertainty-aware scoring that aligns predicted probabilities with empirical frequencies of behavioral modes, tighter bridges to closed-loop evaluation through receding-horizon deployment with temporal consistency regularization, and metrics that measure closed loop and intervention sensitivity rather than only log-matching accuracy.

\subsection{Centralized Modeling and Simulation}
\label{subsec:centralized_simulation}

Centralized simulation models scene evolution through a single learned model that jointly determines how all agents move. Rather than assigning independent policies to each agent. A centralized simulator takes the complete scene state as input and produces the next scene state, enforcing inter-agent consistency through shared computation. The key distinction from decentralized simulation (Section~\ref{subsec:decentralized_methods}) is that coupling between agents is enforced \emph{within} the model architecture, which means a single forward pass produces globally consistent next states rather than emerging from the interaction of separate per-agent policies. In the following subsections, the major classes of centralized modeling and simulation methods are reviewed. 

\subsubsection{Diffusion-Based Simulators}

Diffusion-based methods generate realistic multi-agent traffic by starting from random noise and gradually refining it into plausible joint trajectories. The approach defines a denoising process over a joint trajectory tensor $\mathbf{Y}\in\mathbb{R}^{N\times T_f\times d}$ stacking future states for all $N$ agents over a short forecast horizon $T_f$ in $d$-dimensional state space. Let $\mathbf{Y}^{(0)}$ denote the clean joint trajectory and $\mathbf{Y}^{(n)}$ denote the noised version at diffusion step $n$. The forward process progressively corrupts trajectories by adding Gaussian noise, and the model learns to reverse this corruption. The training objective minimizes:
\begin{equation}
    \min_{\theta}\;\mathbb{E}_{n,\epsilon}\left[\left\|\epsilon - \epsilon_{\theta}(\mathbf{Y}^{(n)}, n, m, c)\right\|^2\right]
\end{equation}
where $\epsilon \sim \mathcal{N}(0, I)$ is the noise added to trajectories, $\epsilon_{\theta}$ is a learned noise prediction network, $n$ indexes diffusion steps (distinct from simulation time $t$), $m$ represents map context, and $c$ denotes conditioning signals including initial scene state and ego actions. At inference, the model samples from Gaussian noise and iteratively denoises:
\begin{equation}
    \mathbf{Y}^{(n-1)} = \frac{1}{\sqrt{\alpha_n}}\left(\mathbf{Y}^{(n)} - \frac{1-\alpha_n}{\sqrt{1-\bar{\alpha}_n}}\epsilon_{\theta}(\mathbf{Y}^{(n)}, n, m, c)\right) + \sigma_n \mathbf{z}
\end{equation}
where $\alpha_n$ and $\bar{\alpha}_n$ are noise schedule parameters at diffusion step $n$, $\sigma_n$ is the noise level, and $\mathbf{z} \sim \mathcal{N}(0, I)$. The model can be steered toward desired behaviors in two ways: by training it with specific conditions baked into $c$, or by nudging the denoising process at inference time using gradient-based guidance toward goals like collision avoidance. For closed-loop simulation, the model is re-run at every timestep with the latest scene state as input, so traffic agents continuously react to what the ego vehicle does.

In this class of models, CTG~\cite{zhong2023ctg} demonstrates diffusion generating socially consistent multi-agent trajectories with controllability through conditioning signals. CTG++~\cite{zhong2023ctgpp} extends with language conditioning to enable semantically specified scene evolution valuable for scenario-based testing. MotionDiffuser~\cite{jiang2023motiondiffuser}, while primarily being a forecasting model (discussed in Section~\ref{subsec:joint_forecasting}), demonstrated architectural principles (joint diffusion over all agents with constraint-based guidance) that directly inform the design of diffusion-based simulators. Building on these foundations, diffusion-based approaches have been extended with increasingly expressive controllability interfaces for scenario-level generation, including differentiable cost function guidance and language-based conditioning. These controllable synthesis methods are discussed in Section~\ref{sec:scenario_control_synthesis} as they serve primarily a scenario generation role. 

The primary challenge for diffusion-based simulators is computational cost.  Iterative denoising requires multiple neural network evaluations per simulation step, and this cost multiplies when the sampler must be re-invoked at each rollout timestep for closed-loop operation. Closed-loop stability presents a second challenge, as even locally realistic samples can drift over long horizons when successive re-conditionings accumulate small inconsistencies. A third concern is that guidance-based controllability provides soft rather than hard constraint satisfaction. For instance, biasing the sampling distribution toward collision avoidance does not guarantee it, which is problematic for safety-critical simulation. Promising directions include accelerated sampling through diffusion distillation (reducing the number of denoising steps), consistency models that generate high-quality samples in fewer steps, uncertainty-aware guidance that becomes conservative under distributional ambiguity, and hybrid architectures combining diffusion sampling for multi-modal diversity with deterministic components enforcing hard physical constraints.

\subsubsection{Token-Based and Autoregressive Simulators}

Token-based simulators discretize continuous agent motion into sequences of discrete tokens, which is basically recasting multi-agent simulation as conditional sequence modeling analogous to language generation. This formulation leverages the scalability and expressiveness of transformer architectures developed for natural language processing by applying them to the structured sequential prediction problem of traffic evolution. Let $\mathcal{V}$ denote a learned vocabulary of motion tokens obtained through vector quantization of trajectory segments, and let $z_t^{i} \in \mathcal{V}$ represent the discretized motion token for agent $i$ at time $t$. Given map context $m$ encoding road geometry and topology, an autoregressive simulator learns the conditional distribution over the next-step tokens for all $N$ agents, factorized over a serialization of tokens within the timestep:
\begin{equation}
    p_\theta\!\left(z_{t+1}^{1:N}\mid z_{\le t}^{1:N}, m\right) = \prod_{j=1}^{N} p_\theta\!\left(z_{t+1}^{(j)} \mid z_{\le t}^{1:N},\, m,\, z_{t+1}^{(<j)}\right)
\end{equation}
where $z_{\le t}^{1:N}$ denotes all motion tokens up to time $t$ for all agents, and $z_{t+1}^{(<j)}$ denotes tokens already generated at timestep $t+1$ before serial position $j$. This intra-timestep factorization is critical. Later tokens in the serialization are conditioned on earlier tokens within the same timestep, which implicitly models within-step agent interactions through the autoregressive ordering. Training uses standard cross-entropy next-token prediction over the full sequence. For closed-loop simulation, tokens are sampled sequentially at each timestep, decoded back to continuous states through the inverse of the vector quantization mapping, and the process iterates with the decoded states forming the conditioning for the next step.

MotionLM~\cite{seff2023motionlm} introduced the paradigm of discretizing trajectories into motion tokens and training autoregressive transformers for joint multi-agent futures. Trajeglish~\cite{philion2024trajeglish} refined this approach by modeling traffic as next-token prediction with fine spatial resolution and explicit accounting for intra-timestep agent interactions through careful serialization ordering. BehaviorGPT~\cite{zhou2024behaviorgpt} proposed next-patch prediction (NP3), where multi-step motion patches are predicted as single units rather than individual tokens. This diminishes the ``copying shortcut'' problem where models learn to simply repeat the previous token rather than generating meaningful motion updates. SMART~\cite{wu2024smart} extended the tokenization framework with explicit road tokens encoding map structure within the same vocabulary by using decoder-only transformers. It demonstrates scalability across datasets, cross-dataset generalization, and real-time inference capability.

The central tension in token-based simulation is between discretization fidelity and sequence tractability. Finer-grained tokenization (larger vocabularies and higher spatial resolution) reduces quantization error but increases sequence length, which increases the computational cost and exacerbates exposure to compounding errors over long rollouts. The serialization order of agents within each timestep introduces an arbitrary asymmetry, which means the first agent's token is generated without conditioning on other agents' current actions, while later agents benefit from this information. Compounding error is particularly critical since a single incorrect token can shift the agent to an implausible state and potentially cascading into increasingly unrealistic rollouts. Further research could focus on hybrid approaches combining discrete token generation with continuous refinement layers, learned serialization orderings that minimize information asymmetry, training paradigms incorporating scheduled sampling or rollout-aware losses to improve long-horizon robustness, and constrained beam search or rejection sampling for enforcing hard physical constraints.

\subsubsection{Regression-Based Reactive Simulators}

Regression-based reactive simulators directly predict continuous state updates for all agents through learned regression functions, without the intermediate discretization of token-based methods or the iterative sampling of diffusion models. These methods emphasize computational efficiency and deterministic prediction, which makes them well suited for real-time closed-loop deployment.

SimNet~\cite{bergamini2021simnet} established the paradigm of learning reactive traffic simulation by training on large-scale perception outputs to produce agents that respond to ego deviations during closed-loop rollout. The architecture encodes scene context through rasterized representations and predicts per-agent trajectory continuations. This work demonstrated that data-driven reactive simulation is feasible at scale. TrafficSim~\cite{suo2021trafficsim} advanced this approach by explicitly modeling agent-to-agent dependencies through graph neural network architectures and shows that interaction-aware scene representations substantially improve the realism of joint rollouts compared to independent per-agent prediction.

The challenges in regression-based reactive methods is the gap between short-horizon accuracy and long-horizon stability, where compounding prediction errors accumulate over extended rollouts. Because regression-based methods typically produce point estimates rather than distributions over next states, they capture limited multi-modality. Causal validity is another concern: learning from passive observational logs does not uniquely identify how agents would respond to counterfactual ego actions, and the model may learn correlations that break under intervention. Future directions include multi-modal regression heads that output mixture distributions over next states, intervention-aware training that augments logged data with synthetic ego deviations, uncertainty estimation that flags when predictions are unreliable, and curriculum-based unrolling strategies that progressively increase rollout horizons during training to improve long-horizon stability.

\subsection{Decentralized Multi-Agent Methods}
\label{subsec:decentralized_methods}

Decentralized multi-agent simulation assigns each traffic participant its own policy, letting scene-level behavior emerge from the coupled execution of independent decision-makers. Formally, this setting is modeled as a Markov game (also called stochastic game) with $N$ agents, global state space $\mathcal{S}$, per-agent action spaces $\{\mathcal{A}^i\}_{i=1}^{N}$, per-agent observation functions, and transition dynamics:
\begin{equation}
    s_{t+1} \sim p(s_{t+1}\mid s_t, a_t^1,\ldots,a_t^N)
\end{equation}
where $s_t \in \mathcal{S}$ is the global state, $a_t^i \in \mathcal{A}^i$ is the action of agent $i$, and $p$ represents the environment transition dynamics that depend on all agents' joint actions. It is worth noting that while the world's transition depends on all joint actions, each agent's policy can only condition on its own observations. Each agent $i$ selects actions based on local observations: $a_t^i \sim \pi_{\theta_i}(\cdot \mid o_t^i)$, where $o_t^i$ is agent $i$'s (partial) observation of $s_t$. The key architectural distinction from centralized simulation (Section~\ref{subsec:centralized_simulation}) is that no single model jointly determines all agents' next states. 
 
When deployed reactively, decentralized methods operate in a closed-loop fashion, as each agent conditions its actions on current observations. However, they can equally be trained via open-loop objectives such as behavior cloning or maximum likelihood, and evaluated through fixed-horizon rollouts without external intervention. The subsubsections below organize decentralized approaches by how agent policies are obtained, that is through reward optimization (MARL), demonstration matching (imitation and adversarial learning), strategic reasoning (game theory), or combination of these approaches (hybrid methods).

\subsubsection{Multi-Agent Reinforcement Learning and Self-Play}

Multi-agent reinforcement learning (MARL) learns decentralized driving policies by optimizing expected return in Markov games. Each agent $i$ has its own reward function $r_i$ and learns policy $\pi_{\theta_i}$ to maximize expected cumulative discounted return:
\begin{equation}
    J_i(\theta_i) = \mathbb{E}_{\pi_{\theta_1}, \ldots, \pi_{\theta_N}}\left[\sum_{t=0}^{\infty}\gamma^t\, r_i(s_t, a_t^1,\ldots,a_t^N)\right]
\end{equation}
where $\gamma \in [0,1]$ is the discount factor and the expectation is over joint trajectories generated by all agent policies simultaneously. The fundamental difficulty is that each agent's optimization landscape depends on all other agents' evolving policies, which creates a non-stationary learning problem where the environment effectively changes as co-learners update their behavior.

A dominant paradigm for managing this non-stationarity is \emph{centralized training with decentralized execution} (CTDE). During training, each agent has access to a centralized critic $Q_{\phi_i}(s_t, a_t^1, \ldots, a_t^N)$ parameterized by $\phi_i$ that observes the global state and all agents' actions, providing more stable value estimates than would be possible from local observations alone. At execution time, each agent acts using only its local observation:
\begin{equation}
    a_t^i \sim \pi_{\theta_i}(\cdot \mid o_t^i)
\end{equation}
This asymmetry (global information for learning, local information for acting) enables agents to internalize the effects of other agents' behavior during training while remaining deployable in partially observed settings. Self-play, a special formulation of MARL where agents train against copies of themselves or a population of past policy checkpoints, provides an automatic curriculum: as policies improve, the training environment becomes correspondingly more challenging and produce increasingly robust interaction strategies.

Foundational work established MARL's viability for traffic applications. Shalev-Shwartz et al.~\cite{shalev2016safe} proposed a safety-oriented multi-agent formulation that decomposes driving into learned high-level ``desires'' (e.g., target lane, desired speed) and constraint-satisfying trajectory planners that ensure feasibility and aims to address the challenge that unconstrained RL policies may produce physically implausible or dangerous actions. Flow~\cite{wu2021flow} provided a computational framework for studying mixed autonomy traffic through MARL and demonstrates that even small penetration rates of RL-controlled vehicles can produce emergent improvements in traffic throughput. Nocturne~\cite{vinitsky2022nocturne} introduced a purpose-built multi-agent driving environment designed for MARL research by featuring 2D partially observed scenarios running at over 2,000 steps per second. This efficiency enables the large-scale experiments (millions of episodes) that MARL algorithms require. Building on the need for high-throughput multi-agent environments, recent simulators, such as GPUDrive~\cite{kazemkhani2025gpudrive} and PufferDrive~\cite{pufferdrive2025github} achieve substantially higher efficiency and flexibility.

Recent work has focused on producing human-compatible rather than merely reward-optimal behavior. The observation that self-play alone can converge to superhuman or non-human strategic equilibria motivates data-regularized approaches. Human-Regularized PPO~\cite{cornelisse2024human} addresses this by augmenting the self-play objective with demonstration regularization, which penalizes policies that deviate from recorded human driving distributions while still benefiting from self-play's robustness to adversarial interactions. This anchoring to human data is critical for simulation applications where the goal is reproducing realistic mixed traffic rather than optimal traffic. Complementing this direction, Robust Autonomy from Self-Play~\cite{cusumanotowner2025robust_selfplay} demonstrated that policies emerging from large-scale self-play exhibit robustness to diverse interaction partners, including out-of-distribution agents not encountered during training. However, compared to \cite{cornelisse2024human}, the reward design and shaping is quite extensive. Bronstein et al.~\cite{bronstein2022latentactions} showed that structuring the action space through learned latent representations simplifies multi-agent policy learning and improves transfer across scenarios.

One of the main challenges for MARL in driving simulation is non-stationarity and credit assignment. Each agent's learning changes the environment for all others. This makes convergence difficult and potentially causes oscillatory training dynamics. Reward specification is also demanding. Small changes in reward weights can produce qualitatively different emergent behaviors, and multi-objective rewards balancing safety, progress, comfort, and rule compliance are difficult to tune. Future directions include stronger CTDE architectures with uncertainty-aware critics that distinguish epistemic from aleatoric uncertainty in other agents' behavior, hybrid training that anchors MARL policies to real driving logs through behavioral priors, population-based training that maintains a diverse pool of driving styles preventing convergence to a single behavioral mode, and explicit safety constraints that remain stable under multi-agent distributional shift.

\subsubsection{Multi-Agent Imitation and Adversarial Learning}

Multi-agent imitation learning extends the imitation and adversarial methods discussed for single agents (Section~\ref{sec:irl_gail}) to decentralized multi-agent settings. The key difference from single-agent adversarial imitation is that the discriminator must evaluate \emph{joint} behavior and see if the collection of agents' actions matches the distributional properties of real multi-agent traffic.

Given expert demonstrations $\mathcal{D}_E = \{(s_j, a_j^{1:N})\}_{j=1}^{M}$ containing $M$ samples of global states $s_j$ and joint actions $a_j^{1:N} = \{a_j^1, \ldots, a_j^N\}$ for all $N$ agents, the goal is to learn per-agent policies $\{\pi_{\theta_i}\}_{i=1}^{N}$ whose joint rollouts are indistinguishable from expert behavior. In the multi-agent GAIL formulation, the objective is:
\begin{equation}
\begin{aligned}
\min_{\{\pi_{\theta_i}\}}\max_{D_{\psi}}\;&
\mathbb{E}_{\pi_\theta}\left[\log D_{\psi}(s, a^{1:N})\right] \\
&+ \mathbb{E}_{\pi_E}\left[\log(1-D_{\psi}(s, a^{1:N}))\right]
- \lambda \sum_{i=1}^{N}\mathcal{H}(\pi_{\theta_i})
\end{aligned}
\end{equation}
where $D_{\psi}$ is a discriminator evaluating joint state-action tuples, $\pi_E$ represents the expert's \textit{joint behavior}, and $\mathcal{H}(\pi_{\theta_i})$ is the entropy of agent $i$'s policy weighted by $\lambda > 0$. Training typically follows the CTDE paradigm. The discriminator operates on global state-action tuples during training, while execution remains decentralized with each agent acting from local observations $a_t^i \sim \pi_{\theta_i}(\cdot \mid o_t^i)$.

PS-GAIL~\cite{bhattacharyya2018multi} first scaled adversarial imitation to multi-agent driving by introducing parameter sharing across agents. All agents use the same policy network conditioned on agent-specific context. Parameter sharing is combined with curriculum learning that progressively increases rollout horizons during training to improve long-horizon interaction stability. The extended version~\cite{bhattacharyya2023titsgail} systematized this framework by adding Reward Augmented Imitation Learning (RAIL) for injecting domain knowledge (e.g., lane-keeping and speed maintenance rewards alongside the adversarial signal) and Burn-InfoGAIL for disentangling latent driving style factors. Symphony~\cite{igl2022symphony} addressed a key failure mode of multi-agent adversarial learning, which is diversity collapse where all agents converge to similar behavior. This is achieved by combining imitation learning with discriminator-guided parallel beam search during rollout and hierarchical goal mechanisms that maintain distinct agent intents over long horizons. DecompGAIL~\cite{guo2025decompgail} identified another failure mode specific to multi-agent settings: ``irrelevant interaction misguidance'', where a global discriminator penalizes an agent for behavior that appears unrealistic only because of \emph{other} agents' errors. The solution decomposes discrimination into ego-map components (evaluating map compliance independently) and ego-neighbor components (evaluating interaction realism separately) to prevent irrelevant signals from destabilizing individual agent learning.

Counterfactual validity is still the most fundamental concern about multi-agent imitation. Since training data consists of passive logs, the discriminator can only evaluate whether joint behavior matches the \emph{observed} distribution of interactions, not whether agents respond correctly to \emph{novel} interventions. Training stability is more difficult than in single-agent settings because the adversarial dynamics must simultaneously shape multiple policies whose interactions create additional non-stationarity beyond the standard generator-discriminator oscillation. The diversity-versus-realism tension is also concerning. Mode collapse in multi-agent settings can lead to loss of behavioral variety and interaction variety, which itself can generate traffic where all agents exhibit similar responses to the same situations. Moving forward, it will be beneficial to explore factored discriminator architectures that isolate sources of unrealism, hybrid training combining behavior cloning pre-training with adversarial fine-tuning for stability, uncertainty-aware discriminators that reduce gradient magnitude in states where expert behavior is ambiguous, and evaluation protocols that explicitly test intervention response by measuring how agent behavior changes when the ego vehicle deviates from logged trajectories.

\subsubsection{Game-Theoretic and Strategic Interaction Models}

Game-theoretic models formalize driving interaction as strategic reasoning among decision-makers with coupled objectives and provide a principled framework for understanding phenomena such as yielding, merging negotiation, and implicit coordination at unsignalized intersections. While MARL also operates in the Markov game framework, the key distinction is that MARL learns policies through trial-and-error interaction, whereas game-theoretic approaches compute or approximate equilibrium solutions analytically without requiring extensive simulation experience.

Mathematically, each agent $i$ seeks to optimize its utility $J_i$ that depends on all agents' strategies:
\begin{equation}
    J_i(\pi_1,\dots,\pi_N) = \mathbb{E}_{\pi_1,\dots,\pi_N}\left[\sum_{t=0}^{T}\gamma^t\, r_i(s_t, a_t^1,\dots,a_t^N)\right]
\end{equation}
where $T$ is the interaction horizon. A Nash equilibrium is a joint policy $(\pi_1^*, \ldots, \pi_N^*)$ where no agent can improve its utility by unilateral deviation:
\begin{equation}
    J_i(\pi_1^*,\dots,\pi_i^*,\dots,\pi_N^*) \geq J_i(\pi_1^*,\dots,\pi_i,\dots,\pi_N^*), \quad \forall \pi_i, \; \forall i
\end{equation}
In Stackelberg formulations, which is natural for modeling ego-vehicle planning among reactive human drivers, the ego vehicle acts as leader, choosing its policy $\pi^{\text{ego}}$ while anticipating that other agents will best-respond:
\begin{equation}
    \pi^{\text{ego},*} = \arg\max_{\pi^{\text{ego}}} J^{\text{ego}}\!\left(\pi^{\text{ego}}, \text{BR}_{-\text{ego}}(\pi^{\text{ego}})\right)
\end{equation}
where $\text{BR}_{-\text{ego}}(\pi^{\text{ego}})$ denotes the collection of best-response policies for all non-ego agents given the leader's policy. Computing best responses requires solving each follower's optimization problem given the leader's committed strategy, making the Stackelberg formulation a bilevel optimization.

Game-theoretic approaches can operate in both open-loop and closed-loop modes. Open-loop game solutions compute complete strategy profiles (sequences of actions) before execution, suitable for short-horizon interactions where replanning is unnecessary. Closed-loop (feedback) strategies condition actions on the current state at each timestep and produce reactive policies that adapt to the evolving interaction.

Kita~\cite{kita1999merging} modeled highway merging as a game where gap acceptance decisions depend on strategic reasoning about the mainline driver's likely response, demonstrating that human merge behavior is better explained by game theory than by fixed gap thresholds. Sadigh et al.~\cite{sadigh2016planning} formulated automated planning as actively influencing human actions through Stackelberg games and showed that an ego vehicle can leverage its commitment power to create more favorable interactions (e.g., nudging forward to claim right-of-way). iLQGames~\cite{fridovichkeil2020ilqgames} introduced efficient iterative linear-quadratic solvers for general-sum dynamic games, which resulted in producing feedback (closed-loop) Nash equilibrium strategies in real time. This is important for planning at the speeds required for driving. GameFormer~\cite{huang2023gameformer} incorporated game-theoretic structure into transformer-based architectures for joint interactive prediction and planning to capture how agents' predicted futures influence each other.

An important insight is that human drivers do not reason with unlimited strategic depth. Bounded rationality models using level-$k$ reasoning~\cite{stahl1995levels,chong2005cognitive} assume that level-0 agents follow simple heuristic policies (e.g., maintain current speed), level-1 agents best-respond to level-0 behavior, level-2 agents best-respond to level-1, and so on. This framework provides a controlled spectrum from reactive heuristics to deep strategic reasoning. This (arguably) better matches the heterogeneity of real driving populations where some drivers plan ahead while others react instinctively. BeTop~\cite{liu2024reasoning} introduces behavioral topology as a structured representation for multi-agent interactions and reasons about relative spatial-temporal relationships between agents (who yields to whom) rather than absolute positions. By modeling topological relations explicitly, BeTop captures the strategic structure of driving interactions in a form that is both interpretable and compatible with downstream planning.

Utility specification is a demanding task in game theoretic approaches. Small changes in reward weights can qualitatively flip equilibrium behavior (e.g., switching from yielding to aggressive merging). Equilibrium selection poses a further problem. Many driving interactions admit multiple Nash equilibria (e.g., both ``I yield, you go'' and ``you yield, I go'' may be equilibria), and selecting among them requires additional assumptions about conventions, precedence, or focal points. Computational cost is significant for dynamic games with many agents and long horizons, as the state space grows combinatorially. Most critically, the assumption of rationality (even bounded rationality) may not capture the full range of human driving behavior, such as distraction, confusion, and rule violations. Future directions could include differentiable game solvers with learned utility functions calibrated from driving data, data-driven bounded rationality models that learn the distribution of reasoning depths from observed interactions, and evaluation protocols testing whether game-theoretic agents produce realistic equilibrium selection patterns under distribution shift.

\subsubsection{Hybrid Data-Driven Agents}

Hybrid approaches combine data-driven imitation with structured mechanisms for controllability and physical feasibility. The motivation is that pure imitation produces realistic but uncontrollable agents, pure RL produces controllable but potentially unrealistic agents, and hybrid methods seek the best of both by starting from data-driven foundations and adding targeted modifications.

TrafficBots~\cite{zhang2023trafficbots} demonstrated the effectiveness of goal-and-personality conditioning for scalable multi-agent simulation. It trains closed-loop policies on large-scale driving datasets where each agent is conditioned on its extracted destination and a latent personality inferred via CVAE. The approach scales efficiently to dense urban scenes because the shared policy handles variable agent counts. TrafficBots V1.5~\cite{zhang2024trafficbots} improved upon this foundation with stronger transformer-based scene encoders and training techniques (including better data augmentation and loss scheduling) for improved closed-loop performance on the Waymo Sim Agents benchmark \cite{montali2023wosac}. CtRL-Sim~\cite{rowe2024ctrl} introduced return-conditioned offline RL to multi-agent simulation setting. It trains on logged data annotated with per-timestep rewards for safety, progress, and comfort, then conditioning the policy on desired return levels at inference. This enables test designers to generate specific behavioral profiles (e.g., a tailgating agent, a hesitant merger) by specifying return targets without retraining. TrajGen~\cite{zhang2022trajgen} takes a two-stage approach, first generating diverse candidate trajectories using a prediction model and then applying RL-based refinement to modify trajectories under kinematic constraints while avoiding collisions.

The central tension in hybrid approaches is between controllability and realism. As behavioral conditioning moves agents further from the modes well-represented in training data, the policy operates increasingly out-of-distribution, and predicted behaviors may become implausible. Aggressive return targets or extreme personality samples can produce ``sim-native'' behaviors that are internally consistent but do not correspond to any observed human driving pattern. Calibration of latent controls to interpretable and consistent behaviors remains difficult. It is often unclear what specific latent values correspond to a specific observable driving characteristic. Future frameworks should focus on evaluation frameworks that measure the controllability-versus-realism trade-off explicitly, and try to disentangle latent spaces where individual dimensions correspond to interpretable behavioral attributes.

\subsection{Agentic and Foundation-Model Coordination}
\label{subsec:multi_agentic_foundation}

Foundation model approaches, introduced for single-agent ego planning in Section~\ref{sec:single_foundation_models}, are increasingly explored for multi-agent coordination as reasoning layers that mediate interactions across multiple traffic participants. The foundation model in this context is aimed at maintaining temporal and inter-agent consistency across multiple simultaneous reasoning processes, avoiding producing over-coordinated behaviors that would be unrealistic without explicit vehicle-to-vehicle communication, and scaling its reasoning across variable numbers of participants with heterogeneous intents.

The general formulation treats the foundation model as producing high-level decisions for agent $i$ conditioned on multi-agent scene context:
\begin{equation}
    y_t^i = f_{\theta}^i\left(\mathcal{P}\left(o_t^{1:N}, m, g^{1:N}, h_t^{1:N}\right), \mathcal{M}_t^i\right)
\end{equation}
where $o_t^{1:N}$ represents observations for all $N$ agents, $m$ denotes map context, $g^{1:N}$ represents goals for all agents, $h_t^{1:N}$ denotes intent or high-level plan representations, $\mathcal{P}(\cdot)$ is a prompt builder structuring multi-agent context, $\mathcal{M}_t^i$ represents memory or retrieved knowledge for agent $i$, and $y_t^i$ is the high-level decision output subsequently grounded into a control action $a_t^i = \mathcal{C}^i(y_t^i, s_t^i)$ through a low-level controller $\mathcal{C}^i$. An important design tension concerns the information structure. Providing each agent's model with observations of all other agents $o_t^{1:N}$ enables globally informed reasoning but is unrealistic. Systems sharing full scene information risk producing coordination that implicitly assumes omniscience or communication. Therefore, restricting each agent to local observations $o_t^i$ is more faithful to real driving although it limits the model's reasoning about distant or occluded agents. 

In this class of models, AgentsCoDriver~\cite{hu2024agentscodriver} formulates multi-vehicle collaborative driving as a Decentralized Partially Observable Markov Decision Process (Dec-POMDP) and equips each LLM-based agent with five modules: observation, reasoning engine, cognitive memory, reinforcement reflection, and communication. The communication module allows agents to determine \emph{when} to communicate (based on potential trajectory conflicts) and \emph{what} to convey (via an LLM-instantiated message generator). This provides a principled framework for selective information sharing rather than broadcasting the global state. The reinforcement reflection module evaluates outcomes and updates experience memory to enable a form of lifelong learning across episodes. KoMA~\cite{jiang2024koma} proposes a knowledge-driven framework where multiple LLM-powered agents analyze surrounding vehicles' behaviors to infer intentions. A multi-step planning module structures reasoning through a goal--plan--action hierarchy. KoMA demonstrates that structured multi-agent reasoning with shared knowledge substantially outperforms single-agent LLM approaches in highway merging scenarios.

CCMA~\cite{zhang2025ccma} introduces a hierarchical architecture by integrating reinforcement learning for individual-level optimization with fine-tuned LLMs for regional and global coordination. Rather than using LLMs for direct control, CCMA leverages them to dynamically tune reward functions to balance traffic flow, cooperation, comfort, and safety. This hybrid RL-LLM design addresses the latency problem by confining LLM inference to less frequent strategic decisions. Wang et al.~\cite{wang2024llmsocialnorms} investigated whether LLM agents playing Markov games in multi-agent driving scenarios can develop emergent social norms. Their experiments showed that social norms do emerge, with agents adopting conservative policies in collision-prone situations and suggested that LLMs' pretraining on human-generated text encodes implicit social reasoning relevant to driving coordination. A comprehensive survey by Hu et al.~\cite{hu2025multiagentllmsurvey} systematically categorizes LLM-based multi-agent driving systems by interaction mode (multi-vehicle, vehicle-infrastructure, agent-human) and architectural pattern (centralized, decentralized, hierarchical).

These systems can operate in both open-loop and closed-loop modes. In open-loop mode, the foundation model generates a complete interaction plan for all agents from an initial scene description, useful for scenario generation and offline analysis. In closed-loop mode, agents are queried at each timestep with updated observations; however, the computational cost of LLM inference has led current implementations to reduce the query frequency (1--2 Hz).

Several trends are emerging in this domain. First, systems are shifting from free-form language outputs toward structured, testable interfaces with well-defined output schemas. Second, multi-agent coordination increasingly employs explicit communication abstractions (message protocols, shared memory pools) and ensuring that the level of coordination is realistic given assumed communication capabilities. Third, hybrid architectures coupling LLM-level reasoning with RL-trained or safety-certified low-level controllers are gaining more attention, as they separate the semantic understanding where LLMs excel from the real-time reactivity where neural controllers are more practical. Fourth, the question of whether LLMs can serve as models of \emph{human} driving cognition (rather than just optimal planners) is gaining attention, with social norm emergence studies suggesting that LLMs' implicit knowledge of human behavior may be directly applicable to simulating human-like traffic participants. Despite these progresses and attentions, these methods face open challenges. \textit{Grounding} remains the primary concern. Translating language-level multi-agent reasoning into geometrically precise, dynamically feasible, and mutually consistent trajectories for multiple agents is significantly harder than the single-agent case. \textit{Scalability} is a fundamental bottleneck.  Running LLM inference for every agent at every timestep is computationally prohibitive for large-scale traffic simulation; even hierarchical designs face challenges when the number of interacting agents grows. \textit{Verification and safety assurance} require checking not only individual outputs but their joint consistency. \textit{Evaluation} requires closed-loop, intervention-based metrics assessing how multi-agent systems respond when individual agents are perturbed. Future directions include lightweight reasoning architectures, such as distilled models, cached reasoning patterns; formal verification layers checking joint physical consistency before execution; integration of LLM reasoning with game-theoretic solvers for principled equilibrium computation; and hybrid architectures allocating expensive LLM reasoning selectively to complex interactions while learned neural policies manage routine driving.

\subsection{Discussion on Multi-Agent Methods}
\label{subsec:multi_challenges}

Multi-agent methods explicitly model interaction but introduce challenges absent in the single-agent setting (Section~\ref{subsec:single_challenges}). While single-agent methods face the fundamental limitation of treating other agents as non-responsive environment components, multi-agent methods must satisfy three simultaneous requirements that create inherent tensions: (1) interaction realism, where agents respond plausibly to each other and to ego interventions; (2) scene-level consistency, where joint rollouts remain collision-free and map-compliant, unless scenarios specifically demand otherwise; and (3) scalability and controllability, where simulators remain computationally tractable and behaviorally steerable across varying traffic densities, road geometries, and testing objectives. 

The most fundamental challenge cutting across all multi-agent approaches is arguably counterfactual validity under intervention. Most models train from passive observational logs that do not uniquely identify how agents would react when the ego vehicle deviates from its recorded behavior. In practice, this means a simulator may produce highly realistic traffic when the ego follows its logged trajectory but generate implausible reactions when the ego deviates, such as agents failing to brake or yielding when they should not. This is a limitation that is particularly problematic for scenario-based testing and verification of automated vehicles. Addressing this requires either interventional training data (rare and expensive to collect), data augmentation strategies, or explicit causal models of agent response. Additionally, the compounding error in multi-agent settings is amplified by interaction feedback. When one agent's prediction error changes the input distribution for all others, cascading effects emerge that are absent in single-agent rollouts. Current mitigation strategies include unrolled training losses, scheduled sampling, and explicit reset mechanisms, but maintaining stable multi-agent rollouts over horizons exceeding 10 seconds remains open, particularly in dense urban traffic.

% Beyond these data and dynamics challenges, multi-agent methods face difficulties in credit assignment, controllability, and the tension between realism and behavioral steering. In decentralized learning, non-stationarity arises because each agent learns while the environments induced by other agents' evolving policies change simultaneously; in adversarial imitation, global discriminators can penalize an individual agent for behavior that appears unrealistic only because of other agents' errors---the ``irrelevant interaction misguidance'' identified by DecompGAIL~\cite{guo2025decompgail}; and in centralized simulators, the same observed scene evolution can be explained by different latent intent configurations, creating identifiability problems. 
The realism-controllability trade-off is an interesting area to explore. Test engineers require controllable agents to create specific stress-testing scenarios, but controllability mechanisms can push agent behavior beyond the support of logged data, also known as ``sim-native'' behaviors. Sim-native behaviors are internally consistent but do not necessarily correspond to observed human driving patterns. Parameter sharing, while essential for scalability, can reduce behavioral diversity unless heterogeneity is explicitly maintained. Quantifying where the boundary lies between controlled-but-realistic and controlled-but-implausible remains an open problem. Related to this, evaluation and benchmarking constitute perhaps the most critical bottleneck limiting the field's ability to compare and validate multi-agent methods. Open-loop prediction metrics such as minADE and minFDE do not guarantee closed-loop simulation quality. Closed-loop benchmarks such as WOSAC~\cite{montali2023wosac} and nuPlan~\cite{caesar2021nuplan} represent important progress but remain sensitive to evaluation infrastructure assumptions, including how ego behavior is specified, how collisions are detected, and what constitutes realistic interaction. Key evaluation dimensions that remain insufficiently addressed include intervention sensitivity, long-horizon distributional stability, calibration of multi-modality (whether behavioral diversity and mode probabilities match real-world frequencies), controllability fidelity (whether steering mechanisms produce intended effects without degrading realism), and geographic and cultural transfer (whether models trained in one city generalize to different road geometries, traffic rules, and driving norms). Developing comprehensive evaluation protocols addressing these dimensions is an important open research problem.

\section{ENVIRONMENT-LEVEL SIMULATION METHODS}
\label{sec:environment_level}

Sections~\ref{sec:single_agent}--\ref{sec:multi_agent} addressed how individual agents and groups of agents make decisions and generate trajectories. This section shifts focus to the \emph{environment side} of the simulation loop. The key architectural distinction from agent-level methods is that the modeling target is the environment state itself---observations, occupancy fields, or complete scene configurations---rather than individual agent actions or policies. We organize this section into generative world models (Section~\ref{sec:worldmodels_sim}), which learn environment dynamics from data, and traffic scenario generation (Section~\ref{sec:scenario_gen}), which constructs initial conditions and test configurations for simulation. It is worth noting that many of the methodological paradigms, especially in traffic scenario generation, are shared with methods presented in multi-agent and single-agent sections; however, the goal here is not to model how agents behave but to construct and generate the environments in which they are trained and tested. Table~\ref{tab:neural_sim_summary} provides a summary of paradigms discussed in this section.

\begin{table*}[!htbp]
\centering
\caption{Summary of neural simulation environment and scene generation methods for mixed autonomy traffic simulation.}
\label{tab:neural_sim_summary}
\renewcommand{\arraystretch}{1.25}
\setlength{\tabcolsep}{3pt}
\scriptsize
\begin{tabularx}{\textwidth}{
    >{\RaggedRight\arraybackslash}p{1.7cm}
    >{\RaggedRight\arraybackslash}p{1.5cm}
    >{\RaggedRight\arraybackslash}X
    >{\RaggedRight\arraybackslash}X
    >{\RaggedRight\arraybackslash}X
    >{\RaggedRight\arraybackslash}p{3.0cm}
}
\toprule
\textbf{Method} &
\textbf{Output Space} &
\textbf{Core Idea} &
\textbf{Strengths} &
\textbf{Limitations} &
\textbf{Representative Studies} \\

%% --- SUB-HEADER: Generative World Models ---
\midrule
\multicolumn{6}{l}{\textbf{\textit{Generative World Models}}} \\
\midrule

Video-Based \& Latent World Models &
Multi-view images; latent tokens; video frames &
Learns action-conditioned video or latent generation via autoregressive token prediction or latent diffusion; produces future visual observations for end-to-end evaluation &
Photorealistic rollouts; end-to-end stack evaluation; controllable via actions, text, and structured inputs; emergent scene dynamics &
Long-horizon drift; high computational cost; causal consistency under intervention not guaranteed; image quality metrics miss driving-critical correctness &
GAIA-1~\cite{hu2023gaia1}, GAIA-2~\cite{russell2025gaia2}, Drive-WM~\cite{wang2024drivewm}, DriveDreamer~\cite{wang2024drivedreamer}, DriveDreamer-2~\cite{zhao2025drivedreamer}, Vista~\cite{gao2024vista}, Copilot4D~\cite{zhang2023copilot4d}, MUVO~\cite{bogdoll2025muvo}, DrivingWorld~\cite{hu2024drivingworld}, GenAD~\cite{yang2024genad} \\

Occupancy-Based World Models &
3D/4D semantic occupancy grids; BEV fields &
Predicts future semantic occupancy conditioned on ego actions via autoregressive or diffusion generation in voxel or triplane space &
Geometric and semantic structure; directly supports planning cost evaluation; captures free-space and drivable-area evolution &
No agent identity or intent; compounding topology errors; memory-intensive voxel grids; expensive 4D labels &
OccWorld~\cite{zheng2024occworld}, OccSora~\cite{wang2024occsora}, DOME~\cite{gu2024dome}, T$^3$Former~\cite{xu2025t3former}, $I^{2}$-World~\cite{liao2025i2}, Drive-OccWorld~\cite{yang2025driving}, OccLLaMA~\cite{wei2024occllama} \\

%% --- SUB-HEADER: Traffic Scenario Generation ---
\midrule
\multicolumn{6}{l}{\textbf{\textit{Traffic Scenario Generation}}} \\
\midrule

Scene Initialization \& Scenario Synthesis &
Static scene layouts; multi-agent trajectories; map + traffic configurations &
Generates traffic configurations ranging from static actor placements to full dynamic scenarios; controllable methods expose control interfaces (language, temporal logic, guidance functions, retrieval tags) for steerable generation &
Data-driven scene diversity; user-specified controllability; intuitive language interfaces; compositional control; supports scenario augmentation and targeted testing &
Realism--control trade-off; stronger constraints push off data manifold; unstandardized control interfaces; scene initialization methods lack dynamic validation; limited closed-loop validation &
SceneGen~\cite{tan2021scenegen}, TrafficGen~\cite{feng2022trafficgen}, DriveSceneGen~\cite{sun2024drivescenegen}, CTG~\cite{zhong2023ctg}, CTG++~\cite{zhong2023ctgpp}, Scenario Diffusion~\cite{pronovost2023scenariodiffusion}, LCTGen~\cite{tan2023lctgen}, ChatSim~\cite{wei2024editable}, RealGen~\cite{ding2024realgen} \\

Safety-Critical Generation \& Adversarial Falsification &
Scenario parameters; adversarial agent trajectories; safety-critical multi-agent configurations &
Discovers failure-inducing scenarios via optimization-based search (RL adversaries, importance sampling, fuzzing) or safety-targeted generative synthesis (guided diffusion, latent-space optimization with learned traffic priors) &
Efficient failure discovery; targeted stress testing; identifies safety-critical edge cases; generative methods maintain realism via learned priors; complements naturalistic evaluation &
Validity--criticality trade-off; adversarial drift to implausible scenarios; coverage gaps; failure finding $\neq$ risk estimation; guided generation provides soft not hard constraint satisfaction &
AST~\cite{koren2018ast_iv,lee2020ast_jair}, Accelerated Eval.~\cite{zhao2017accelerated}, STRIVE~\cite{rempe2022strive}, KING~\cite{hanselmann2022king}, AdvDiffuser~\cite{xie2024advdiffuser}, CAT~\cite{zhang2023cat}, AV-FUZZER~\cite{li2020avfuzzer}, CCDiff~\cite{lin2025causal}, CaDRE~\cite{huang2025cadre}, BridgeGen~\cite{hao2023bridging}, MADS~\cite{mads2024} \\

\bottomrule
\end{tabularx}
\par\vspace{3pt}
{\scriptsize \textit{Output Space} indicates the primary representation generated by each method family. Generative world models produce environment-level observations, while scenario generation methods produce simulation inputs (initial conditions, agent configurations, and trajectory specifications).}
\end{table*}

\subsection{Generative World Models}
\label{sec:worldmodels_sim}

Driving world models learn data-driven simulators of how the environment evolves under ego vehicle actions and optional control variables. A common formulation is an action-conditioned transition model:
\begin{equation}
    p_\theta\!\left(o_{t+1}\mid o_{\le t}, a_{\le t}, c\right)
\end{equation}
where $o_t$ is an observation (multi-camera images, LiDAR, BEV features, or learned latent), $o_{\le t} = \{o_1, o_2, \ldots, o_t\}$ denotes observation history up to time $t$, $a_t$ is an action or planned motion, $a_{\le t} = \{a_1, a_2, \ldots, a_t\}$ denotes action history, $c$ represents structured conditioning such as map context, agent states, or high-level commands, and $\theta$ represents model parameters. Modern driving world models often factorize generation through tokenizers and generative backbones (autoregressive next-token prediction or diffusion in latent space), which helps scale to long horizons and high-dimensional outputs.

World models function as general-purpose environment simulators that can serve multiple downstream applications. For behavior modeling specifically, generative world models enable three capabilities that traditional simulators lack: \emph{imagination-based training}, where behavioral policies can be optimized through rollouts inside the learned model without requiring access to the real environment or a hand-crafted simulator; \emph{closed-loop evaluation}, where learned behavior models are tested against environment responses that reflect data-driven dynamics rather than scripted rules; and \emph{data augmentation}, where the world model generates novel scenarios that expand the training distribution for behavior models beyond what is available in recorded logs.

\subsubsection{Video-Based and Latent World Models}
\label{sec:worldmodels_video_latent}

Video-based world models treat driving simulation as conditional video generation, producing future visual observations conditioned on actions and structured controls. Let $o_t \in \mathbb{R}^{C \times H \times W}$ denote a multi-view observation at time $t$. The observation is encoded into a latent representation $z_t \in \mathbb{R}^d$ or discrete tokens. In autoregressive form with discrete tokenization, let $z_t = \{z_{t,1}, z_{t,2}, \ldots, z_{t,K_t}\}$ represent $K_t$ tokens at time $t$ from vocabulary $\mathcal{V}$. The autoregressive world model learns:
\begin{equation}
    p_\theta(z_{1:T}) = \prod_{t=1}^{T}\prod_{k=1}^{K_t} p_\theta\!\left(z_{t,k}\mid z_{<t,\cdot}, z_{t,<k}, u\right)
\end{equation}
where $z_{<t,\cdot}$ denotes all tokens before time $t$, $z_{t,<k}$ denotes tokens at time $t$ before position $k$, and $u$ aggregates control variables (ego actions $a^{\text{ego}}$, map context $m$, semantic constraints $c$). In diffusion form, models denoise latent representations conditioned on past observations and controls. The key differentiator from occupancy-based models discussed in the next section is targeting view-level appearance and temporal coherence directly, which is useful for training and evaluating end-to-end autonomy stacks that operate on raw sensor inputs.

GAIA-1~\cite{hu2023gaia1} demonstrated that large-scale generative models can produce controllable scenario generation with emergent understanding of scene dynamics, geometry, and agent interactions. GAIA-2~\cite{russell2025gaia2} extends this foundation with latent diffusion and achieved multi-camera consistency, fine-grained control over agent configurations and environmental factors, and geographic diversity across multiple countries. GAIA-2  and its successor GAIA-3 enable the generation of safety-critical scenarios that are rare in naturalistic data but essential for behavior model evaluation. Drive-WM~\cite{wang2024drivewm} demonstrated that world models can directly serve planning by rolling out futures under different maneuvers and scoring with image-based rewards. This established a pathway from world model predictions to behavioral decisions. DriveDreamer~\cite{wang2024drivedreamer} proposed diffusion-based world models trained on real-world data with structured constraints (3D bounding boxes, HD maps, ego actions) to improve controllability. Its extension, DriveDreamer-2~\cite{zhao2025drivedreamer} added LLM interfaces converting natural language user intent into trajectories for customized driving video generation. Vista~\cite{gao2024vista} emphasized high fidelity and versatile controllability for generalizable driving world models by demonstrating long-horizon action-conditioned video prediction across diverse driving domains.

Additional architectural contributions address scalability and multimodal fusion. Copilot4D~\cite{zhang2023copilot4d} learns unsupervised world models by tokenizing LiDAR observations and predicting future tokens via discrete diffusion. MUVO~\cite{bogdoll2025muvo} bridges video-based and geometry-aware representations by fusing camera and LiDAR inputs to predict future observations in multiple formats (images, point clouds, 3D occupancy grids). It shows that multimodal sensor fusion improves both camera and LiDAR forecasting quality. DrivingWorld~\cite{hu2024drivingworld} introduces spatial-temporal fusion mechanisms for long-duration rollouts within GPT-style autoregressive frameworks, and GenAD~\cite{yang2024genad} proposes a generalized predictive model that rolls latent states forward to sample ego-conditioned futures, which enables end-to-end planning through world model imagination.

While these video-level world models produce visually compelling rollouts, their utility for behavior modeling depends on whether the generated dynamics are causally consistent under intervention. Evaluation remains a further challenge and interesting area to investigate. Standard image quality metrics (FID, FVD) do not measure driving-critical correctness, and standardized closed-loop benchmarks testing causal consistency under interventions are lacking. In addition, current models face long-horizon drift, where compounding errors accumulate over extended rollouts. Computational cost leads high-resolution multi-view generation become expensive for real-time closed-loop deployment. Hybrid supervision combining perceptual and dynamics losses, structured latent spaces separating geometry from appearance, and uncertainty-aware generation that flags unreliable rollouts could be considered for future research. A further open question is how to incorporate physical priors, such as vehicle dynamics, collision geometry, and road constraints, into video generation objectives to ensure rollouts are within physically valid space.

\subsubsection{Occupancy-Based World Models}
\label{sec:wm_occ}

Occupancy-based world models represent environment state using semantic occupancy fields (typically 3D voxel grids or BEV representations with height) and learn to predict future evolution conditioned on ego action. Let $O_t \in \{1,\dots,C\}^{H\times W\times Z}$ denote semantic occupancy at time $t$, where $H$, $W$, and $Z$ are spatial dimensions and $C$ is the number of semantic classes (road, vehicle, pedestrian, background). An occupancy world model learns:
\begin{equation}
    p_\theta\!\left(O_{t+1:t+T_f} \mid O_{1:t}, a_{t:t+T_f-1}\right)
\end{equation}
where $O_{1:t}$ represents occupancy history and $a_{t:t+T_f-1}$ denotes future ego actions over forecast horizon $T_f$. Compared with video world models, occupancy models trade pixel-level realism for geometric and semantic structure closer to planning cost functions, such as free space, drivable area, dynamic occupancy.

OccWorld~\cite{zheng2024occworld} learns scene tokenizers for 3D occupancy using GPT-like spatiotemporal transformers to autoregressively generate future scene tokens jointly with ego motion, which established the foundational autoregressive paradigm for occupancy prediction. The other paradigm, diffusion-based approaches, improve generation fidelity and controllability. OccSora~\cite{wang2024occsora} treats 4D occupancy generation as a core simulation primitive using diffusion-style generation for long sequences with semantic structure, and DOME~\cite{gu2024dome} uses continuous occupancy latents via an occupancy VAE with spatiotemporal diffusion transformers and trajectory-based resampling for strengthened controllability. Efficiency-focused architectures address the computational cost of dense 3D prediction. T$^3$Former~\cite{xu2025t3former} compresses 3D occupancy into triplanes and predicts triplane deltas autoregressively for real-time rollouts. $I^{2}$-World~\cite{liao2025i2} decouples tokenization into intra-frame and inter-frame components, which reduces redundancy. Drive-OccWorld~\cite{yang2025driving} intends to Connecting occupancy world models to planning and foundation-model interfaces. It adapts vision-centric 4D occupancy forecasting for end-to-end planning by evaluating candidate trajectories against occupancy-based cost functions.  OccLLaMA~\cite{wei2024occllama} introduces unified occupancy-language-action modeling for dynamics modeling, language conditioning, and multi-task outputs. It jointly tokenizes 3D occupancy, natural language instructions, and ego actions into a shared discrete token space, which enables a single model to simultaneously predict future scene states and plan actions. This unified token representation allows language commands to directly condition occupancy rollouts that bridges scene understanding and decision-making within a single generative framework.

For behavior modeling, occupancy world models offer a geometric interface that is particularly well suited for evaluating whether simulated agents maintain safe distances, avoid collisions, and respect road boundaries because there are assessments that are more natural in occupancy space than in pixel space. However, although occupancy captures geometry current approaches do not explicitly capture agent identity or intent. In other words, an occupied voxel does not carry information about which agent occupies it, what that agent's goal is, or how it would respond to interventions. This makes occupancy world models complementary to, but not a replacement for, the agent-level behavior models of Sections~\ref{sec:single_agent}--\ref{sec:multi_agent}. Additional challenges include compounding error in closed-loop rollouts, representation trade-offs (voxel grids are memory-intensive while tokenizers introduce compression loss), and expensive 4D occupancy labels. Future research could focus on semi-supervised training schemes that reduce label requirements, hybrid pipelines connecting occupancy rollouts to neural rendering for sensor-realistic evaluation, and explicit agent-identity tracking within occupancy representations.

\subsection{Traffic Scene and Scenario Generation}
\label{sec:scenario_gen}

While world models focus on learning environment dynamics, scene and scenario generation focus on constructing the traffic configurations, such as initial states and trajectory specifications. These methods address two complementary needs. First, populating simulation with diverse, realistic traffic scenes that let users steer generation toward specific configurations. Second, evaluating AV safety by generating and discovering rare, failure-inducing conditions. We organize this subsection accordingly, focusing on these two purposes. 

\subsubsection{Scene Initialization and Scenario Synthesis}
\label{sec:scenario_control_synthesis}

This section covers methods that generate traffic configurations, ranging from static scene layouts to full dynamic scenarios with explicit control interfaces. These methods cover a spectrum from unconditional scene initialization, which generates plausible starting conditions by sampling from learned data distributions, to controllable scenario synthesis, which exposes control variables that steer generated outcomes toward user-specified constraints while maintaining consistency with real-world traffic statistics.

\paragraph{Scene initialization and unconditional scenario generation.}
Scene initialization methods generate the starting conditions for simulation without producing subsequent dynamic behavior. SceneGen~\cite{tan2021scenegen} introduces autoregressive models that sequentially insert actors (with class, position, bounding box, and velocity) given ego state and HD map, which leads to generating realistic static traffic snapshots that can serve as initial conditions for downstream simulation or trajectory prediction models. Extending beyond static initialization, TrafficGen~\cite{feng2022trafficgen} generates both initial actor layouts and dynamic multi-agent trajectories via learning from fragmented real-world logs and supporting scenario augmentation through sampling from the learned distribution. DriveSceneGen~\cite{sun2024drivescenegen} further advances this direction by jointly synthesizing both static map elements and dynamic traffic participants from scratch, which reduces reliance on preselected map regions and enabling generation of entirely novel road geometries paired with plausible traffic. These methods generate diverse scenarios by sampling from learned data distributions but do not offer explicit user control over the generated outcomes, which is a limitation addressed by the controllable methods discussed next.

\paragraph{Controllable scenario synthesis.}
Controllable scenario synthesis learns generative distributions over scenarios by exposing \emph{control variables} that steer generated outcomes. Assume a scenario consisting of HD map $m$ and multi-agent trajectories $\tau = \{\tau^1, \ldots, \tau^N\}$, controllable synthesis learns the conditional distribution, $p_{\theta}(\tau \mid m, c)$, where $c$ encodes user intent or constraints, which can take multiple forms, such as semantic tokens, behavior tags, goal endpoints, interaction types, rule specifications, or natural language descriptions. Control is applied either by explicit conditioning during training or through guided sampling that biases generation toward satisfying constraints via auxiliary score functions.

In this class, CTG~\cite{zhong2023ctg} proposes conditional diffusion with sampling-time guidance using differentiable constraints including temporal-logic specifications, demonstrating that diffusion models can generate socially consistent multi-agent trajectories while offering controllability through differentiable cost function guidance. In practice, users may specify constraints such as collision avoidance or goal reaching, and the reverse diffusion process is steered toward satisfying them without retraining. CTG++~\cite{zhong2023ctgpp} extends this framework with language-based conditioning to translate natural language scenario descriptions into guidance signals that steer multi-agent trajectory generation. Scenario Diffusion~\cite{pronovost2023scenariodiffusion} frames controllability through conditioning tokens and map context using latent diffusion. These foundational controllable generation mechanisms support several safety-critical generation methods discussed in Section~\ref{sec:safety_critical_gen}, where the same architectural principles are specialized toward adversarial and failure-inducing objectives.

\paragraph{Language-conditioned and retrieval-augmented approaches.}
Language-conditioned approaches provide intuitive interfaces for scenario specification, bridging natural language intent and generated traffic behavior. LCTGen~\cite{tan2023lctgen}   uses natural language as supervision and combines LLMs with generative decoders for map selection, initial traffic distributions, and agent dynamics,. InteractTraj~\cite{xia2024language} translates language descriptions into structured codes conditioning trajectory generators. ChatSim~\cite{wei2024editable} operates at a different level of abstraction: rather than generating traffic behavior, it introduces collaborative LLM agents for editable photo-realistic 3D scene rendering via neural radiance fields, which enables language-commanded modification of the visual appearance of driving scenes (for instance, adding, removing, or repositioning rendered vehicles). While ChatSim's controllability interface is relevant to scenario specification, its primary contribution lies in sensor-level scene editing rather than traffic behavior synthesis. RealGen~\cite{ding2024realgen} introduces retrieval-augmented generation retrieving template scenarios, including rare tagged cases for compositional behavior synthesis.

A key concern in scene and scenario generation methods is the progression from unconditional to controllable generation because stronger constraints push samples off the data manifold. Moreover, evaluation metrics and pipelines that measure both constraint satisfaction and realism are lacking. For scene initialization methods, ensuring that generated starting conditions lead to realistic spawn of vehicles and plausible dynamic evolution when paired with downstream behavior models is still an open problem. Future research directions include compositional control building scenarios from reusable primitives, hybrid retrieval-plus-generation systems targeting rare events while producing novel variations, tighter coupling between scenario synthesis and downstream evaluation objectives, and unified frameworks that jointly generate scene layouts, initial conditions, and controllable dynamic trajectories within a single generative pipeline. Also, verification against traffic flow theory foundations seems essential for ensuring the realism of generated scenes and scenarios. 

\subsubsection{Safety-Critical Scenario Generation and Adversarial Falsification}
\label{sec:safety_critical_gen}

While the methods of Section~\ref{sec:scenario_control_synthesis} generate diverse and controllable traffic scenarios for general-purpose simulation, a particularly important sub-problem is the efficient discovery of \emph{safety-critical} scenarios, such as near-misses, or edge cases. This section surveys methods that address this sub-problem through two complementary strategies: \emph{optimization-based adversarial search}, which uses reinforcement learning, importance sampling, or fuzzing to search the scenario space for failure-inducing conditions; and \emph{safety-targeted generative synthesis}, which specializes the generative models introduced in Section~\ref{sec:scenario_control_synthesis} by directing their sampling toward safety-critical regions of the scenario distribution. The central challenge shared across both strategies is balancing \emph{criticality} (finding scenarios that expose failures) with \emph{validity} (ensuring those scenarios remain physically plausible and represent situations that could actually occur).

It is important to distinguish the purpose of these methods from the adversarial imitation methods of Section~\ref{sec:irl_gail} and multi-agent adversarial learning of Section~\ref{subsec:decentralized_methods}: adversarial imitation learns realistic agent policies by matching expert distributions, whereas the methods here use optimization to \emph{test} those policies by searching for failure-inducing conditions. The generated scenarios serve as evaluation infrastructure rather than as behavior models themselves.

\paragraph{Optimization-based adversarial search.}
A first family of methods treats safety-critical scenario discovery as explicit optimization. 
% Given simulator rollout function $\textsc{Sim}(\omega; \pi)$ that simulates scenario $\omega \in \Omega$ with AV policy $\pi$, producing trajectory $\tau$, and safety requirement $\varphi$, the adversarial optimization problem seeks:
% \begin{equation}
%     \omega^\star \in \arg\max_{\omega \in \Omega_{\text{valid}}} \; \mathbb{E}_{\tau \sim \textsc{Sim}(\omega;\pi)}\!\left[J(\tau)\right]
% \end{equation}
% where $\Omega_{\text{valid}} \subseteq \Omega$ encodes physical plausibility and traffic feasibility constraints, and $J(\tau)$ is a scalar criticality score measuring violation severity. A common instantiation uses robustness of temporal-logic specifications, where falsification corresponds to finding scenarios minimizing robustness $\rho(\tau, \varphi)$, with $\rho(\tau, \varphi) < 0$ indicating violation.
Adaptive Stress Testing (AST)~\cite{koren2018ast_iv,lee2020ast_jair} casts failure search as MDPs where adversaries select disturbance actions that drive the system toward failure while keeping trajectories probable under nominal disturbance models. Accelerated evaluation methods~\cite{zhao2017accelerated} modify surrounding vehicle behavior and use importance sampling for statistically meaningful risk estimates with far fewer simulation miles than na\"ive Monte Carlo. Multi-agent adversarial RL extends this by training adversarial agents that exploit AV policy weaknesses~\cite{sharif2022adversarial,kuutti2020training_adversarial}, while hierarchical RL architectures manage reward sparsity and promote diverse scenario generation~\cite{zhu2025critical}. Closed-loop Adversarial Training (CAT)~\cite{zhang2023cat} iterates between safety-critical scenario generation and driving policy optimization using probabilistic factorization, and through this, turn the log-replay into adversarial scenario. Fuzzing frameworks provide a complementary approach, mutating scenario parameters guided by surrogate scores. AV-FUZZER~\cite{li2020avfuzzer} uses genetic algorithms with learned fitness functions, DriveFuzz~\cite{kim2022drivefuzz} combines mutation-based fuzzing with neural guidance, and TM-fuzzer~\cite{lin2024tmfuzzer} targets traffic management scenarios with learned surrogate models.

\paragraph{Safety-targeted generative synthesis.}
A second family specializes the generative scenario synthesis frameworks of Section~\ref{sec:scenario_control_synthesis} by directing their sampling toward safety-critical regions. These methods leverage learned traffic priors to ensure that adversarial scenarios remain realistic rather than drifting into implausible configurations.

STRIVE~\cite{rempe2022strive} uses VAE-based traffic models for gradient-based optimization in latent space. The learned prior constrains the search to realistic behaviors even as the objective pushes toward dangerous configurations. KING~\cite{hanselmann2022king} uses proxy dynamics and kinematics gradients for efficient adversarial perturbation of recorded scenarios. Hao et al.~\cite{hao2024adversarial_naturalistic} combine GAIL-based driver representations with PPO-based optimization to balance adversariality and naturalness. AdvDiffuser~\cite{xie2024advdiffuser} builds on the guided diffusion foundations established by CTG (Section~\ref{sec:scenario_control_synthesis}) and replaced general-purpose controllability objectives with auxiliary reward functions that steer the diffusion sampling process specifically toward safety-critical scenarios. CCDiff~\cite{lin2025causal} similarly extends the controllable diffusion paradigm and targeted long-tail and safety-critical settings by injecting causal structure and decomposition into the generation process for closed-loop evaluation. CaDRE~\cite{huang2025cadre} emphasizes controllability plus diversity for safety-critical generation, specifically addressing mode collapse around single failure patterns to ensure a broad coverage of the safety-critical scenario space. BridgeGen~\cite{hao2023bridging} bridges data-driven and knowledge-driven approaches by combining statistical patterns from driving data with explicit domain knowledge about safety-critical situations to generate scenarios that are both realistic and targeted. For modeling hazardous interactions in mixed autonomy traffic, specifically, MADS~\cite{mads2024} develops a multi-agent driving simulation framework designed to characterize safety-critical encounters between automated and manual vehicles, which allows systematic evaluation of AV safety under realistic adversarial conditions involving human-driven vehicles.

The validity-criticality trade-off remains the central challenge for safety-critical scenario generation methods. Adversarial search and safety-targeted generation can drift into unrealistic behaviors unless constrained by strong priors. On the other hand, overly conservative validity constraints may prevent discovery of genuine edge cases. Coverage accounting is a further concern. Finding many failures does not necessarily imply meaningful operational domain coverage, and the relationship between generated scenario distributions and real-world occurrence frequencies requires statistical frameworks that connect failure finding to principled risk estimation \cite{rahmani2024systematic}. For safety-targeted generative methods, the challenge of ensuring that guided sampling produces \emph{causally valid} agent reactions, and not just statistically plausible trajectories that happen to be dangerous, is shared with the counterfactual validity concerns discussed in Section~\ref{sec:neural_sim_challenges}. Worthwhile directions for investigation include tighter integration between optimization-based search and generative priors (using learned models to constrain the feasible search space while optimization targets criticality), likelihood-aware adversarial methods that yield unbiased risk estimates rather than simply finding failures, multi-objective frameworks that simultaneously optimize for criticality, diversity, and validity, and standardized evaluation protocols that measure both the realism and the coverage of generated safety-critical scenario sets.

\subsection{Discussions on Environment-level Simulations}
\label{sec:neural_sim_challenges}

Neural simulation environments promise advances in realism and controllability by learning environment dynamics through generative world models and scenario inputs through data-driven synthesis. However, deploying these components as reliable infrastructure for behavior modeling raises challenges distinct from those faced by agent-level methods.

A fundamental challenge is \textbf{counterfactual validity under intervention}. Neural simulators must remain correct when ego policy changes, agents are edited, or scene conditions are modified. This is inherently difficult because most training data is observational and does not uniquely identify counterfactual reactions. World models can produce visually convincing rollouts that respond incorrectly when the ego vehicle deviates from logged behavior. Similarly, in scenario generation, adversarial search can exploit simulator weaknesses, producing unrealistic high-criticality failures unless validity constraints are strong and calibrated.

\textbf{Long-horizon stability and compounding error} presents a second challenge. Closed-loop testing requires rollouts over many seconds with repeated feedback. Generative models suffer from drift where small early errors move simulated distributions out-of-support, which produces increasingly implausible states, particularly acute for high-dimensional representations and in multi-agent settings where interaction feedback creates coupling between agents' errors. \textbf{The gap between environment fidelity and behavioral realism} is specific to the environment-level perspective. A world model may generate visually or geometrically plausible scenes whose \emph{behavioral content} (how agents move and interact) does not match the quality of purpose-built agent models. Conversely, sophisticated agent-level behavior models may be undermined by environment models that provide unrealistic feedback. Aligning environment-level and agent-level fidelity is essential but rarely addressed explicitly. A further gap is \textbf{risk estimation}. Adversarial failure finding differs from estimating real-world risk, which necessitates statistical frameworks connecting generated scenario distributions to real-world occurrence frequencies.

Promising future directions include: (1) \emph{hybrid simulators} combining explicit structure (maps, kinematics, traffic rules) with learned residual generation; (2) \emph{uncertainty-aware simulation} exposing confidence estimates and supporting conservative rollout when out-of-distribution; (3) \emph{structured controllability interfaces} aligned with engineering workflows; (4) \emph{causal and interventional training objectives} going beyond matching observational distributions to explicitly model counterfactual responses; and (5) \emph{standardized evaluation protocols} that assess both environmental realism and behavioral fidelity of generated scenarios.

\section{COGNITIVE AND PHYSICS INFORMED AI METHODS}
\label{sec:cognitive_ai}

The data-driven methods surveyed in Sections~\ref{sec:single_agent} to \ref{sec:environment_level} optimize for trajectory accuracy against recorded behavior. However, they often cannot explain \textit{why} human drivers behave as they do. A model that matches logged trajectories with low displacement error may still fail to reproduce the cognitive mechanisms that generate those trajectories. This distinction matters for mixed autonomy simulation in at least three ways. First, purely data-driven agents cannot predict human \textit{error modes}, such as inattention, misjudgment, and overload failures that dominate real-world crash causation. Second, without cognitive grounding, behavioral diversity in simulation relies on arbitrary latent variable sampling (Section~\ref{subsec:decentralized_methods}) rather than on principled variation of psychologically meaningful parameters such as reaction time, risk tolerance, or attentional capacity. Third, as mixed autonomy traffic introduces novel interaction patterns between humans and AVs (patterns absent from historical driving logs) cognitively grounded models offer a basis for extrapolation and generalization that distribution-matching methods lack.

This section reviews methods that incorporate validated theories from cognitive psychology, human factors, and traffic physics into learning-based behavior models. We organize the discussion around three complementary perspectives: how drivers \textit{decide} under uncertainty (Section~\ref{subsec:bounded_rationality}), how they \textit{perceive}, \textit{attend} to, and \textit{evaluate} the driving environment including risk assessment and trust toward other agents (Section~\ref{subsec:cognitive_attention}), and how \textit{physical priors} can be encoded into neural architectures to enforce dynamic plausibility (Section~\ref{subsec:physics_informed}). We conclude with a discussion of integration challenges and promising directions for bridging cognitive and data-driven paradigms (Section~\ref{subsec:cognitive_integration}). Table~\ref{tab:cognitive_ai_summary} provides a summary of the reviewed methods and studies in this section.

\begin{table*}[!htbp]
\centering
\caption{Summary of cognitive and physics-informed AI methods for mixed autonomy traffic simulation.}
\label{tab:cognitive_ai_summary}
\renewcommand{\arraystretch}{1.25}
\setlength{\tabcolsep}{3pt}
\scriptsize
\begin{tabularx}{\textwidth}{
    >{\RaggedRight\arraybackslash}p{1.7cm}
    >{\RaggedRight\arraybackslash}p{1.6cm}
    >{\RaggedRight\arraybackslash}X
    >{\RaggedRight\arraybackslash}X
    >{\RaggedRight\arraybackslash}X
    >{\RaggedRight\arraybackslash}p{2.8cm}
}
\toprule
\textbf{Method} &
\textbf{Theory Basis} &
\textbf{Core Idea} &
\textbf{Strengths} &
\textbf{Limitations} &
\textbf{Representative Studies} \\

%% --- SUB-HEADER: Bounded Rationality & Decision-Making ---
\midrule
\multicolumn{6}{l}{\textbf{\textit{Bounded Rationality and Decision-Making Models}}} \\
\midrule

Prospect-Theoretic Decision Models &
Cumulative Prospect Theory (CPT); Prospect Balancing &
Replaces expected utility with S-shaped value function and nonlinear probability weighting; captures loss aversion and overweighting of rare catastrophic outcomes in driving decisions &
Interpretable style variation via CPT parameters; reproduces conservative/aggressive profiles without separate rewards; psychologically validated &
Parameter calibration from driving data is difficult; limited to single-decision settings; does not model temporal dynamics of choices &
Tversky \& Kahneman~\cite{tversky1992advances}, Schmidt et al.~\cite{schmidt2014prospect}, Sun et al.~\cite{sun2019interpretable} \\

Risk Allostasis \& Homeostatic Models &
Risk Allostasis Theory; Task-Capability Interface &
Drivers adjust behavior to maintain a subjective target risk level rather than optimizing utility; explains systematic deviations from time-optimal behavior &
High fidelity to naturalistic car-following and speed choice; captures risk compensation effects; simple parameterization &
Target risk level is latent and hard to estimate; limited to longitudinal control; does not explain strategic interaction &
Fuller~\cite{fuller2011driver}, Mohammadian et al.~\cite{mohammadian2023net}, Kashifi~\cite{kashifi2025modelling}  \\

Drift-Diffusion Models (DDMs) &
Sequential Sampling; Evidence Accumulation &
Decision variable accumulates noisy perceptual evidence toward boundaries; models the temporal dynamics and variability of gap acceptance, braking, and yielding decisions &
Captures full RT distribution, including dangerous late decisions; perceptual inputs (looming, TTC) as natural drift rates; differentiable neural implementations &
Binary decisions only; limited multi-alternative extensions; requires careful specification of evidence signals; calibration from naturalistic data is challenging &
Ratcliff~\cite{ratcliff1978theory}, Markkula et al.~\cite{markkula2018models}, Zgonnikov et al.~\cite{zgonnikov2022should}, McDonald et al.~\cite{mcdonald2021modeling}, Fengler et al.~\cite{fengler2021neuralddm} \\

Level-$k$ Reasoning &
Cognitive Hierarchy; Behavioral Game Theory &
Agents have heterogeneous strategic depth: level-0 follows heuristics, level-$k$ best-responds to level-$(k{-}1)$; models realistic variation in anticipatory reasoning &
Principled heterogeneity in interactive behavior; avoids full-rationality assumption; calibratable from observed interactions &
Requires specifying level-0 heuristic; distribution over levels is scenario-dependent; computational cost grows with $k$ &
Stahl \& Wilson~\cite{stahl1995levels}, Chong et al.~\cite{chong2005cognitive}, Li et al.~\cite{li2017levelk} \\

%% --- SUB-HEADER: Cognitive Architectures, Attention, Risk Perception & Trust ---
\midrule
\multicolumn{6}{l}{\textbf{\textit{Cognitive Architectures, Attention, Risk Perception, and Trust Models}}} \\
\midrule

Cognitive Architectures (ACT-R, SOAR) &
ACT-R; SOAR; Production Systems &
Unified computational theories of perception, memory, and motor execution with constrained processing resources; predict realistic latencies and error modes from cognitive bottlenecks &
Interpretable parameters with psychological meaning; predict error modes (inattention, overload); principled generalization to novel scenarios; validated against human data &
Hand-crafted production rules; operate in simplified environments; difficult to scale to high-fidelity scenarios; limited continuous control &
Anderson~\cite{anderson2004integrated}, Salvucci~\cite{salvucci2006modeling}, Cao et al.~\cite{cao2014qnactr}, Ebadi et al.~\cite{ebadi2024qnactrsa}, Laird~\cite{laird2012soar}, Zhou et al.~\cite{zhou2025hybrid} \\

Attention \& Gaze Prediction &
Selective Attention; Visual Saliency; Foveal--Peripheral Processing &
Models where human drivers look using learned gaze maps combining bottom-up saliency and top-down task cues; weights features by predicted attention for downstream models &
Prevents causal confusion; provides explainability; captures perceptual biases; large-scale gaze datasets available &
Gaze $\neq$ attention (covert shifts missed); dataset bias toward normal driving; limited integration with closed-loop control &
Palazzi et al.~\cite{palazzi2018dreyeve}, Xia et al.~\cite{xia2018predicting}, Fang et al.~\cite{fang2021dada2000}, Kultrera et al.~\cite{kultrera2020attention} \\

Brain-Inspired \& Affordance Models &
Ecological Perception; Multiple Resource Theory; Cognitive Control Hypothesis &
Spiking neural networks mimic visual cortex; affordance-based representations extract driving-relevant variables; resource theory predicts dual-task interference and distraction effects &
Robust to high-speed/low-latency scenarios; interpretable intermediate affordances; validated dual-task degradation predictions ($R^2 > 0.90$) &
SNNs lack mature training frameworks; affordance definition requires domain expertise; resource models calibrated from lab, not naturalistic settings &
Ma et al.~\cite{ma2021traffic}, Chen et al.~\cite{chen2015deepdriving}, Wickens~\cite{wickens2008mrt}, Horrey \& Wickens~\cite{horrey2006wickens}, Engstr\"{o}m et al.~\cite{engstrom2017cch} \\

Risk Fields \& Potential-Based Models &
Driving Safety Field; Artificial Potential Fields &
Surrounding objects and road elements emit repulsive potentials; trajectories minimize cumulative risk exposure; learned risk fields capture subjective risk perception differing from objective TTC &
Implicit safety guarantees; reduced data requirements; captures subjective risk biases (overestimation of head-on, underestimation of lateral); interpretable field visualization &
Risk field calibration is driver-specific; limited to reactive control; does not model strategic anticipation; field superposition assumptions may be simplistic &
Wang et al.~\cite{wang2015driving}, Ji et al.~\cite{ji2020navigating}, Kolekar et al.~\cite{kolekar2020drf} \\

Trust Dynamics in Human--AV Interaction &
Trust Calibration; Three-Layer Trust Model &
Models how human drivers adjust behavior based on evolving trust toward AVs; captures dispositional, situational, and learned trust influencing headway, gap acceptance, and interaction willingness &
Essential for mixed autonomy transition modeling; ML-based prediction from physiological and behavioral features; captures asymmetric trust repair after failures &
Trust is latent and difficult to measure at scale; models calibrated from simulator studies, not naturalistic driving; limited integration with trajectory-level behavior models &
Lee \& See~\cite{lee2004trust}, Hoff \& Bashir~\cite{hoff2015trust}, Ayoub et al.~\cite{ayoub2021trust}, Kaufman et al.~\cite{kaufman2025predicting} \\

%% --- SUB-HEADER: Physics-Informed Learning ---
\midrule
\multicolumn{6}{l}{\textbf{\textit{Physics-Informed Learning}}} \\
\midrule

Physics-Informed Deep Learning (PIDL) &
IDM; OVM; FVDM; Car-Following Theory &
Encodes classical traffic flow models as structural priors or regularization in neural networks; combines physics rigor with data-driven flexibility for car-following and trajectory prediction &
Outperforms pure physics and pure neural models; strong in sparse-data regimes; enforces kinematic plausibility; distributional estimates via PIDL-GAN &
Limited to longitudinal models (car-following); physics priors may be too rigid for complex urban scenarios; fusion strategy selection is ad hoc &
Mo et al.~\cite{mo2021pidl}, Mo et al.~\cite{mo2022uncertainty}, Xu et al.~\cite{xu2022car}, Geng et al.~\cite{geng2023physics}, Wang et al.~\cite{wang2025knowledge} \\

\bottomrule
\end{tabularx}
\par\vspace{3pt}
{\scriptsize \textit{Theory Basis} indicates the primary cognitive, psychological, or physical theory grounding each method family. Methods are organized by whether they address decision-making under uncertainty, cognitive processing and evaluation including risk perception and trust, or physical and kinematic constraints.}
\end{table*}

\subsection{Bounded Rationality and Decision-Making Models}
\label{subsec:bounded_rationality}

Standard reinforcement learning (Section~\ref{sec:rl}) and game-theoretic models (Section~\ref{subsec:decentralized_methods}) assume agents that maximize expected utility. However, behavioral research demonstrate that human decision-making under uncertainty may be governed by bounded rationality~\cite{simon1955behavioral}, which means drivers use heuristics, overweight rare catastrophic outcomes, exhibit loss aversion, and make choices that are \textit{satisfactory} rather than optimal. Incorporating these biases into behavior models produces agents whose decisions are not merely statistically plausible but psychologically supported.

Cumulative Prospect Theory (CPT)~\cite{tversky1992advances} provides a formal framework for bounded rationality. CPT replaces the linear utility and probability weighting of expected utility theory with an S-shaped value function, and a probability weighting function that overweights small probabilities and underweights large ones. Applied to driving, this means that drivers disproportionately weigh the small probability of a catastrophic collision relative to the certain time cost of driving slowly. Sun et al.~\cite{sun2019interpretable} show that the CPT-based formulation yields an interpretable explanation of behaviors such as cautious versus risk-seeking responses during interaction, and it can reproduce different driving tendencies by tuning CPT parameters. Risk allostasis theory~\cite{fuller2011driver} and its computational instantiations~\cite{mohammadian2023net, kashifi2025modelling} provide a complementary perspective. They suggest that rather than optimizing any utility function, drivers continuously adjust their behavior to maintain a subjective target risk level. Models incorporating this homeostatic mechanism yield higher fidelity to naturalistic driving data than efficiency-maximizing algorithms. Prospect-balancing theory~\cite{schmidt2014prospect} further formalizes this insight by conceptualizing speed choice as a trade-off between the prospect of arriving sooner and the prospect of avoiding a collision, with subjective probability distortions favoring cautious choices.

For the dynamic and timely decisions characteristic of driving, such as gap acceptance at intersections, merge initiation, and pedestrian yielding, drift-diffusion models (DDMs) from cognitive psychology offer a process-level account of how decisions unfold over time~\cite{ratcliff1978theory}. In DDMs, a decision variable accumulates noisy evidence toward one of two boundaries (e.g., ``accept gap'' versus ``wait''). Markkula et al.~\cite{markkula2018models} demonstrated that variable-drift DDMs accurately capture not just mean response times but the \textit{full distribution} of response times in gap acceptance scenarios. This distributional accuracy is critical for safety validation, where the probability of rare late responses determines crash risk. Subsequent work extended DDMs to looming-based evidence accumulation for braking decisions~\cite{zgonnikov2022should,mcdonald2021modeling}. They showed that perceptual variables (optical expansion rate, time-to-collision) serve as natural drift-rate inputs. Neural implementations of evidence accumulation~\cite{fengler2021neuralddm} enable end-to-end learning of accumulation dynamics from behavioral data, which bridges the gap between cognitive process models and scalable deep learning architectures.

The level-$k$ reasoning framework introduced in Section~\ref{subsec:decentralized_methods} for game-theoretic planning has deep roots in cognitive science~\cite{stahl1995levels,chong2005cognitive}. Its relevance here is that it provides a principled cognitive model of \textit{strategic depth heterogeneity}: some drivers plan several steps ahead (high-$k$ reasoners anticipating others' responses), while others react to immediate stimuli (level-0 heuristic followers). Calibrating the distribution of reasoning levels from naturalistic data could produce traffic populations with realistic heterogeneity in interactive scenarios.

For simulation, the key advantage of bounded rationality models is that behavioral diversity emerges from variation in \textit{interpretable cognitive parameters}, such as risk sensitivity, evidence accumulation rate, decision boundary, reasoning depth, rather than from obscured latent variables. This enables test engineers to construct specific driver profiles (e.g., ``distracted driver with slow evidence accumulation and high risk tolerance'') grounded in psychological foundations, which could support targeted scenario design for safety-critical evaluation.

\subsection{Cognitive Architectures and Attention-Guided Models}
\label{subsec:cognitive_attention}

While bounded rationality models address the decision-making process itself (how drivers evaluate options, accumulate evidence, and select actions) this subsection covers the perceptual and cognitive processing from which those decisions arise, such as the processing bottlenecks and capacity limitations that cause errors. It also covers two closely related aspects of cognitive evaluation: how drivers assess risk through subjective threat perception, and how they calibrate trust when interacting with automated vehicles.

\paragraph{Cognitive architectures.}
Cognitive architectures provide unified computational theories of human information processing. ACT-R (Adaptive Control of Thought-Rational)~\cite{anderson2004integrated} has become one of the most extensively validated architectures for driving. Salvucci's foundational work~\cite{salvucci2006modeling} established a comprehensive driver model based on ACT-R by integrating control (steering and speed regulation), monitoring (situation assessment through visual sampling), and decision-making (tactical choices such as lane changes). Later, QN-ACTR~\cite{cao2014qnactr} integrated queueing network theory to model multitask performance and predict how secondary tasks degrade driving, and QN-ACTR-SA~\cite{ebadi2024qnactrsa} extended situation awareness modeling, which resulted in better performance in predicting brake response times. SOAR-ACPPO~\cite{zhou2025hybrid} combined a rule-based cognitive control~\cite{laird2012soar} with deep reinforcement learning for lane change decisions. It uses SOAR to structure the decision space and RL to optimize within it. This hybrid paradigm addresses the problem of hand-crafting production rules for complex continuous control while preserving the interpretability and cognitive fidelity that pure RL lacks.

For simulation, cognitive architectures offer three advantages that neural networks alone cannot provide. First, their parameters have direct psychological meaning (e.g., visual sampling frequency, memory decay rate, production rule latency). This property allows principled transfer across scenarios and populations. Second, they predict \textit{error modes}, the specific failures arising from cognitive overload, inattention, or memory interference. This is important because these are the human behaviors that AVs must handle safely. Third, they support principled generalization to novel scenarios that lack training data, because behavior is generated from cognitive mechanisms rather than pattern-matched to observed distributions.

\paragraph{Attention and perception models.}
Human drivers do not process visual scenes uniformly: foveal vision captures detail in a narrow central region while peripheral vision detects motion and salience, and selective attention determines which scene elements receive processing resources~\cite{carrasco2011visual}. This stands in contrast to the uniform feature extraction assumed by most neural driving models, which process entire bird's-eye-view representations or multi-camera inputs with equal weight. Large-scale driving attention datasets, such as DR(eye)VE~\cite{palazzi2018dreyeve} (74 sequences with eye-tracking and physiological signals), BDD-A~\cite{xia2018predicting} (over 1,200 videos with gaze annotations), and DADA-2000~\cite{fang2021dada2000} (2,000 accident videos with driver attention labels), could help with training attention prediction networks that model where human drivers actually look.

Xia et al.~\cite{xia2018predicting} fused bottom-up visual saliency (scene-driven conspicuity) with top-down task-driven cues (goal-directed attention allocation) by using predicted gaze maps to weight feature importance in downstream driving models. This attention-guided weighting prevents \textit{causal confusion}~\cite{codevilla2019exploring} where end-to-end models learn spurious correlations (e.g., associating stopped vehicles with red traffic lights in the background rather than with the vehicle itself). Kultrera et al.~\cite{kultrera2020attention} demonstrated that attention-based architectures provide explainability for end-to-end steering and help verify whether a model's failures correspond to plausible human perceptual limitations.

Brain-inspired architectures push this further. DeepDriving~\cite{chen2015deepdriving} operationalized Gibson's ecological theory of affordances and trained CNNs to estimate driving-relevant affordances (distance to lane markings, angles to surrounding vehicles) as intermediate representations rather than mapping directly from images to control. The aim was to produce more interpretable and transferable behavior models. Wickens' Multiple Resource Theory~\cite{wickens2008mrt} provides a complementary framework for predicting dual-task interference. When secondary tasks (phone use, conversation, navigation) compete for the same perceptual or cognitive resources as driving, performance degrades predictably. Horrey and Wickens~\cite{horrey2006wickens} could predict driver attention allocation under dual-task conditions with high accuracy, and Engstr\"{o}m et al.'s Cognitive Control Hypothesis~\cite{engstrom2017cch} refined this by showing that cognitive load selectively impairs executive-dependent sub-tasks, such as hazard detection and tactical decisions, while preserving automatized behaviors like lane-keeping and speed maintenance. Collectively, these brain-inspired and resource-theoretic approaches demonstrate that encoding the structure of human cognition and attention into computational models yields not only more faithful representations of driver behavior but also principled predictions of the context-dependent failure modes that autonomous vehicles must be validated against.

\paragraph{Risk fields and potential-based models.}
The Driving Safety Field (DSF) theory~\cite{wang2015driving} conceptualizes driving behavior as navigation through subjective risk fields generated by surrounding objects, road boundaries, and traffic infrastructure. Each obstacle and road element emits a repulsive potential that decays with distance and depends on relative kinematics, and the driver's trajectory minimizes cumulative exposure to the resulting composite field. Deep learning extensions learn to map sensor inputs to continuous potential surfaces, with ego-vehicles performing gradient descent on learned risk. Building on this, recent approaches employ graph representations paired with deep reinforcement learning to navigate these fields. By modeling traffic participants and road topology as interconnected graphs, these DRL agents can efficiently process complex, multi-agent potential fields to learn safe navigation policies \cite{ji2020navigating}. This formulation provides implicit safety guarantees while drastically reducing data requirements compared to unconstrained end-to-end learning. Kolekar et al.~\cite{kolekar2020drf} introduced the Driver's Risk Field (DRF), a learned two-dimensional field representing \textit{subjective} collision probability that varies across drivers and scenarios. The DRF captures the empirical finding that drivers' perceived risk differs systematically from objective risk metrics such as TTC. Drivers overestimate risk from vehicles approaching on a collision course while underestimating risk from vehicles in adjacent lanes.

\paragraph{Trust and adaptation in human--AV interaction.}
For mixed autonomy traffic, an important behavioral dimension is \textit{trust}: how human drivers calibrate their expectations and adjust their behavior when interacting with automated vehicles whose driving patterns may differ from familiar human norms. Lee and See~\cite{lee2004trust} establish three trust dimensions: performance (competence-based), process (understanding-based), and purpose (intent-based). Their framework explains why the same driver may follow an AV closely after observing smooth driving but maintain excessive headway after witnessing an abrupt stop, and why trust recovery after automation failures is slow and asymmetric. In a similar work, Hoff and Bashir~\cite{hoff2015trust} distinguish dispositional trust (personality-based baseline), situational trust (context-dependent adjustment), and learned trust (experience-based calibration). 

Machine learning approaches increasingly enable data-driven trust modeling. Ayoub et al.~\cite{ayoub2021trust} used XGBoost with SHAP explanations to predict trust from driving context and behavioral indicators, which provided accuracy and interpretability regarding which factors most influence trust dynamics. Kaufman et al.~\cite{kaufman2025predicting} achieved high accuracy in classifying trust levels using random forests trained on physiological and behavioral features. For simulation, trust-aware agent models are useful for representing the transition period of mixed autonomy deployment. As AV penetration rates increase, human drivers gradually calibrate their interaction strategies. Integrating trust dynamics into the hybrid data-driven agents discussed in Section~\ref{subsec:decentralized_methods} represents a promising direction for generating realistic mixed traffic populations.

\subsection{Physics-Informed Learning}
\label{subsec:physics_informed}

A complementary approach to cognitive grounding is encoding physical laws and traffic flow theory directly into neural architectures. While cognitive models and bounded rationality methods constrain how agents process information and make decisions, physics-informed methods constrain what behaviors are physically and dynamically plausible. This addresses the well-known problem that purely data-driven models can generate trajectories violating basic kinematics, producing negative headways, or ignoring road geometry.

\paragraph{Physics-informed deep learning for traffic behavior.}
The physics-informed deep learning (PIDL) paradigm, introduced by Mo et al.~\cite{mo2021pidl}, encodes established car-following models, including Intelligent Driver Model (IDM), Optimal Velocity Model (OVM), and Full Velocity Difference Model (FVDM), directly into neural network architectures as structural priors or regularization terms. Rather than learning car-following behavior from scratch and with no prior knowledge from data, PIDL networks are constrained to produce outputs consistent with the functional forms of these classical models while retaining the flexibility to capture residual complexity that physics alone misses. They demonstrate that physics-informed networks substantially outperform both pure physics models and pure neural networks, particularly in sparse-data regimes where training samples are limited. Xu et al.~\cite{xu2022car} extended this paradigm with improved IDM formulations and complementary fusion strategies. Geng et al. proposed a physics-informed Transformer architecture (PIT-IDM)~\cite{geng2023physics} by combining the long-range temporal modeling of self-attention with IDM-based structural constraints for highway trajectory prediction and achieved improvements over both physics-uninformed transformers and standalone IDM. Mo et al.~\cite{mo2022uncertainty} developed a physics-informed generative adversarial network for uncertainty quantification in car-following. They produced point predictions and calibrated distributional estimates of acceleration conditioned on traffic regime. Most recently, the Knowledge-Informed Deep Learning (KIDL) paradigm~\cite{wang2025knowledge} distills high-level driving knowledge from large language models into lightweight neural car-following models, which combines LLM-derived priors about traffic regulations and driving norms with data-driven calibration for improved cross-dataset generalization.

\subsection{Integration Challenges and Future Directions}
\label{subsec:cognitive_integration}

The methods reviewed above offer complementary benefits. Bounded rationality models explain decision variability through psychologically meaningful parameters, cognitive architectures predict processing limitations and error modes, attention models capture perceptual biases, risk fields represent subjective threat evaluation, trust models represent adaptation in human--AV interaction, and physics-informed methods enforce dynamic plausibility. However, each operates at a different level of abstraction and computational cost, and integrating them into cohesive, scalable simulation agents remains an open challenge. A key challenge is the trade-off between \textit{cognitive validity} and \textit{data-driven scalability}. Cognitive architectures such as ACT-R produce interpretable, theory-grounded behavior but require hand-crafted production rules and operate in simplified environments. Conversely, data-driven models scale to complex scenarios but lack cognitive foundations, which results in agents that match trajectory distributions without capturing the generative mechanisms behind them. 

Several promising integration strategies could be considered. First, instead of building full cognitive architectures, key cognitive constraints (processing delays, attention bottlenecks, bounded lookahead) can be encoded as structural priors or regularization terms in neural network training, which is analogous to how PIDL encodes physics. Evidence accumulation dynamics, for instance, can be implemented as differentiable neural layers~\cite{fengler2021neuralddm} within end-to-end architectures. This combines DDM-like decision processes with learned feature extraction. Second, LLM-enhanced cognitive modeling seems to be a promising direction~\cite{gao2026foundation}. Wu et al.~\cite{wu2024llmactr} explored using large language models to assist in developing and extending cognitive architectures, leveraging LLMs' broad knowledge to generate production rules and parameter hypotheses that would otherwise require extensive manual effort. Third, cognitive architectures can generate synthetic behavioral data, such as rare error modes and edge cases underrepresented in naturalistic datasets, to augment training sets for neural models. It has been demonstrated by recent work using cognitive-based driver models to produce critical scenarios for AV testing~\cite{jawad2024accident}. Fourth, combining physics-informed constraints (kinematic feasibility, road compliance) with cognitive constraints (bounded reaction time, attentional capacity) within a single training framework would produce agents that are simultaneously physically plausible and cognitively realistic. 

A further open challenge is \textit{calibration at scale}. Cognitive model parameters (risk sensitivity, attention capacity, trust propensity) are typically estimated from small-sample laboratory experiments. Developing methods to calibrate these parameters from large-scale naturalistic driving datasets would unlock the full potential of cognitively grounded simulation. The recent availability of datasets combining vehicle trajectories with driver physiological signals (eye tracking, electrodermal activity) opens new possibilities for this calibration.

To summarize, cognitive and physics-informed methods provide value that purely data-driven approaches cannot replicate. They predict the variability and failure modes that matter most for safety, their parameters carry psychological or physical meaning that allows principled transfer across populations and scenarios, and they capture the bounded rationality that fundamentally distinguishes human driving from the optimization-based behavior of automated systems. Therefore, the integration challenge represents one of the most important and valuable research directions in behavior modeling for automated vehicle simulation.

\section{CHRONOLOGICAL LANDSCAPE OF AI METHODS FOR DRIVING BEHAVIOR MODELING}\label{sec:timeline}

\begin{figure*}
    \centering
    \includegraphics[width=0.95\linewidth]{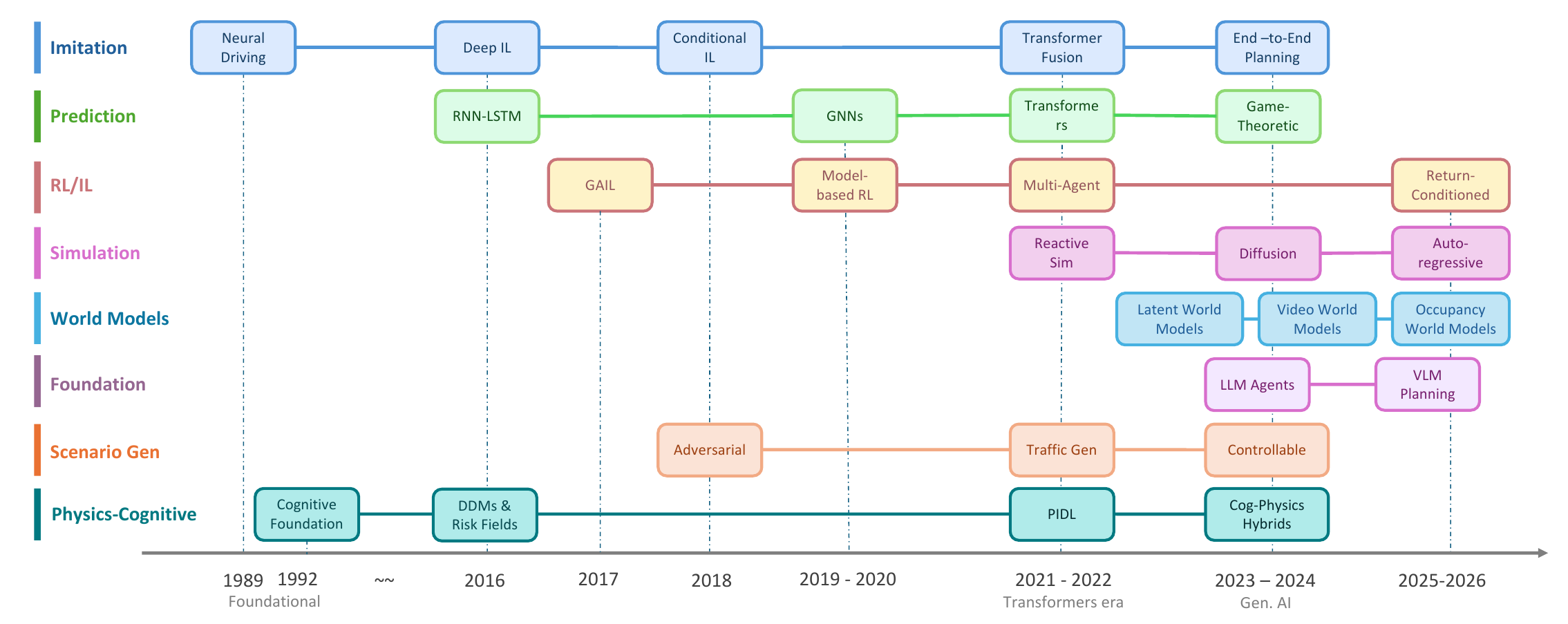}
    \caption{Timeline of model development in driving simulation and virtual evaluation}
    \label{fig:timeline}
\end{figure*}

Figure~\ref{fig:timeline} presents a timeline view of the AI methods reviewed in this survey, organized by methodological family and mapped to the period in which key contributions emerged. The timeline reveals three broad eras of development. The foundational era (pre-2016) was marked by pioneering but mainly isolated efforts, most notably early neural network--based imitation learning and recurrent architectures for trajectory prediction. The transformers era (roughly 2021–2022) brought a convergence of advances: transformer-based architectures permeated nearly every methodological family simultaneously, from prediction and imitation learning to multi-agent reinforcement learning and reactive simulation, while latent world models and traffic scenario generation methods also matured during this period. The most recent generative AI era (2023–2024) is characterized by an explosion in both breadth and diversity of approaches: diffusion-based simulation, occupancy and video world models, LLM/VLM-based planning, game-theoretic prediction, and controllable scenario generation all emerged in a compressed timeframe. 

Three patterns in this evolution are particularly noteworthy. First, there is a clear top-down temporal cascade across abstraction levels: agent-level methods such as imitation and prediction matured earliest, followed by environment-level approaches like simulation and world models, and most recently by foundation model methods that attempt to unify perception, reasoning, and planning. Second, the timeline illustrates an accelerating methodological convergence. While early work in each family developed largely in isolation, recent methods increasingly draw on shared architectural building blocks (e.g., transformers, diffusion processes) and blur traditional boundaries between prediction, planning, and simulation. This convergence suggests that the field is moving toward more holistic frameworks. However, the cognitive and physics-informed methods seem to follow a distinct developmental trajectory compared to the data-driven families. Rather than advancing through architectural innovations (from RNNs to transformers to diffusion models), this family evolved from established theoretical foundations in psychology, human factors, and traffic flow theory toward computational integration with deep learning. This dual evolution highlights a maturing field that must reconcile the rapid scalability of data-driven architectures with the slower but more principled development of theory-grounded models to achieve truly realistic mixed autonomy simulation.

\section{EVALUATION AND BENCHMARKING}
\label{sec:evaluation}

For learning-based behavior models in mixed autonomy traffic, the core question is whether simulated rollouts are (1) behaviorally realistic for both automated vehicles and human drivers, (2) interactive under feedback (agents react plausibly to each other), and (3) safe and rule-compliant under controlled stress. This section presents evaluation protocols, metrics, and benchmarking practices organized by evaluation paradigm and application context.

\subsection{Open-Loop Evaluation Metrics}
\label{subsec:openloop_metrics}

Open-loop evaluation compares predicted trajectories or actions against ground-truth recordings without environmental feedback. Given predicted trajectory $\hat{\mathbf{Y}} = \{\hat{y}_1, \ldots, \hat{y}_T\}$ and ground-truth $\mathbf{Y} = \{y_1, \ldots, y_T\}$ over prediction horizon $T$, standard metrics quantify displacement accuracy, multi-modal coverage, and probabilistic calibration.

\paragraph{Displacement metrics.}
The most widely adopted metrics measure Euclidean distance between predictions and ground-truth. \textit{Average Displacement Error} (ADE) computes the mean L2 distance across all predicted timesteps:
\begin{equation}
\text{ADE} = \frac{1}{T_f} \sum_{t=1}^{T_f} \|\hat{y}_t - y_t\|_2
\end{equation}

\textit{Final Displacement Error} (FDE) measures accuracy at the prediction horizon endpoint:
\begin{equation}
\text{FDE} = \|\hat{y}_{T_f} - y_{T_f}\|_2
\end{equation}

For multi-modal predictions generating $K$ trajectory hypotheses $\{\hat{\mathbf{Y}}^{(k)}\}_{k=1}^{K}$, the minimum variants select the best-matching mode:
\begin{equation}
\text{minADE}_K = \min_{k \in \{1,\ldots,K\}} \frac{1}{T_f} \sum_{t=1}^{T_f} \|\hat{y}_t^{(k)} - y_t\|_2, 
\end{equation}

\begin{equation}
\text{minFDE}_K = \min_{k \in \{1,\ldots,K\}} \|\hat{y}_{T_f}^{(k)} - y_{T_f}\|_2
\end{equation}

Common choices include $K \in \{1, 5, 6, 10\}$ depending on benchmark conventions~\cite{ettinger2021womd,wilson2021argoverse2}.

\paragraph{Miss rate and probabilistic metrics.}
\textit{Miss Rate} (MR) measures the fraction of predictions where no hypothesis falls within distance threshold $\delta$ of ground-truth:
\begin{equation}
\text{MR}_K^{\delta} = \mathbb{1}\left[\min_{k} \|\hat{y}_T^{(k)} - y_T\|_2 > \delta\right]
\end{equation}
Typical thresholds include $\delta = 2.0$m for the Waymo Open Motion Dataset~\cite{ettinger2021womd}. The \textit{Brier minimum FDE} used in Argoverse~2~\cite{wilson2021argoverse2} jointly evaluates displacement and probability calibration:
\begin{equation}
\text{brier-minFDE} = \min_{k} \|\hat{y}_T^{(k)} - y_T\|_2 + (1 - p_k)^2
\end{equation}
where $p_k$ is the predicted probability for mode $k$. For methods outputting trajectory distributions, \textit{Negative Log-Likelihood} (NLL) measures how well the predicted distribution covers ground-truth. \textit{Soft mAP} from the Waymo Motion Prediction Challenge adapts object detection Average Precision to trajectory forecasting, using soft assignment based on distance thresholds across multiple horizons and incorporating precision-recall curves over semantic trajectory modes.

\paragraph{Map and kinematic compliance.}
Beyond positional accuracy, practical evaluation requires checking physical feasibility. \textit{Off-road Rate} measures the fraction of predicted positions falling outside drivable areas. \textit{Overlap Rate} (OR) quantifies the frequency of predicted agent trajectories whose bounding boxes overlap with other agents or environment, signaling physically implausible predictions. \textit{Kinematic Feasibility} checks whether predicted trajectories satisfy velocity, acceleration, and curvature bounds consistent with vehicle dynamics.

\subsection{Closed-Loop Evaluation Metrics}
\label{subsec:closedloop_metrics}

Closed-loop evaluation deploys models as interactive agents within simulation by measuring realized behavior over extended rollouts. Unlike open-loop metrics comparing against fixed recordings, closed-loop metrics assess how agents respond to evolving scenarios and interventions~\cite{caesar2021nuplan,montali2023wosac}.

\paragraph{Safety metrics.}
\textit{Collision Rate} (CR) measures the fraction of episodes involving collisions:
\begin{equation}
\mathrm{CR} = \frac{1}{N_{\mathrm{ep}}}\sum_{i=1}^{N_{\mathrm{ep}}} \mathbf{1}\!\left[\text{collision in episode } i\right]
\end{equation}
where $N_{\text{ep}}$ is number of episodes.

\textit{Post-Encroachment Time} (PET) measures the time gap between consecutive occupancy of the same conflict region by different agents, and is commonly used to quantify intersection conflicts:
\begin{equation}
\mathrm{PET}_{ij}(R) = t^{\text{enter}}_{j}(R) - t^{\text{exit}}_{i}(R) \quad \text{if } t^{\text{enter}}_{j}(R) \ge t^{\text{exit}}_{i}(R)
\end{equation}
where $R$ denotes a conflict region, $t^{\text{exit}}_{i}(R)$ is the time agent $i$ exits $R$, and $t^{\text{enter}}_{j}(R)$ is the time agent $j$ enters $R$ (with $i$ being the first agent to traverse $R$ and $j$ the second).

\paragraph{Progress and task completion.}
\textit{Route Completion Rate} measures the fraction of intended route successfully traversed. \textit{Goal Achievement Rate} indicates whether agents reach designated destinations within time limits. The nuPlan benchmark~\cite{caesar2021nuplan,karnchanachari2024towards} introduces a \textit{Closed-Loop Score} (CLS) as a weighted aggregate combining progress (fraction of route completed), safety-related violations (collision, off-road, wrong direction), comfort metrics (acceleration, jerk), speed-limit adherence, and minimum TTC, with multiplier penalties that zero the score for any collision, off-road departure, or failure to make progress.

\paragraph{Comfort and rule compliance.}
\textit{Longitudinal Jerk} and \textit{Lateral Acceleration} measure ride comfort through motion smoothness. Metrics are typically reported as maximum values or fraction of time exceeding comfort thresholds (e.g., jerk $> 4$ m/s$^3$, lateral acceleration $> 3$ m/s$^2$). Traffic rule adherence is evaluated through \textit{Red Light Violation Rate}, \textit{Speed Limit Compliance}, and \textit{Lane Violation Rate}, aggregated into composite \textit{Traffic Rule Score}.

\subsection{Multi-Agent and Interaction Metrics}
\label{subsec:multiagent_metrics}

Multi-agent simulation requires metrics capturing not only individual agent quality but also scene-level consistency and interaction realism~\cite{montali2023wosac}.

\paragraph{Scene-level consistency.}
\textit{Scene Collision Rate} measures collisions between any pair of simulated agents (excluding ego), indicating whether joint rollouts produce physically consistent traffic. For trajectory prediction, \textit{Joint minADE/FDE} evaluates coordinated accuracy across multiple agents simultaneously rather than averaging independent scores.

\paragraph{Interaction realism.}
The Waymo Open Sim Agents Challenge (WOSAC)~\cite{montali2023wosac} introduced metrics specifically targeting behavioral realism for simulation agents. The \textit{Realism Meta-Metric} combines multiple distributional comparisons between simulated and real traffic, computed via negative log-likelihood of logged futures under densities estimated from model rollouts. The NLL objective WOSAC aims to approximate is:
\begin{equation}
\mathrm{NLL}^{\ast}
=
-\frac{1}{|\mathcal{D}|}\sum_{i=1}^{|\mathcal{D}|}
\log q_{\mathrm{world}}\!\left(\mathbf{o}_{\ge 1,i}\mid \mathbf{o}_{<1,i}\right)
\end{equation}
To avoid scoring the full high-dimensional future $\mathbf{o}_{\ge 1,i}$, WOSAC parameterizes scenarios using a set of component measurements and computes a time-series likelihood for each component metric $m$ as an average (in log-space) over time, masked by validity $v_t$:
\begin{equation}
m
=
\exp\!\left(
-\frac{\sum_{t}\mathbf{1}\{v_t\}\,\mathrm{NLL}_t}
{\sum_{t}\mathbf{1}\{v_t\}}
\right)
\end{equation}
After obtaining component metrics for each measurement, WOSAC aggregates them into a single composite metric $M^{K}$:
\begin{equation}
M^{K}
=
\frac{1}{N M}\sum_{i=1}^{N}\sum_{j=1}^{M} w_j\, m^{K}_{i,j},
\qquad
\sum_{j=1}^{M} w_j = 1
\end{equation}
where $N$ is the number of scenarios, $M=9$ is the number of component metrics, and $K=32$ is the number of stochastic rollouts. The metric aggregates three categories: \textit{Kinematic Metrics} comparing distributions of linear speed, linear acceleration magnitude, angular speed, and angular acceleration magnitude; \textit{Interactive Metrics} comparing TTC distributions, distance to nearest object, and collision likelihood; and \textit{Map-Based Metrics} comparing distance to road edge, road departure frequency, and off-road rate. Collisions and road departures are typically double-weighted to emphasize safety.

\paragraph{Interaction-specific evaluation.}
For targeted interaction assessment, metrics include \textit{Gap Acceptance Rate} at merging scenarios, \textit{Yielding Compliance} at unprotected turns, \textit{Time Headway} (THW) distributions for car-following behavior, and \textit{Lane Change Success Rate}. The INTERACTION dataset~\cite{zhan2019interaction} and highD dataset~\cite{krajewski2018highd} emphasize such scenario-specific metrics, with INTERACTION covering diverse international driving contexts (roundabouts, intersections, merging) and highD providing naturalistic highway trajectories from German highways for calibration and validation.

\section{SIMULATION TOOLS AND DATASETS}
\label{sec:dataset_sim}

This section reviews the simulation platforms, datasets, and benchmarks used to train and evaluate learning-based traffic behavior models in mixed autonomy.

\subsection{Datasets for Behavior Modeling}
\label{subsec:datasets}

Large-scale naturalistic driving datasets have been essential in advancing AI-based behavior modeling for automated vehicles. These datasets provide the trajectory data, sensor recordings, and contextual information necessary for training and evaluating prediction, planning, and simulation models. Table~\ref{tab:datasets} summarizes the major publicly available datasets organized by their primary characteristics: geographic coverage, recording duration, agent diversity, and key features.

Early trajectory datasets such as NGSIM~\cite{colyar2007ngsim} established foundational benchmarks for car-following and lane-changing analysis. However, they were limited in scale and scenario diversity. The drone-based recording methodology pioneered by the highD dataset~\cite{krajewski2018highd} and its successors (inD~\cite{bock2020ind}, rounD~\cite{krajewski2020round}, exiD~\cite{moers2022exid}) enabled high-precision trajectory extraction. This set of datasets provide naturalistic behavior data for highways, intersections, roundabouts, and merging scenarios across German road networks. The INTERACTION dataset~\cite{zhan2019interaction} extended this approach internationally and covers diverse driving cultures and highly interactive scenarios, including adversarial and cooperative maneuvers with semantic map annotations. Finally, pNEUMA~\cite{barmpounakis2020pneuma} provides a large-scale drone-based trajectory dataset recorded over the congested central business district of Athens using a coordinated swarm of 10 drones over five days. It contains high-frequency trajectories across multiple traffic modes (cars, buses, taxis, powered two-wheelers, and heavy vehicles).

Modern large-scale datasets have shifted toward multimodal sensor fusion and standardized benchmarking. The Waymo Open Motion Dataset (WOMD)~\cite{ettinger2021womd} provides over 570 hours of driving data across six U.S. cities with high-fidelity trajectory annotations, HD maps, and curated interactive splits emphasizing multi-agent scenarios. Argoverse~2~\cite{wilson2021argoverse2} offers 250,000 motion forecasting scenarios with 3D lane boundaries and ground height information, while nuScenes~\cite{caesar2020nuscenes} provides diverse urban driving with full sensor suite annotations. For closed-loop planning evaluation, nuPlan~\cite{caesar2021nuplan,karnchanachari2024towards} represents a paradigm shift by offering 1,282 hours of data with scenario taxonomies, reactive simulation capabilities, and planning-specific metrics across four cities with distinct driving cultures. These datasets collectively enable the training and rigorous evaluation of the behavior modeling methods reviewed in this survey, with dataset selection depending on the target task (prediction vs.\ planning vs.\ simulation), required scenario diversity, and evaluation paradigm (open-loop vs.\ closed-loop).

\begin{table*}[htbp]
\centering
\scriptsize
\setlength{\tabcolsep}{2.5pt}
\renewcommand{\arraystretch}{1.2}
\caption{Major publicly available datasets for trajectory prediction, behavior modeling, and planning in automated driving. Duration indicates total recording time; scenarios indicates number of extracted segments for benchmarking where applicable.}
\label{tab:datasets}
\begin{tabularx}{\linewidth}{>{\raggedright\arraybackslash}p{0.13\linewidth} >{\raggedright\arraybackslash}p{0.11\linewidth} >{\raggedright\arraybackslash}p{0.08\linewidth} >{\raggedright\arraybackslash}p{0.11\linewidth} >{\raggedright\arraybackslash}p{0.08\linewidth} >{\raggedright\arraybackslash}p{0.12\linewidth} >{\raggedright\arraybackslash}p{0.11\linewidth} >{\raggedright\arraybackslash}p{0.16\linewidth}}
\toprule
\textbf{Dataset} & \textbf{Location(s)} & \textbf{Year} & \textbf{Duration / Scenarios} & \textbf{Freq. (Hz)} & \textbf{Environment} & \textbf{Agent Types} & \textbf{Key Features} \\
\midrule
\multicolumn{8}{l}{\textit{\textbf{Drone-Based Trajectory Datasets}}} \\
\midrule
NGSIM~\cite{colyar2007ngsim} & CA, GA (USA) & 2005--06 & 90 min & 10 & Highway, arterial & Vehicles & Foundational; car-following analysis \\
highD~\cite{krajewski2018highd} & Germany & 2018 & 16.5 h / 110k vehicles & 25 & Highway (6 sites) & Cars, trucks & High precision ($<$10 cm); 5,600 lane changes \\
inD~\cite{bock2020ind} & Germany & 2020 & 10 h / 11.5k road users & 25 & Urban intersections (4 sites) & Vehicles, cyclists, pedestrians & Naturalistic intersection behavior \\
rounD~\cite{krajewski2020round} & Germany & 2020 & 6 h / 13k road users & 25 & Roundabouts (3 sites) & Vehicles, cyclists, pedestrians & Complex interaction patterns \\
pNEUMA~\cite{barmpounakis2020pneuma} & Athens (Greece) & 2018 & 12.5 h / 0.5M trajectories & 25 & Urban CBD (congested) & Cars, taxis, buses, PTWs, medium/heavy vehicles & Swarm of 10 drones over 5 days; dense multimodal trajectories over large urban network \\
exiD~\cite{moers2022exid} & Germany & 2022 & 16 h / 69k vehicles & 25 & Highway ramps (7 sites) & Vehicles & Merging and lane change scenarios \\
INTERACTION~\cite{zhan2019interaction} & USA, China, Germany & 2019 & 16.5 h / 40k+ trajectories & 10 & Mixed (roundabouts, intersections, highways) & Vehicles & International; adversarial \& cooperative scenarios; semantic maps \\
\midrule
\multicolumn{8}{l}{\textit{\textbf{Vehicle-Mounted Sensor Datasets (Motion Forecasting)}}} \\
\midrule
nuScenes~\cite{caesar2020nuscenes} & Boston, Singapore & 2020 & 5.5 h / 1,000 scenes & 2 (keyframes) & Urban & Vehicles, pedestrians, cyclists & Full sensor suite; 23 classes; 3D annotations \\
Argoverse 1~\cite{chang2019argoverse} & Miami, Pittsburgh & 2019 & 320 h / 324k scenarios & 10 & Urban & Vehicles & First large-scale; HD maps with centerlines \\
Argoverse 2~\cite{wilson2021argoverse2} & 6 U.S. cities & 2023 & --- / 250k scenarios & 10 & Urban & 30 classes & 3D lane boundaries; ground height; 6s prediction \\
Waymo Open Motion~\cite{ettinger2021womd} & 6 U.S. cities & 2021 & 570+ h / 103k segments & 10 & Urban, suburban & Vehicles, pedestrians, cyclists & Interactive split; realism meta-metric; WOSAC benchmark \\
Lyft Level 5~\cite{houston2021one} & Palo Alto (USA) & 2021 & 1,118 h / 170k scenes & 10 & Urban & Vehicles, pedestrians, cyclists & Semantic maps; large scale \\
\midrule
\multicolumn{8}{l}{\textit{\textbf{Planning and Closed-Loop Evaluation Datasets}}} \\
\midrule
nuPlan~\cite{caesar2021nuplan,karnchanachari2024towards} & Boston, Pittsburgh, Las Vegas, Singapore & 2021--24 & 1,282 h / 75 scenario types & 20 & Urban (4 cities) & Vehicles, pedestrians, cyclists & Closed-loop benchmark; reactive agents; planning metrics \\
CommonRoad~\cite{althoff2017commonroad} & Synthetic + recorded & 2017 & --- / 5,700 scenarios & varies & Highway, urban & Vehicles & Motion planning benchmark; formal specifications \\
\bottomrule
\end{tabularx}
\end{table*}

\subsection{Simulation Tools}
\label{subsec:simtools}

Table~\ref{tab:simtools} summarizes widely used simulation tools for developing and evaluating learning-based driving methods. Tool selection depends on the target application, required fidelity level, dataset compatibility, and computational constraints. High-throughput state-level simulators (GPUDrive, Waymax, Nocturne) are most suitable for methods requiring billions of environment steps, typical in reinforcement learning, self-play, and large-scale ablations. Sensor-level simulators (CARLA, AWSIM, Isaac Sim) are preferred when contributions depend on perception realism, sensor modeling, or full autonomy stack integration. Traffic-level simulators (SUMO, CityFlow) support network-wide analysis and mixed autonomy penetration studies. Neural sensor simulators (UniSim, VISTA) generate realistic sensor outputs from real driving logs, which allows closed-loop evaluation under counterfactual scenarios.

\begin{table*}[t]
\centering
\scriptsize
\setlength{\tabcolsep}{3pt}
\renewcommand{\arraystretch}{1.2}
\caption{Simulation tools commonly used with learning-based driving methods. Simulation level indicates output granularity: \textit{sensor} (camera/LiDAR images), \textit{state} (bounding boxes, trajectories), or \textit{traffic} (aggregate flow). Background agents refers to how non-ego vehicles are modeled during simulation.}
\label{tab:simtools}
\begin{tabularx}{\linewidth}{>{\raggedright\arraybackslash}p{0.11\linewidth} >{\raggedright\arraybackslash}p{0.17\linewidth} >{\raggedright\arraybackslash}p{0.08\linewidth} >{\raggedright\arraybackslash}p{0.17\linewidth} >{\raggedright\arraybackslash}p{0.17\linewidth} >{\raggedright\arraybackslash}p{0.17\linewidth}}
\toprule
\textbf{Tool} & \textbf{Primary Application} & \textbf{Sim Level} & \textbf{Supported Datasets} & \textbf{Background Agents} & \textbf{Throughput} \\
\midrule
\multicolumn{6}{l}{\textit{\textbf{State-Level / Data-Driven Simulators}}} \\
\midrule
GPUDrive & Multi-agent RL training & State & WOMD & Learned policies, rule-based & Very high (GPU-native, 1M+ steps/s) \\
Waymax & Motion planning, prediction evaluation & State & WOMD & Log replay, IDM, learned & High (JAX/GPU/TPU batching) \\
nuPlan Devkit & Planning benchmark and evaluation & State & nuPlan & IDM, log replay, learned (SMART) & Moderate (multi-process) \\
Nocturne & Multi-agent RL research & State & WOMD-derived scenarios & Log replay, learned (BC baseline) & High (C++ core, $>$2k steps/s) \\
highway-env & RL algorithm prototyping & State & Synthetic (procedural) & IDM, MOBIL & High (vectorized environments) \\
MetaDrive & RL training, generalization & State & Procedural, WOMD, nuScenes, nuPlan, Lyft & Rule-based (IDM), log replay & High (1k+ FPS) \\
SMARTS & Multi-agent RL, social driving & State & NGSIM, Argoverse, WOMD, SUMO & Social agents, rule-based, log replay & Moderate (multi-instance) \\
\midrule
\multicolumn{6}{l}{\textit{\textbf{Traffic-Level Simulators}}} \\
\midrule
SUMO & Traffic flow, mixed autonomy & Traffic & OpenStreetMap, custom & Car-following (Krauss, IDM) & High (large-scale networks) \\
Flow & RL for traffic control & Traffic & SUMO networks & SUMO models & Moderate (RL overhead) \\
CityFlow & Traffic signal optimization & Traffic & Custom networks & Rule-based & High (city-scale) \\
\midrule
\multicolumn{6}{l}{\textit{\textbf{Sensor-Level / 3D Simulators}}} \\
\midrule
CARLA & End-to-end driving, perception & Sensor & Custom, OpenDRIVE maps & Autopilot (rule-based), traffic manager & Moderate (rendering-limited) \\
AWSIM & Autoware stack testing & Sensor & Lanelet2 maps & Rule-based & Moderate \\
Isaac Sim & Perception, robotics & Sensor & Custom environments & Scripted behaviors & Moderate (GPU rendering) \\
BeamNG.tech & Vehicle dynamics, crash testing & Sensor & Custom scenarios & Scripted, AI traffic & Moderate \\
\midrule
\multicolumn{6}{l}{\textit{\textbf{Neural / Data-Driven Sensor Simulators}}} \\
\midrule
VISTA 2.0 & Data-driven policy learning & Sensor & Custom logs (MIT AVT) & Log-based with viewpoint synthesis & Dataset-dependent \\
UniSim & Closed-loop counterfactual testing & Sensor & PandaSet, custom logs & Neural reconstruction, actor manipulation & Moderate (GPU inference) \\
\midrule
\multicolumn{6}{l}{\textit{\textbf{Scenario Engines and Benchmarks}}} \\
\midrule
esmini & OpenSCENARIO execution & State & OpenDRIVE/OpenSCENARIO & Scripted trajectories & Lightweight \\
CommonRoad & Motion planning benchmark & State & Recorded + synthetic & Replay, interactive & Offline evaluation \\
\bottomrule
\end{tabularx}
\end{table*}

\section{DISCUSSION}
\label{sec:discussion}

The methods reviewed in this survey cover a wide range. While each section has identified challenges specific to individual method families, a broader view of the field reveals several cross-cutting themes that can guide future efforts in the domain of AI-driven mixed autonomy traffic simulation. This section synthesizes these themes into nine actionable challenges and future directions that go beyond individual methods.

\subsection{The Causality Gap: From Correlation to Counterfactual Validity}

Perhaps the most pervasive challenge across all methods surveyed is the reliance on observational training data to build models that must function under \textit{interventional} conditions. Imitation learning, trajectory prediction, multi-agent imitation, and even generative world models all learn from passively recorded driving logs, while the primary use case for simulation demands \emph{counterfactual} reasoning: what would surrounding agents do if the ego vehicle behaved differently from what was recorded? This gap between observational training and interventional deployment is not merely a technical limitation of any single method family; it is a fundamental constraint that limits causal validity under policy interventions. Current approaches address this indirectly through data augmentation, adversarial perturbation, or self-play, but none provide principled guarantees of counterfactual validity. Bridging this gap will likely require several measures, such as large-scale collection of interventional driving data (e.g., through controlled AV deployments that deliberately probe human responses), development of causal inference frameworks adapted to multi-agent sequential decision-making, and hybrid architectures that combine data-driven pattern matching with structured causal models that encode domain knowledge about driver response mechanisms. The intersection of causal machine learning and traffic simulation remains underdeveloped, but it is a high-impact research direction.

\subsection{The Evaluation Crisis}

A recurring finding throughout this survey is that evaluation methodology lags behind modeling methodology. Open-loop displacement metrics (minADE, minFDE) remain the dominant evaluation paradigm despite mounting evidence that they correlate poorly with closed-loop simulation quality. Models that achieve state-of-the-art prediction accuracy can produce unsafe or unrealistic behavior when deployed as reactive simulation agents. Closed-loop benchmarks such as WOSAC and nuPlan represent important progress, but they introduce their own complications: sensitivity to evaluation infrastructure assumptions, dependence on how ego behavior is specified, lack of standardized intervention protocols, and in the case of WOSAC, reliance on available fine-grained datasets. More fundamentally, the field lacks consensus on what ``realistic simulation'' means operationally. Is the goal to match distributional statistics of real traffic (a statistical fidelity criterion), to produce plausible reactions to arbitrary ego interventions (a counterfactual validity criterion), or to reproduce the logged data or generative cognitive mechanisms behind human driving? These are distinct objectives that may require different models, training paradigms, and evaluation metrics. Developing a unified evaluation framework that addresses behavioral realism, interaction quality, counterfactual robustness, and long-horizon stability simultaneously, while remaining computationally tractable and reproducible across research communities, is one of the most important infrastructure challenges facing the field.

\subsection{The Scalability, Fidelity, and Controllability Trilemma}

Across method families, a persistent three-way tension emerges between scalability, behavioral fidelity, and controllability. High-fidelity multi-agent simulators like diffusion-based reactive simulators produce realistic joint rollouts but face computational costs that limit their applicability to large-scale testing campaigns. Scalable approaches such as parameter-shared policies or autoregressive token-based simulators achieve throughput suitable for reinforcement learning and large-scale ablations, but may sacrifice behavioral diversity or interaction quality. Controllable methods that allow test engineers to specify desired behavioral profiles (e.g., aggressive merging, hesitant yielding) risk pushing agents beyond the support of training data into ``sim-native'' behaviors that are internally consistent but not grounded in observed human driving. To the best of our knowledge, no current method satisfactorily resolves all three requirements simultaneously. Future architectures may need to adopt hierarchical designs that allocate computational resources adaptively. They can use expensive high-fidelity reasoning selectively for complex interactions and use lightweight policies for routine driving. They also need calibrated uncertainty estimates that indicate when controlled behaviors depart from training data.

\subsection{The Mixed Autonomy Representation Problem}

While ``mixed autonomy'' is the motivating context for this entire survey, surprisingly few methods explicitly model the heterogeneous behavioral dynamics that arise when AVs and human drivers share road infrastructure. Most current approaches either model all agents as human-like (ignoring the distinct behavioral signatures of AVs) or focus exclusively on ego-vehicle planning (ignoring how human drivers adapt to AV presence). The critical missing ingredient is the \emph{co-adaptation} between human drivers and AVs: as AV penetration rates increase, human drivers encounter AVs more frequently and may adjust their gap acceptance, following distances, and interaction strategies. Trust dynamics, as reviewed in Section~\ref{sec:cognitive_ai}, offer one formalization of this adaptation, but integration into scalable multi-agent simulation remains an open research direction. Constructing simulation environments that capture the evolving dynamics of the transition period, rather than assuming static behavioral distributions, is essential for realistic testing of AV deployment strategies. This requires not only new modeling approaches but also new datasets that capture longitudinal human-AV interaction patterns across varying AV penetration rates, which currently do not exist at scale.

\subsection{Bridging the Research and Deployment Gap}

Currently, a notable disconnect exists between the methods developed in the research community and what simulation platforms actually deploy. Mainstream simulation tools (CARLA, SUMO, VISSIM) still predominantly rely on rule-based car-following and lane-changing models or log replay for background traffic, despite the rich landscape of learned behavior models reviewed in this survey. This gap persists for practical reasons: learned models require careful integration with map representations, traffic rule enforcement, and scenario specification workflows that research prototypes rarely address. They also lack the interpretability and predictability that safety engineers require for systematic test design. Addressing these issues demands not only continued improvement in model accuracy and efficiency but also attention to engineering concerns, such as standardized APIs for behavior model integration, modular architectures that allow mixing learned and rule-based agents within the same simulation, and validation frameworks that quantify when a learned model can be trusted as a substitute for a calibrated rule-based model. The emergence of data-driven simulation platforms (Waymax, GPUDrive, Nocturne)~\cite{NEURIPS2023_1838feeb, kazemkhani2025gpudrive, vinitsky2022nocturne} designed specifically for learned behavior models represents a promising trend, but these remain largely confined to research settings.

\subsection{Data Limitations and Geographic Bias}

The datasets underpinning the methods reviewed in this survey exhibit notable geographic and cultural concentration. The dominant benchmarks, such as Waymo Open Motion Dataset, Argoverse, LevelX  Dataset, nuScenes, and nuPlan, are collected primarily in a handful of U.S. and European cities, and Singapore. Driving behavior, however, is deeply influenced by local traffic culture, road infrastructure, regulatory environments, and social norms. Models trained on U.S. highway data may produce implausible behavior in the dense, mixed-mode traffic of South and Southeast Asian cities, or in the roundabout-heavy networks of European towns. Beyond geographic bias, existing datasets overwhelmingly capture nominal driving conditions, with safety-critical events (near-misses, erratic behaviors, AV-human conflicts) dramatically underrepresented relative to their importance for simulation-based AV testing. Addressing these limitations requires concerted data collection efforts across diverse geographies and interaction contexts, as well as synthetic data generation techniques that can augment rare-event coverage without sacrificing distributional validity. Moreover, longitudinal collection of AV datasets is essential for studying and modeling behavioral adaptation in mixed autonomy traffic.

\subsection{Toward Unified Architectures}

The chronological analysis presented in Section~\ref{sec:timeline} reveals accelerating methodological convergence: transformers, diffusion processes, and autoregressive generation now serve as shared building blocks across prediction, simulation, world modeling, and scenario generation. This convergence suggests that the traditional boundaries between single-agent and multi-agent methods, between agent-level and environment-level simulation, and between behavior modeling and scenario generation are becoming increasingly artificial. The logical endpoint of this trajectory is unified simulation architectures that jointly generate environment evolution, multi-agent behavior, and scenario conditions within a single learned framework. This can combine the strengths of world models (environment dynamics), reactive simulators (interaction fidelity), and scenario generators (controllability) into cohesive systems. Early steps in this direction are visible in recent works that combine diffusion-based scene generation with reactive rollout capabilities. However, the risk of monolithic architectures is reduced interpretability and reduced ease of debugging. A more promising path may be modular but differentiable pipelines where specialized components (perception, prediction, interaction reasoning, and physical dynamics) are connected but remain individually interpretable and replaceable.

\subsection{Simulation-to-Real Transfer Gap}
Despite the increasing realism of AI-driven traffic simulators, a persistent challenge remains the gap between simulation performance and real-world behavior. Many behavior models are trained and evaluated in simulated environments whose dynamics, interaction patterns, and long-tail events only partially reflect real traffic conditions. As a result, policies that perform well in simulation may fail to generalize when exposed to the variability, uncertainty, and rare edge cases encountered in real deployments. This issue is particularly pronounced in mixed autonomy traffic, where subtle differences in human driving behavior, local traffic culture, and infrastructure can significantly affect interaction dynamics. Bridging this simulation-to-real gap requires improved environment fidelity, better calibration against real-world data, and evaluation frameworks that explicitly measure transferability between simulated and real-world settings. Hybrid approaches combining data-driven learning with physics-based constraints and real-world validation pipelines represent promising directions for improving the reliability of simulation-trained models.

\subsection{The Role of Cognitive Grounding}

Finally, the cognitive and physics-informed methods reviewed in this survey highlight a fundamental research direction. The dominant research trend is toward larger data-driven models that match behavioral distributions with increasing fidelity. However, for mixed autonomy simulation specifically, understanding \emph{why} drivers behave as they do, not just reproducing \emph{what} they do, carries distinct practical value. Cognitively grounded models offer three capabilities that purely data-driven approaches currently lack: (1) principled generation of human error modes critical for safety testing, (2) interpretable behavioral diversity parameterized by psychologically meaningful variables rather than obscured latent information, (3) and the ability to generalize to novel interaction patterns (such as those arising from AV introduction) that lie outside training data distributions. The integration challenge is also worth noting. Cognitive architectures operate at different abstraction levels and computational costs than modern deep learning. Nevertheless, hybrid approaches that encode cognitive constraints as inductive biases within neural architectures, rather than building full cognitive simulations, represent a pragmatic and promising approach that could combine the scalability of data-driven methods with the explanatory power of cognitive science.

\section{CONCLUSION}\label{sec:conclusion}

This survey provides a comprehensive and structured review of artificial intelligence methods for modeling mixed automated and human traffic in simulation environments, organized along three complementary axes: agent-level behavior models (single-agent and multi-agent), environment-level simulation methods (world models and scenario generation), and cognitive and physics-informed approaches. Across these methodological families, several cross-cutting challenges were identified: the gap between open-loop trajectory accuracy and closed-loop interactive realism remains fundamental, as matching recorded driving logs does not guarantee plausible behavior under counterfactual interventions; the tension between data-driven fidelity and user-specified controllability limits the practical utility of generative approaches for systematic testing; long-tail safety-critical scenarios remain underrepresented in both training data and evaluation benchmarks; and standardized validation frameworks that can establish trust in learned simulators for safety assurance are still lacking. Looking ahead, the field is converging toward hybrid architectures that combine the scalability of data-driven learning with the interpretability of cognitive models and the guarantees of physics-based constraints, while foundation models and diffusion-based generation are opening new possibilities for unified perception–prediction–planning frameworks. We argue that bridging the transportation engineering and machine learning communities will be essential for realizing the full potential of AI-driven simulation in enabling the safe deployment of automated vehicles in mixed autonomy traffic. This could be achieved by jointly addressing behavioral realism, counterfactual validity, domain generalization, and integration with safety assurance pipelines.

% \section*{Acknowledgments}

%Bibliography
\bibliographystyle{unsrt}  
\bibliography{references}

\begin{IEEEbiography}[{\includegraphics[width=1in,height=1.25in,clip,keepaspectratio]{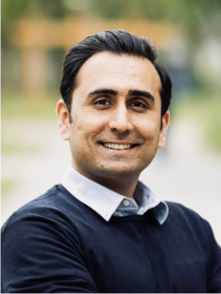}}]{Saeed Rahmani } (Graduate Student Member, IEEE) is a PhD Candidate at Delft University of Technology and a visiting scholar at Oxford University. He received the B.Sc. and M.Sc. degrees in civil engineering, transportation and traffic systems engineering. Since 2022, He has been pursuing a Ph.D. degree, focusing on motion planning and decision-making for automated vehicles focusing on mixed automated and human traffic.
\end{IEEEbiography}

\begin{IEEEbiography}[{\includegraphics[width=1in,height=1.25in,clip,keepaspectratio]{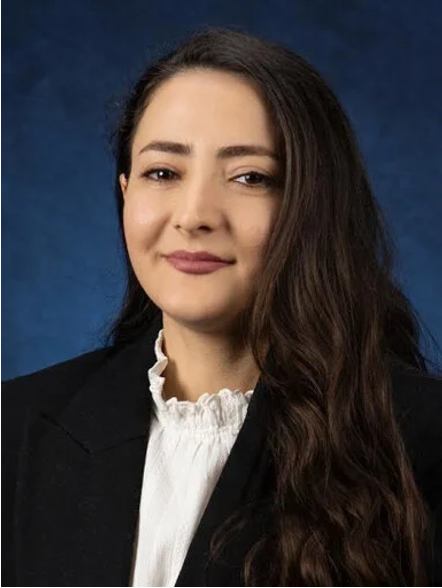}}]{Shiva Rasouli } is a Ph.D. candidate and Research Assistant in the Department of Industrial and Systems Engineering at the University of Michigan, Dearborn, Michigan, USA. Her current research interests include affective computing, human-computer interaction, driver behavior, and user state estimation to facilitate human and automation teaming.  
\end{IEEEbiography}

\begin{IEEEbiography}[{\includegraphics[width=1in,height=1.25in,clip,keepaspectratio]{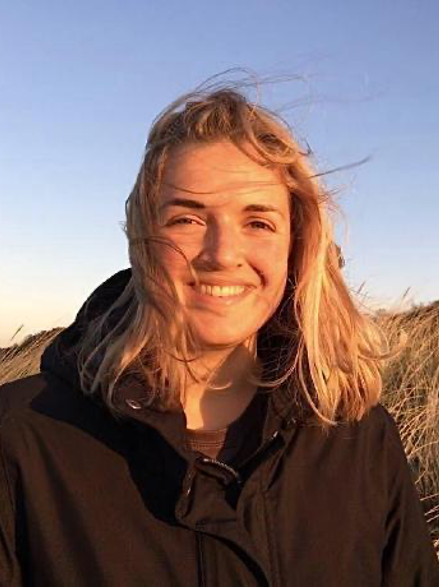}}]{Daphne Cornelisse } is a Ph.D. student at the Tandon School of Engineering, New York University, New York, United States. Her research applies reinforcement learning and imitation learning to multi-agent systems. She is particularly focused on developing effective, human-compatible simulation agents for autonomous driving, aiming to scale self-play for reliable traffic modeling and evaluation.  
\end{IEEEbiography}

\begin{IEEEbiography}[{\includegraphics[width=1in,height=1.25in,clip,keepaspectratio]{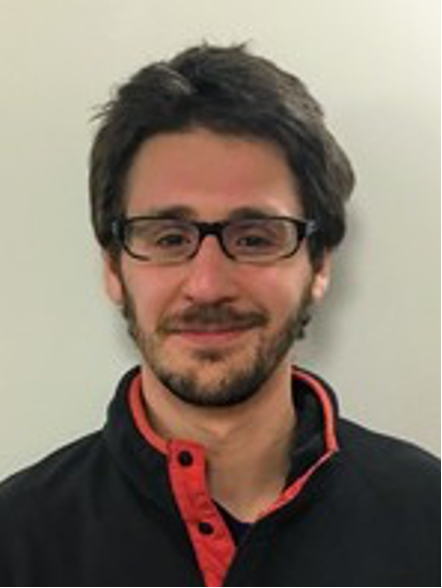}}]{Dr. Eugene Vinitsky }  is an Assistant Professor at the Tandon School of Engineering, New York University, New York, United States. He received his Ph.D. in Control from the University of California, Berkeley. His research applies multi-agent reinforcement learning to robotics and transportation, with a focus on synthesizing complex, human-like behavior from unsupervised interaction between groups of learning agents and designing scalable data-driven simulators.  
\end{IEEEbiography}

\begin{IEEEbiography}[{\includegraphics[width=1in,height=1.25in,clip,keepaspectratio]{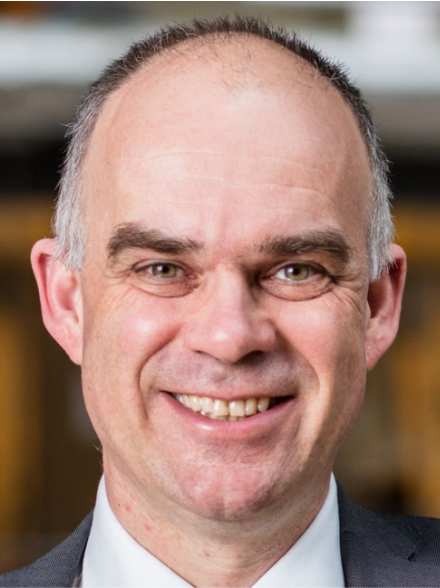}}]{Prof.Dr.ir. Bart van Arem } was appointed as full professor Transport Modelling at the department of Transport and Planning in 2009 and serves as Pro Vice Rector for Doctoral Affairs of TU Delft since 2021  He was head of the department Transport \& Planning from 2010 till 2017 and served as director of the TU Delft Transport Institute from 2012-2021. His research focuses on analysing and modelling the implications of intelligent transportation systems, such as automated, electric and shared vehicles.  
\end{IEEEbiography}

\begin{IEEEbiography}[{\includegraphics[width=1in,height=1.25in,clip,keepaspectratio]{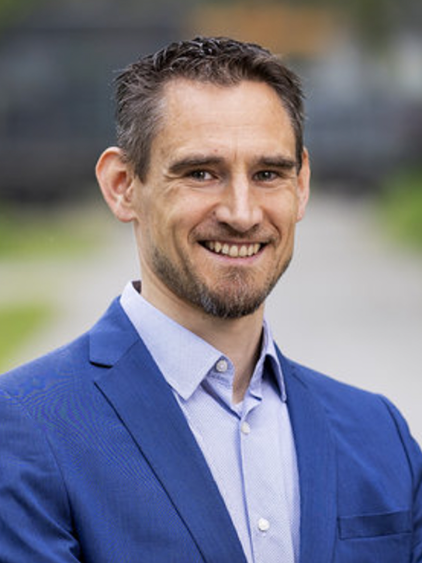}}]{Dr.ir. Simeon C. Calvert } is an Associate Professor of traffic and network management at TU Delft. He is the director of the Automated Driving \& Simulation (ADaS) Lab in the Department of Transport \& Planning and co-leads the Delft AI Lab on urban mobility behaviour: CiTy- AI. From 2010 to 2016, he worked as a Research Scientist with TNO. His research interests include ITS, impacts of vehicle automation, traffic management, traffic flow theory, and network analysis.
\end{IEEEbiography}

\end{document}